\pdfoutput=1

\documentclass[11pt]{article}

\usepackage{emnlp2022}

\usepackage{times}
\usepackage{latexsym}
\usepackage{soul}

\usepackage[T1]{fontenc}

\usepackage[utf8]{inputenc}

\usepackage{microtype}

\usepackage{graphicx} 
\usepackage{float} 
\usepackage{hyperref}
\usepackage{url}
\usepackage{booktabs}
\usepackage{multirow}
\usepackage{graphics}
\usepackage{amsmath}
\usepackage{multicol}
\usepackage{lipsum}
\usepackage{mwe}
\usepackage{hyperref}
\usepackage{url}
\usepackage{xcolor}
\usepackage[ruled,vlined]{algorithm2e}
\usepackage{bbm}
\usepackage{upgreek}
\usepackage{svg}
\usepackage{makecell}
\SetAlFnt{\small}
\usepackage{wrapfig}
\usepackage{colortbl}
\usepackage{amssymb}
\usepackage{bbm}
\usepackage{subcaption}
\usepackage{grffile}

\usepackage{amsmath,amsfonts,bm}

\def\eqref#1{equation~\ref{#1}}

\def\1{\bm{1}}

\DeclareMathAlphabet{\mathsfit}{\encodingdefault}{\sfdefault}{m}{sl}
\SetMathAlphabet{\mathsfit}{bold}{\encodingdefault}{\sfdefault}{bx}{n}

\newcommand{\R}{\mathbb{R}}

\DeclareMathOperator*{\argmax}{\arg\!\max}

\newcommand{\brihi}[1]{{\color{red} [{\bf Brihi}: #1]}}
\newcommand{\aaron}[1]{{\color{blue} [{\bf Aaron}: #1]}}
\newcommand{\ziyi}[1]{{\color{purple} [{\bf Ziyi}: #1]}}
\newcommand{\xiang}[1]{{\color{red} [{\bf Xiang}: #1]}}
\newcommand{\snie}[1]{{\color{brown} [{\bf Shaoliang}: #1]}}

\newcommand{\ertestemoji}{\raisebox{-2pt}{\includegraphics[width=1em]{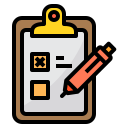}}}

\newcommand{\method}{\textsc{ER-Test}\xspace}
\newcommand{\eg}{\textit{e.g., }}
\newcommand{\ie}{\textit{i.e., }}
\newcommand{\versus}{\textit{vs. }}

\colorlet{lred}{red!15}
\colorlet{lblue}{blue!15}
\colorlet{lgreen}{green!20}
\colorlet{lpink}{pink!75}

\title{\ertestemoji \method: Evaluating Explanation \\ Regularization Methods for Language Models}

\author{Brihi Joshi$^{\clubsuit}$\thanks{~~Equal contribution.} \hspace{3mm} Aaron Chan$^{\clubsuit*}$ \hspace{3mm} Ziyi Liu$^{\clubsuit*}$ \\ \textbf{Shaoliang Nie}$^{\diamondsuit}$ \hspace{3mm} \textbf{Maziar Sanjabi}$^{\diamondsuit}$ \hspace{3mm} \textbf{Hamed Firooz}$^{\diamondsuit}$ \hspace{3mm} \textbf{Xiang Ren}$^{\clubsuit}$ \vspace{1mm} \\
$^{\clubsuit}$University of Southern California \hspace{1mm} $^{\diamondsuit}$Meta AI \\
\small{\texttt{\{brihijos, chanaaro, zliu2803, xiangren\}@usc.edu}} \\
\small{\texttt{\{snie, maziars, mhfirooz\}@fb.com}}
}

\begin{document}
\maketitle

\begin{abstract}
\renewcommand{\aaron}[1]{}
\renewcommand{\brihi}[1]{}
\renewcommand{\xiang}[1]{}
\renewcommand{\ziyi}[1]{}
\renewcommand{\snie}[1]{}



By explaining how humans would solve a given task, human rationales can provide strong learning signal for neural language models (LMs).
Explanation regularization (ER) aims to improve LM generalization by pushing the LM's machine rationales (\textit{Which input tokens did the LM focus on?}) to align with human rationales (\textit{Which input tokens would humans focus on?}).
Though prior works primarily study ER via in-distribution (ID) evaluation, out-of-distribution (OOD) generalization is often more critical in real-world scenarios, yet ER's effect on OOD generalization has been underexplored.
In this paper, we introduce \method, a framework for evaluating ER models' OOD generalization along three dimensions: unseen dataset tests, contrast set tests, and functional tests.
Using \method, we extensively analyze how ER models' OOD generalization varies with different ER design choices.
Across two tasks and six datasets, \method shows that ER has little impact on ID performance but can yield large OOD performance gains.
Also, we find that ER can improve OOD performance even with limited rationale supervision.
\method's results help demonstrate ER's utility and establish best practices for using ER effectively.\footnote{Code is available at \href{https://github.com/INK-USC/er-test}{github.com/INK-USC/ER-Test}.}

\end{abstract}

\begin{figure}[t!]
\vspace{-0.3cm}
\centering
\includegraphics[width=0.89\columnwidth]{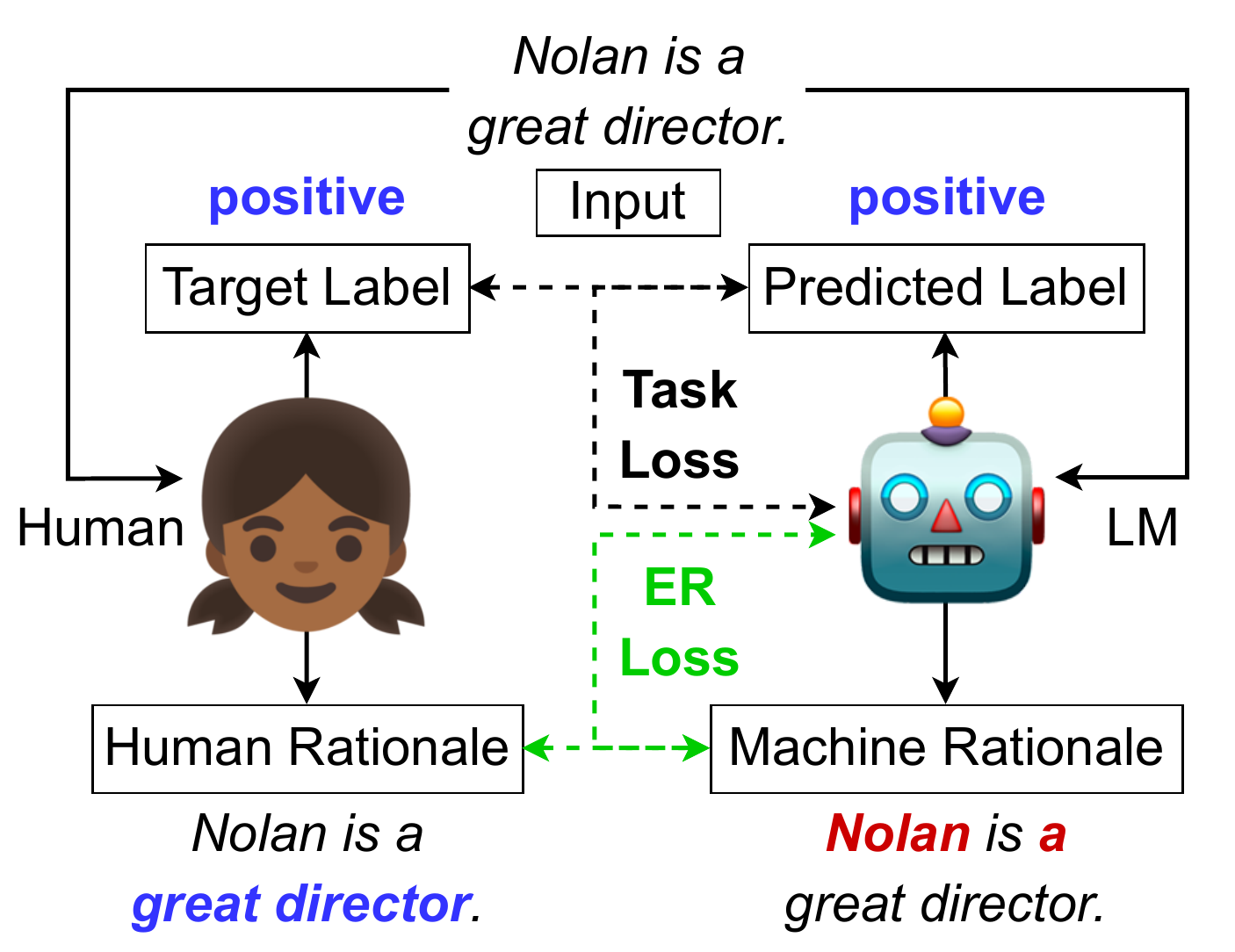}
\vspace{-0.2cm}
\caption{\small \textbf{Explanation Regularization (ER).} 
    Sometimes, task labels alone provide insufficient supervision for language model (LM) generalization. ER aims to improve generalization by training the LM so that its machine rationales (\textit{Which input tokens did the LM focus on?}) align with human rationales (\textit{Which input tokens would humans focus on?}) (\textsection \ref{sec:background}).
}
\label{fig:intro:illustration}
\vspace{-0.6cm}
\end{figure}

\section{Introduction} 
\label{sec:intro}

Neural language models (LMs) have achieved state-of-the-art performance on a broad array of natural language processing (NLP) tasks \citep{devlin2018bert,liu2019roberta}.
Even so, LMs' reasoning processes are notoriously opaque \citep{rudin2019stop, doshi2017towards, lipton2018mythos}, which has spurred significant interest in designing algorithms to automatically explain LM behavior \citep{denil2014extraction, sundararajan2017axiomatic, camburu2018snli, rajani2019explain, luo2021local}.
Most of this work has focused on \textit{rationale extraction}, which explains an LM's output on a given task instance by highlighting the input tokens that most influenced the output \cite{denil2014extraction, sundararajan2017axiomatic, li2016understanding, jin2019towards, lundberg2017unified, chan2022unirex}.

Recent studies have investigated how \textit{machine rationales} outputted by rationale extraction algorithms can be utilized to improve LM decision-making \cite{hase2021can, hartmann-sonntag-2022-survey}.
Among these works, one prevalent paradigm is \textit{explanation regularization} (ER), which aims to improve LM behavior by regularizing the LM to yield machine rationales that align with \textit{human rationales} (Fig. \ref{fig:intro:illustration}) \cite{ross2017right, huang2021exploring, ghaeini2019saliency, zaidan2008modeling, kennedy2020contextualizing, rieger2020interpretations, liu2019incorporating}.
Human rationales can be created by annotating each training instance individually \cite{lin2020triggerner, camburu2018snli, rajani2019explain} or by applying task-level human priors across all training instances \cite{rieger2020interpretations, ross2017right, liu2019incorporating}.

Though prior works primarily evaluate ER models' in-distribution (ID) generalization, the results are mixed, and it is unclear when ER is actually helpful. Furthermore,
out-of-distribution (OOD) generalization is often more crucial in real-world settings \cite{chrysostomou2022an, ruder2021benchmarking}, yet ER's impact on OOD generalization has been underexplored \cite{ross2017right, kennedy2020contextualizing}.
In particular, due to prior works' lack of unified comparison, little is understood about how OOD performance is affected by major ER design choices, such as the rationale alignment criterion, human rationale type (instance-level \versus task-level), number and choice of rationale-annotated instances, and time budget for rationale annotation.

\begin{figure*}[t!]
\vspace{-0.4cm}
    \centering
    \includegraphics[width=0.99\linewidth]{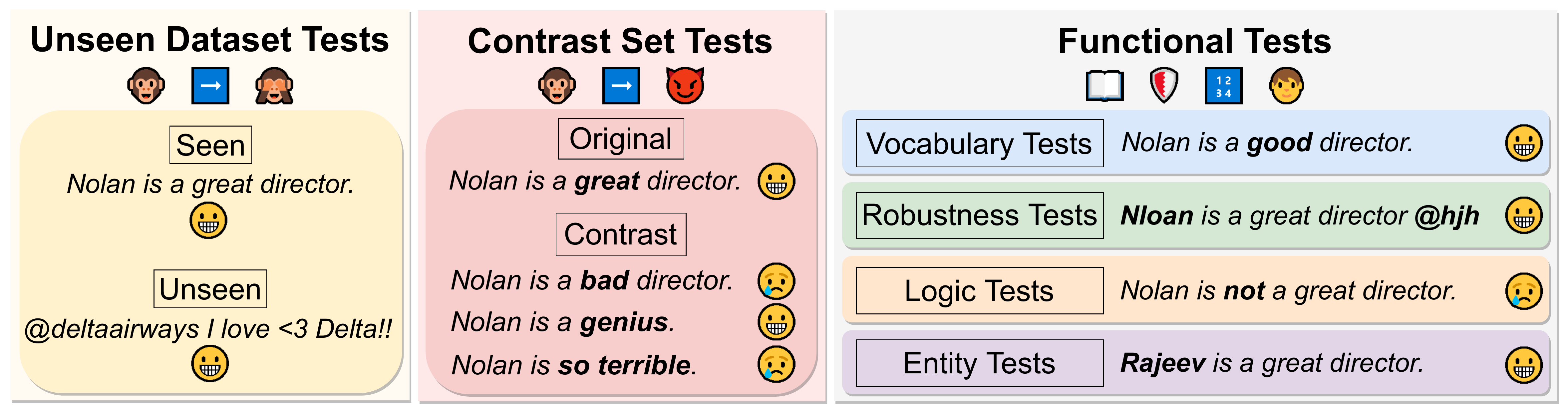}
    \caption{\textbf{\method.} While existing works focus on ER models' in-distribution (ID) generalization, the \method framework is designed to evaluate ER models' out-of-distribution (OOD) generalization with respect to: (A) unseen dataset tests, (B) contrast set tests, and (C) functional tests (\textsection \ref{sec:method}).}
    \label{fig:method:er_test}
\vspace{-0.3cm}
\end{figure*}

In this paper, we propose \ertestemoji \textbf{\method} (Fig. \ref{fig:method:er_test}), a framework for evaluating ER methods' OOD generalization via: (A) \textit{unseen dataset tests}, (B) \textit{contrast set tests}, and (C) \textit{functional tests}.
For (A), \method tests ER models' performance on datasets beyond their training distribution \cite{ross2017right, kennedy2020contextualizing}.
For (B), \method tests ER models' performance on real-world data instances that are semantically perturbed \cite{gardner2020evaluating}.
For (C), \method tests ER models' performance on synthetic data instances created to capture specific linguistic capabilities \cite{ribeiro2020beyond}. 
Using \method, we study four questions: (1) Which rationale alignment \textit{criteria} are most effective? (2) Is ER effective with \textit{task-level} human rationales? (3) How is ER affected by the \textit{number and choice} of rationale-annotated instances? (4) How does ER performance vary with the rationale annotation \textit{time budget}?

For two text classification tasks and six datasets, \method shows that ER has little impact on ID performance but yields large gains on OOD performance, with the best ER criteria being task-dependent (\textsection \ref{sec:exp:rq1}). 
Furthermore, ER can improve OOD performance even with distantly-supervised (\textsection \ref{sec:exp:rq2}) or few (\textsection \ref{sec:exp:rq3}) human rationales.
Finally, we find that rationale annotation is more time-efficient than label annotation, in terms of impact on OOD performance (\textsection \ref{sec:exp:rq4}).
These results from \method help demonstrate ER's utility and establish best practices for using ER effectively.

\section{Explanation Regularization (ER)} 
\label{sec:background}



Given an LM for an NLP task, the goal of ER is to improve LM generalization on the task by pushing the LM's (extractive) machine rationales (\textit{Which input tokens did the LM focus on?}) to align with human rationales (\textit{Which input tokens would humans focus on?}).
The hope is that this inductive bias encourages the LM to solve the task in a manner that follows humans' reasoning process.

Given a set of classes $C$, let $\mathcal{F}$ be an LM for $M$-class text classification, where $|C| = M$.
We assume $\mathcal{F}$ has a BERT-style architecture \citep{devlin2018bert, liu2019roberta}, consisting of a Transformer encoder \citep{vaswani2017attention} followed by a linear layer with softmax classifier.
$\mathcal{F}$ can be used for either sequence or token classification.
Let $\mathbf{x}_i = [x_{i}^{t}]_{t=1}^{n}$ be the $n$-token input sequence (\eg a sentence) for task instance $i$.
For sequence classification, $\mathcal{F}$ predicts a single class for $\mathbf{x}_i$, such that $\mathcal{F}(\mathbf{x}_i) \in \R^{M}$ are the logits for $\mathbf{x}_i$.
In this case, let $y_i = \argmax_{\hspace{0.5mm} c \in C} \mathcal{F}(\mathbf{x}_i)_c$ denote $\mathcal{F}$'s predicted class for $\mathbf{x}_i$.
For token classification, $\mathcal{F}$ predicts a class for each token $x_{i}^{t}$, such that $\mathcal{F}(\mathbf{x}_i) \in \R^{n \times M}$ are the logits for the $n$ tokens in $\mathbf{x}_i$.
In this case, let $y_{i,t} = \argmax_{\hspace{0.5mm} c \in C} \mathcal{F}(\mathbf{x}_{i})_{t,c}$ denote $\mathcal{F}$'s predicted class for $x_{i}^{t}$, while $y_i = [y_{i,t}]_{t=1}^{n}$ collectively denotes all of $\mathcal{F}$'s predicted token classes for $\mathbf{x}_i$.

Given $\mathcal{F}$, $\mathbf{x}_i$, and $y_i$, the goal of rationale extraction is to output feature attribution vector $\mathbf{r}_i = [r_{i}^{t}]_{t=1}^{n}$, where each $0 \leq r_{i}^{t} \leq 1$ is an \textit{importance score} indicating how strongly token $x_{i}^{t}$ influenced $\mathcal{F}$ to predict class $y_i$ \citep{luo2021local}.
In practice, the final machine rationale is obtained by binarizing $\mathbf{r}_i$ via strategies like top-$k\%$ thresholding \cite{deyoung2019eraser, jain2020learning, pruthi2020evaluating}.
However, for convenience, we refer to $\mathbf{r}_i$ as the machine rationale in this work, since the binarized $\mathbf{r}_i$ is not explicitly used in ER. 
Let $\mathcal{G}$ denote a rationale extractor, such that $\mathbf{r}_i = \mathcal{G}(\mathcal{F}, \mathbf{x}_i, y_i)$.


$\mathcal{G}$ can also be used to compute machine rationales with respect to other classes besides $y_i$ (\eg target class $\dot{y}_i$).
Let $\mathbf{\hat{r}}_{i}$ denote the (non-binarized) machine rationale for $\mathbf{x}_i$ with respect to $\dot{y}_i$.
Given $\mathbf{\hat{r}}_{i}$ obtained via $\mathcal{G}$ and $\mathcal{F}$, many works have explored ER, in which $\mathcal{F}$ is regularized such that $\mathbf{\hat{r}}_{i}$ aligns with human rationale $\mathbf{\dot{r}}_{i}$ \cite{zaidan2008modeling, rieger2020interpretations, ross2017right}. 
$\mathbf{\dot{r}}_{i}$ can either be human-annotated for individual instances, or generated via human-annotated lexicons for a given task. 
Typically, $\mathbf{\dot{r}}_{i}$ is a binary vector, where ones and zeros indicate important and unimportant tokens, respectively.

We formalize the ER loss as: $\mathcal{L_{\text{ER}}} = \Phi(\mathbf{\hat{r}}_{i}, \mathbf{\dot{r}}_{i})$,
where $\Phi$ is an ER criterion measuring alignment between $\mathbf{\hat{r}}_{i}$ and $\mathbf{\dot{r}}_{i}$.
Thus, the full learning objective is: $\mathcal{L} = \mathcal{L_{\text{task}}} + \lambda_{\text{ER}} \mathcal{L_{\text{ER}}}$, where $\mathcal{L_{\text{task}}}$ is the task loss (\eg cross-entropy loss) $\lambda_{\text{ER}} \in \R$ is the \textit{ER strength} (\ie loss weight) for $\mathcal{L_{\text{ER}}}$.
Also, as a baseline, let $\mathcal{F}_{\text{No-ER}}$ denote an LM that is trained without ER, such that $\mathcal{L} = \mathcal{L_{\text{task}}}$.

\section{\method} 
\label{sec:method}





Existing works primarily evaluate ER models via ID generalization 
\cite{zaidan2008modeling, lin2020triggerner, rieger2020interpretations, liu2019incorporating, ross2017right, huang2021exploring, ghaeini2019saliency, kennedy2020contextualizing},
though a small number of works have done auxiliary evaluations of OOD generalization \cite{ross2017right, kennedy2020contextualizing, rieger2020interpretations, stacey2022supervising}.
However, these OOD evaluations have been relatively small-scale, only covering a narrow range of OOD generalization aspects.
As a result, little is understood about ER's impact on OOD generalization.
To address this gap, we propose \method (Fig. \ref{fig:method:er_test}), a framework for designing and evaluating ER models' OOD generalization along three dimensions: (1) unseen dataset tests; (2) contrast set tests; and (3) functional tests.

Let $\mathcal{D}$ be an $M$-class text classification dataset, which we call the ID dataset.
Assume $\mathcal{D}$ can be partitioned into training set $\mathcal{D}_{\text{train}}$, development set $\mathcal{D}_{\text{dev}}$, and test set $\mathcal{D}_{\text{test}}$, where $\mathcal{D}_{\text{test}}$ is the ID test set for $\mathcal{D}$.
After training $\mathcal{F}$ on $\mathcal{D}_{\text{train}}$ with ER, we measure $\mathcal{F}$'s ID generalization via task performance on $\mathcal{D}_{\text{test}}$ and $\mathcal{F}$'s OOD generalization via (1)-(3).



\vspace{-0.1cm}
\subsection{Unseen Dataset Tests}
\label{sec:method:ood:unseen_datasets}
\vspace{-0.1cm}

First, we evaluate OOD generalization with respect to unseen dataset tests (Fig. \ref{fig:method:er_test}A).
Besides $\mathcal{D}$, suppose we have datasets $\{ \tilde{\mathcal{D}}^{(1)}, \tilde{\mathcal{D}}^{(2)}, ...  \}$ for the same task as $\mathcal{D}$.
Each $\tilde{\mathcal{D}}^{(i)}$ has its own training/development/test sets and distribution shift from $\mathcal{D}$.
After training $\mathcal{F}$ with ER on $\mathcal{D}_{\text{train}}$ and hyperparameter tuning on $\mathcal{D}_{\text{dev}}$, we measure $\mathcal{F}$'s performance on each OOD test set $\tilde{\mathcal{D}}_{\text{test}}^{(i)}$.
This tests ER's ability to help $\mathcal{F}$ learn general (\ie task-level) knowledge representations that can (zero-shot) transfer across datasets.



\vspace{-0.1cm}
\subsection{Contrast Set Tests}
\label{sec:method:ood:contrast}
\vspace{-0.1cm}

Second, we evaluate OOD generalization with respect to contrast set tests (Fig. \ref{fig:method:er_test}B). 
Dataset annotation artifacts \cite{gururangan-etal-2018-annotation} can cause LMs to learn spurious decision rules that work on the test set but do not capture linguistic abilities that the dataset was designed to assess.
Thus, we test $\mathcal{F}$ on contrast sets \cite{gardner2020evaluating}, which are constructed by manually perturbing the test instances of real-world datasets to express counterfactual meanings.
Contrast set tests help probe the decision boundaries intended by the original dataset's design and if $\mathcal{F}$ has learned undesirable dataset-specific shortcuts.
Given $\tilde{\mathcal{D}}_{\text{test}}^{(i)}$, we can convert $\tilde{\mathcal{D}}_{\text{test}}^{(i)}$ to contrast set $\tilde{\mathcal{C}}_{\text{test}}^{(i)}$ using various types of semantic perturbation, such as inversion (\eg ``\textit{big} dog'' $\rightarrow$ ``\textit{small} dog''), numerical modification (\eg ``\textit{one} dog'' $\rightarrow$ ``\textit{three} dogs''), and entity replacement (\eg ``good \textit{dog}'' $\rightarrow$ ``good \textit{cat}'').
Also, each original instance in $\tilde{\mathcal{D}}_{\text{test}}^{(i)}$ can have multiple corresponding contrast instances in $\tilde{\mathcal{C}}_{\text{test}}^{(i)}$.
Note that it may not be possible to create contrast sets for every instance, in which case these instances are omitted from the contrast set test.

With $\tilde{\mathcal{D}}_{\text{test}}^{(i)}$ and $\tilde{\mathcal{C}}_{\text{test}}^{(i)}$, we evaluate $\mathcal{F}$ using the \textit{contrast consistency} metric.
This is defined as the percentage of instances for which both the original instance and all of its contrast instances are predicted correctly, so higher contrast consistency is better \cite{gardner2020evaluating}.
However, since contrast sets are built from real-world datasets, they provide less flexibility in testing linguistic abilities, as a given perturbation type may not apply to all instances in the dataset.
Note that, unlike adversarial examples \cite{gao2019perturbation}, contrast sets are not conditioned on $\mathcal{F}$ specifically to attack $\mathcal{F}$.

\vspace{-0.1cm}
\subsection{Functional Tests}
\label{sec:method:ood:functional}
\vspace{-0.1cm}
Third, we evaluate OOD generalization with respect to functional tests  (Fig. \ref{fig:method:er_test}C).
Whereas contrast sets are created by perturbing real-world datasets, functional tests evaluate $\mathcal{F}$'s prediction performance on synthetic datasets, which are manually created via templates to assess specific linguistic abilities \cite{ribeiro2020beyond, li-etal-2020-linguistically}.
While contrast set tests focus on semantic abilities, functional tests consider both semantic (\eg perception of word/phrase sentiment, sensitivity to negation) and syntactic (\eg robustness to typos or punctuation addition/removal) abilities.
Therefore, functional tests trade off data realism for evaluation flexibility.
If ER improves $\mathcal{F}$'s functional test performance for a given ability, then ER may be a useful inductive bias for OOD generalization with respect to that ability.
Across all tasks, \method contains four general categories of functional tests: Vocabulary, Robustness, Logic, and Entity \cite{ribeiro2020beyond}. 
See \textsection \ref{sec:app:functional} for more details.

\begin{figure*}[!ht]
\vspace{-0.4cm}
    \centering
    \includegraphics[width=0.98\linewidth]{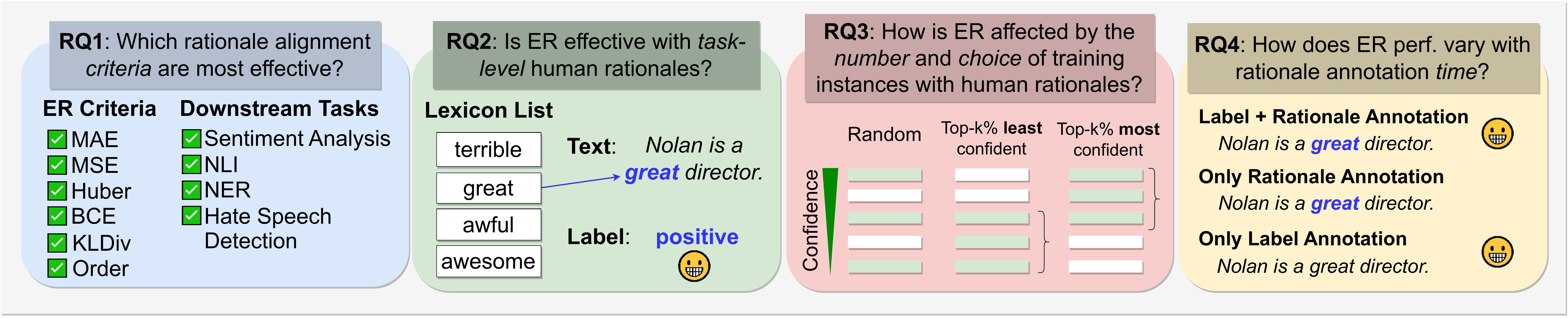}
    \caption{\textbf{\method Research Questions.} 
    To demonstrate \method's utility, we use \method to study four important yet underexplored research questions (RQs). Each RQ considers a different category of ER design choices: rationale alignment criteria (RQ1), human rationale type (RQ2), number/choice of rationale-annotated instances (RQ3), and rationale annotation time (RQ4). With \method, we have a system for identifying ER design choices that are effective for improving OOD generalization (\textsection \ref{sec:exp}).}
    \label{fig:method:er_analysis}
\vspace{-0.5cm}
\end{figure*}

\section{ER Design Choices}
\label{sec:eval}


ER consists of four main components: machine rationale extractor, rationale alignment criterion, human rationale type, and instance selection strategy. With \method, we have a standard tool for evaluating design choices for each component.

\subsection{Machine Rationale Extractors}
\label{sec:eval:rationale_extractors}
We consider three types of machine rationale extractors: \textit{gradient-based}, \textit{attention-based}, and \textit{learned}.
While other rationale extractor types, like \textit{perturbation-based} \citep{li2016understanding}, can also be used, we focus on the first three types since they are relatively less compute-intensive.
We describe these three rationale extractor types below.

\paragraph{Gradient-Based}
Gradient-based rationale extractors compute rationales via the gradient of  logits $\mathcal{F}(\mathbf{x}_i)$ with respect to $\mathbf{x}_i$ \cite{sundararajan2017axiomatic, sanyal-ren-2021-discretized, deeplift}.
In our experiments, we use Input*Gradient (IxG) \cite{denil2014extraction} as a representative gradient-based rationale extractor.
Compared to more expensive gradient-based methods that require multiple backward passes per instance \cite{sundararajan2017axiomatic, lundberg2017unified}, IxG only requires one backward pass per instance.

\paragraph{Attention-Based}
Attention-based rationale extractors compute rationales via the attention weights used by $\mathcal{F}$ to predict $y_i$ \cite{pruthi2020evaluating, stacey2022supervising, saliency, wiegreffe2019attention}.
Following existing Transformer-based works, we consider a variant that uses the attention weights in the final layer of $\mathcal{F}$ \cite{pruthi2020evaluating, stacey2022supervising}. 
In this paper, we simply refer to this variant as Attention.

\paragraph{Learned}
Learned rationale extractors train a model to compute rationales given task input $\mathbf{x}_i$. 
The learned rationale extractor can be trained with respect to faithfulness, plausibility, task performance, and/or knowledge distillation objectives \cite{chan2022unirex, bhat2021self, situ2021learning}.
Given its generality, we consider UNIREX \cite{chan2022unirex} as a representative learned rationale extractor in our experiments.

\subsection{Rationale Alignment Criteria}
\label{sec:eval:er_criteria}


We consider six representative rationale alignment criteria (\ie choices of $\Phi$), described below.

\paragraph{Mean Squared Error (MSE)}
MSE is used in \citet{liu2019incorporating}, \citet{kennedy2020contextualizing}, and \citet{ross2017right}: $\Phi_{\text{MSE}}(\mathbf{\hat{r}}_{i}, \mathbf{\dot{r}}_{i}) = \frac{1}{n} \| \mathbf{\hat{r}}_{i} - \mathbf{\dot{r}}_{i} \|_2^2$.

\paragraph{Mean Absolute Error (MAE)}
MAE is used in \citet{rieger2020interpretations}: $\Phi_{\text{MAE}}(\mathbf{\hat{r}}_{i}, \mathbf{\dot{r}}_{i}) = \frac{1}{n} | \mathbf{\hat{r}}_{i} - \mathbf{\dot{r}}_{i} |$.

\paragraph{Huber Loss}
Huber loss \cite{huber1992robust} is a hybrid of MSE and MAE, but is still unexplored for ER. 
Following the PyTorch library's default settings, our experiments use $\delta=1$ \citep{paszke2019pytorch}.
\begin{align}
\small
\begin{split}
\label{eq:huber}
\Phi_{\text{Huber}}&(\mathbf{\hat{r}}_{i}, \mathbf{\dot{r}}_{i}) \\ =
    &\begin{cases}
        \frac{1}{2} \Phi_{\text{MSE}}(\mathbf{\hat{r}}_{i}, \mathbf{\dot{r}}_{i}), \hspace{-1mm} & \Phi_{\text{MAE}}(\mathbf{\hat{r}}_{i}, \mathbf{\dot{r}}_{i}) < \delta \\
        \delta (\Phi_{\text{MAE}}(\mathbf{\hat{r}}_{i}, \mathbf{\dot{r}}_{i}) - \frac{1}{2} \delta), \hspace{-1mm} &\text{otherwise}
    \end{cases}
\end{split}
\end{align}

\paragraph{Binary Cross Entropy (BCE)}
BCE loss is used in \citet{chan2022unirex} and \citet{chan2021salkg}: $\Phi_{\text{BCE}}(\mathbf{\hat{r}}_{i}, \mathbf{\dot{r}}_{i}) = - \frac{1}{n} \sum_{t=1}^n \dot{r}_i^t\log(\hat{r}_i^t)$.

\paragraph{KL Divergence (KLDiv)}
KLDiv is used by \citet{pruthi2020evaluating}, \citet{chan2022unirex}, and \citet{chan2021salkg}: $\Phi_{\text{KLDiv}}(\mathbf{\hat{r}}_{i}, \mathbf{\dot{r}}_{i}) = \frac{1}{n} \sum_{t = 1}^n \dot{r}_i^t \log(\dot{r}_i^t \hspace{0.5mm} / \hspace{0.5mm} \hat{r}_i^t)$.

\paragraph{Order Loss}
Recall that the human rationale $\mathbf{\dot{r}}_{i}$ labels each token as important (one) or unimportant (zero).
Whereas other criteria generally push important/unimportant tokens' importance scores to be as high/low as possible, order loss \cite{huang2021exploring} relaxes MSE to merely enforce that all important tokens' importance scores are higher than all unimportant tokens' importance scores.
This ranking-based criterion is especially useful if $\mathbf{\dot{r}}_{i}$ is somewhat noisy (\eg if some tokens labeled as important are not actually important).
\vspace{-0.1cm}
\begin{align}
\label{eq:order}
\small
\begin{split}
\Phi_{\text{Order}}(\mathbf{\hat{r}}_{i}, \mathbf{\dot{r}}_{i}) = \sum_{\dot{r}_i^t = 1} \Bigg{(} \min \bigg{(} \frac{\hat{r}_i^t}{\underset{\dot{r}_j^t = 0}{\max} \hspace{1mm} \hat{r}_j^t} - 1, 0 \bigg{)} \Bigg{)}^2
\end{split}
\end{align}

\subsection{Human Rationale Types}
\label{sec:eval:human_rationales}

To construct human rationale $\mathbf{\dot{r}}_{i}$, we consider both \textit{instance-level} and \textit{task-level} human rationales.

\paragraph{Instance-Level Rationales}
Human rationales are often created by annotating each training instance individually \cite{lin2020triggerner, camburu2018snli, rajani2019explain}.
For each instance, humans are asked to mark tokens that support the gold label as important, with the remaining tokens counted as unimportant.
Here, each human rationale is specifically conditioned on the input and gold label for the given instance.
However, instance-level rationales are expensive to obtain, given the high manual effort required per instance.

\paragraph{Task-Level Rationales}
Some works construct distantly-supervised human rationales by applying task-level human priors across all training instances \cite{kennedy2020contextualizing, rieger2020interpretations, ross2017right, liu2019incorporating}.
Given a task-level token lexicon, each instance's rationale is created by marking input tokens present in the lexicon as important and the rest as unimportant, or vice versa.
Here, rationales are not as fine-grained or tailored for the given dataset, but may provide a more general learning signal for solving the task.

\subsection{Instance Selection Strategies}
\label{sec:eval:inst_select}

In real-world applications, it is often infeasible to annotate instance-level human rationales $\mathbf{\dot{r}}_{i}$ for all training instances \cite{chiang2022reexam, kaushik2019learning}.
Besides task-level rationales, another approach for addressing this issue could be to annotate only a subset $\mathcal{S}_{\text{train}} \subset \mathcal{D}_{\text{train}}$ of training instances.
Given a budget of $|\mathcal{S}_{\text{train}}| = \frac{k}{100} |\mathcal{D}_{\text{train}}|$ instances, where $0 < k < 100$, our goal is to select $\mathcal{S}_{\text{train}}$ such that ER with $\mathcal{S}_{\text{train}}$ maximizes $\mathcal{F}$'s task performance.
There exist various ways that the annotation budget can be allocated.
However, for simplicity, we assume that all $|\mathcal{S}_{\text{train}}|$ instances are selected and annotated in a single round, so that ER model training only occurs once.

While instance selection via active learning is well-studied for general classification \cite{schroder2020survey}, this problem has not been explored in ER.
Given non-ER LM $\mathcal{F}_{\text{No-ER}}$, we use \method to compare five active-learning-inspired instance selection strategies.
Note that these are just basic strategies, used to demonstrate a proof of concept for \method's utility.
In practice, one could consider more sophisticated strategies that account for other factors like data diversity.

\paragraph{Random Sampling (Rand)}
constructs $\mathcal{S}_{\text{train}}$ by uniformly sampling $|\mathcal{S}_{\text{train}}|$ instances from $\mathcal{D}$. 

\paragraph{Lowest Confidence (LC)}
selects the $|\mathcal{S}_{\text{train}}|$ instances for which $\mathcal{F}_{\text{No-ER}}$ yields the \textit{lowest} target class confidence probability $\mathcal{F}_{\text{No-ER}}(\dot{y}_i | x_{i})$ \cite{lcsampling}.

\paragraph{Highest Confidence (HC)}
selects the $|\mathcal{S}_{\text{train}}|$ instances for which $\mathcal{F}_{\text{No-ER}}$ yields the \textit{highest} target class confidence probability $\mathcal{F}_{\text{No-ER}}(\dot{y}_i | x_{i})$.
This is the opposite of LC.

\paragraph{Lowest Importance Scores (LIS)} 
Given machine rationale $\mathbf{\hat{r}}_i$ for $\mathcal{F}_{\text{No-ER}}$ and $0 < k' < 100$, let $\mathbf{\hat{r}}_i^{(k')}$ denote a vector of the top-$k\%$ highest importance scores in $\mathbf{\hat{r}}_i$.
With $r_{\mathcal{S}} = (1 / |\mathbf{\hat{r}}_i^{(k')}|) \sum \mathbf{\hat{r}}_i^{(k')}$ as the mean score in $\mathbf{\hat{r}}_i^{(k')}$, LIS selects the $|\mathcal{S}_{\text{train}}|$ instances for which $r_{\mathcal{S}}$ is \textit{lowest}.
This is similar to selecting instances with the highest $\mathbf{\hat{r}}_i$ entropy.

\paragraph{Highest Importance Scores (HIS)} 
Given $r_{\mathcal{S}}$, HIS selects the $|\mathcal{S}_{\text{train}}|$ instances for which $r_{\mathcal{S}}$ is \textit{highest}.
This is the opposite of LIS.





\section{Experiments} 
\label{sec:exp}


Using \method's unified evaluation protocol, we study the effectiveness of ER and various ER design choices with respect to OOD generalization.
In particular, we focus on the following four important research questions (Fig. \ref{fig:method:er_analysis}), which have been underexplored in prior works.

(\textbf{RQ1}) First, \textit{which rationale alignment criteria are most effective for ER?}
Despite rationale alignment criteria being central to ER, little is understood about their influence on ER models' generalization ability (\textsection \ref{sec:exp:rq1}).
(\textbf{RQ2}) Second, compared to instance-level human rationales, \textit{how effective are task-level human rationales for ER?}
Currently, the generalization trade-offs between these two human rationale types is unclear, since existing ER works do not explicitly compare instance-level and task-level human rationales' impact on ER (\textsection \ref{sec:exp:rq2}).
(\textbf{RQ3}) Third, \textit{how is ER affected by the number and choice of training instances with human rationales?}
Sometimes, it is only only feasible to annotate rationales for a small number of training instances, but determining how many and which instances to annotate has been underexplored in ER (\textsection \ref{sec:exp:rq3}).
(\textbf{RQ4}) Fourth, \textit{how is ER affected by the time taken to annotate human rationales?}
Instead of doing ER (\ie with rationale-annotated instances), it is also possible to improve LM generalization by simply providing more label-annotated instances.
To verify the practical utility of ER, we compare the time-efficiency of label and rationale annotation, in terms of their respective impact on model generalization (\textsection \ref{sec:exp:rq4}).


\begin{table*}[ht]
\vspace{-0.5cm}
\centering
\scalebox{0.70}{
\begin{tabular}{ccccccccc}
    \toprule
    \multirow{4}{*}{\makecell[c]{\textbf{Machine} \\ \textbf{Rationale} \\ \textbf{Extractor}}} & \multirow{4}{*}{\makecell[c]{\textbf{Rationale} \\ \textbf{Alignment} \\ \textbf{Criterion}}} & \multicolumn{4}{c}{\textbf{Sentiment Analysis}} & \multicolumn{2}{c}{\textbf{NLI}} \\
    \cmidrule(lr){3-6} \cmidrule(lr){7-8} 
    & & Seen Acc ($\uparrow$)& \multicolumn{3}{c}{Unseen Acc ($\uparrow$)} & Seen F1 ($\uparrow$) & \multicolumn{1}{c}{Unseen F1 ($\uparrow$)} \\
    \cmidrule(lr){3-3} \cmidrule(lr){4-6} \cmidrule(lr){7-7} \cmidrule(lr){8-8} 
    & & SST & Amazon & Yelp & Movies & e-SNLI & MNLI \\
    \midrule
    {-} & No-ER & 94.22~($\pm$0.77) & 90.72 ($\pm$1.36) & 92.07~($\pm$2.66) & 89.83~($\pm$6.79) & 76.18~($\pm$1.28) & 46.15~($\pm$4.38) \\
    \midrule
    \multirow{6}*{IxG}&{MSE}  & 94.29~($\pm$0.05) & 90.58~($\pm$0.77) & 92.17~($\pm$0.64) & 90.00~($\pm$5.63) & \cellcolor{lblue} 78.98~($\pm$1.00) & 54.23~($\pm$2.67) \\&
    {MAE}  & 94.11~($\pm$0.38) & \cellcolor{lblue} 92.02~($\pm$0.25) & \cellcolor{lblue} 94.55~($\pm$0.30) & \cellcolor{lblue} 95.50~($\pm$1.32) & \cellcolor{lblue} 78.77~($\pm$1.01) & \cellcolor{lblue} 52.41~($\pm$4.50) \\&
    {Huber} & 94.19~($\pm$0.19) & 90.43 ($\pm$1.45) & 92.38~($\pm$2.11) & 91.83~($\pm$3.75) & \cellcolor{lblue} 78.99~($\pm$0.81) & 53.97~($\pm$3.11)\\&
    {BCE} & 94.15~($\pm$0.53) & 90.70~($\pm$1.19) & 91.82~($\pm$2.30) & 92.00~($\pm$6.98) & \cellcolor{lblue} 79.07~($\pm$0.83) & \cellcolor{lblue} 53.68~($\pm$4.15) \\&
    {KLDiv} & 94.62~($\pm$0.61) & 91.63~($\pm$0.51) & 93.55~($\pm$1.69) & 93.00~($\pm$2.18) & 73.68~($\pm$4.77) & 46.57~($\pm$1.35)  \\ &
    {Order} & 94.37~($\pm$0.11) & 89.47 ($\pm$2.71) & 87.95~($\pm$6.36) & 84.50~($\pm$10.15) & \cellcolor{lblue} 79.11~($\pm$0.87) & \cellcolor{lblue} 55.26~($\pm$3.56) \\
    \midrule
    \multirow{6}*{Attention}&{MSE} & 94.71~($\pm$0.75) & 91.88~($\pm$0.53) & \cellcolor{lblue} 94.70~($\pm$0.18) & \cellcolor{lblue} 95.83~($\pm$1.15) & 76.04~($\pm$0.43) & 48.60~($\pm$2.55)  \\
    &
    {MAE} & 93.89~($\pm$0.89) & \cellcolor{lblue} 92.18~($\pm$0.59) & \cellcolor{lblue} 94.75~($\pm$0.22) & \cellcolor{lblue} 96.17~($\pm$2.02) & 76.94~($\pm$0.89) & 50.26~($\pm$3.14)  \\&
    {Huber} & 93.92~($\pm$0.94) & 91.93~($\pm$0.75) & \cellcolor{lblue} 94.55~($\pm$0.78) &  \cellcolor{lblue} 96.00~($\pm$0.00) & 76.54~($\pm$1.04) & 50.32~($\pm$2.12)  \\&
    {BCE} & 94.89~($\pm$0.71) & 91.55~($\pm$1.56) & 93.92~($\pm$1.14) & \cellcolor{lblue} 94.83~($\pm$0.58) & 76.17~($\pm$0.65) & 51.28~($\pm$2.06)  \\&
    {KLDiv} & 94.29~($\pm$0.65) & 91.43~($\pm$0.71) &  \cellcolor{lblue} 94.58~($\pm$0.51) & \cellcolor{lblue} 96.67~($\pm$0.76) & \cellcolor{lblue} 77.35~($\pm$0.59) & 49.66~($\pm$2.47) \\ &
    {Order} & 94.92~($\pm$0.35) & 91.45~($\pm$0.00) & 93.90~($\pm$0.92) &  \cellcolor{lblue} 95.50~($\pm$0.71) & 76.76~($\pm$0.12) & 50.66~($\pm$2.55)  \\
    \midrule
    \multirow{6}*{UNIREX} &{MSE} &93.81~($\pm$1.20) & 85.08~($\pm$7.22) & 82.32~($\pm$10.43) & 82.50~($\pm$8.23) & 74.20~($\pm$1.86) & 46.98~($\pm$2.11)\\&
    {MAE}& 94.65~($\pm$0.46) & 90.15~($\pm$1.19) & 92.07~($\pm$1.59) & \cellcolor{lblue} 94.50~($\pm$0.50) & 73.28~($\pm$1.23) & 44.23~($\pm$0.89)\\&
    {Huber}& 94.03~($\pm$0.93) & 87.93~($\pm$4.45) & 89.90~($\pm$4.88) & 88.67~($\pm$6.81) & 73.75~($\pm$2.47) & 46.73~($\pm$2.15)\\&
    {BCE} &94.05~($\pm$0.55) & 86.37~($\pm$3.59) & 83.55~($\pm$6.38) & 74.67~($\pm$14.49) & 69.05~($\pm$1.64) & 40.94~($\pm$1.15)\\&
    {KLDiv} &94.23~($\pm$0.47) & 90.12~($\pm$1.19) & 93.35~($\pm$1.18) & 91.00~($\pm$6.08) & 74.69~($\pm$1.31) & 47.03~($\pm$3.09)\\ &
    {Order} &93.96~($\pm$0.75) & 86.95~($\pm$5.20) & 90.18~($\pm$5.77) & 91.83~($\pm$5.92) & 72.86~($\pm$0.75) & 44.41~($\pm$2.13)\\
    \bottomrule 
\end{tabular}
}
\caption{\small \textbf{RQ1 - Unseen Dataset Tests (\textsection \ref{sec:exp:rq1:unseen}).} We compare various ER \textit{rationale alignment criteria} (as well as the No-ER baseline), with respect to performance on seen (ID) and unseen (OOD) datasets. Performance is measured using accuracy (Acc) for sentiment analysis and macro F1 (F1) for NLI. Numbers highlighted in \colorbox{lblue}{blue} indicate statistically significant improvement over the No-ER baseline ($p < 0.05$).}
\label{tab:exp:rq1:unseen}
\end{table*}

\subsection{Tasks and Datasets}
\label{sec:eval:tasks_datasets}

To demonstrate \method's usage, we consider a diverse set of text classification tasks.
In this paper, we focus on sentiment analysis and natural language inference (NLI) in the main text (\textsection \ref{sec:exp:rq1}-\ref{sec:exp:rq4}), but also present experiments on named entity recognition (NER) (\textsection \ref{sec:app:rq1:ner}) and hate speech detection (\textsection \ref{sec:app:rq1:hate}) in the appendix.

For unseen dataset tests, we focus on the most popular datasets for the given task.
First, for sentiment analysis, we use SST (short movie reviews) \cite{socher2013recursive, carton2020evaluating} as the ID dataset.
As OOD datasets, we use Yelp (restaurant reviews) \citep{zhangCharacterlevelConvolutionalNetworks2015}, Amazon (product reviews) \citep{mcauley2013hidden}, and Movies (long movie reviews) \cite{zaidan2008modeling, deyoung2019eraser}.
Second, for NLI, we use e-SNLI \cite{camburu2018snli, deyoung2019eraser} as the ID dataset and MNLI \cite{williams2017broad} as the OOD dataset.

Since contrast set tests and functional tests are unavailable for the above datasets yet very expensive to construct, we instead use existing contrast set tests and functional tests released by prior works for sentiment analysis and NLI.
Although these existing tests are created from different datasets than the ones mentioned earlier, they can still provide valuable signal for evaluating LMs' OOD generalization.
For sentiment analysis, we use contrast set tests created from the IMDb dataset \cite{gardner2020evaluating} and functional tests created from the Flights dataset \cite{ribeiro2020beyond}.
For NLI, we use contrast set tests created from the Linguistically-Informed Transformations (LIT) dataset \cite{li-etal-2020-linguistically} and functional tests created from the AllenNLP textual entailment (ANLP-NLI) dataset \cite{Gardner2017AllenNLP}.
 





\subsection{Evaluation Metrics}
\label{sec:exp:eval}

\paragraph{Unseen Dataset Tests}
For unseen dataset tests, we evaluate an LM's task performance on unseen datasets using their respective standard metrics.
For sentiment analysis datasets, we measure accuracy \cite{socher2013recursive, zhangCharacterlevelConvolutionalNetworks2015, mcauley2013hidden, zaidan2008modeling}.
For NLI datasets, we measure macro F1 \cite{camburu2018snli, williams2017broad}.

\paragraph{Contrast Set Tests}
For contrast set tests, we primarily evaluate each LM using the contrast consistency metric.
As described in \textsection \ref{sec:method:ood:contrast}, contrast consistency is defined as the percentage of instances for which both the original instance and all of its contrast instances are predicted correctly, so higher contrast consistency is better \cite{gardner2020evaluating}.
In addition, we report the LM's task performance on the original test set and the contrast set.
For sentiment analysis, as described before, task performance metric is measured using accuracy. 
For NLI, task performance metric is measured using accuracy (instead of macro F1), since accuracy is the standard metric for the special LIT dataset used for contrast set tests \cite{li-etal-2020-linguistically}.

\paragraph{Functional Tests}
For all functional tests, we evaluate the LM using the failure rate metric \cite{ribeiro2020beyond}, defined as the percentage of instances predicted incorrectly by the LM.
Since different functional tests' failure rates may have different scales, we use min-max scaling to compute the \textit{normalized failure rate} for each functional test \cite{chan2022unirex}.
Thus, an LM's aggregate functional test performance can be computed as the mean normalized failure rate across all functional tests.

\subsection{Implementation Details}
\label{sec:exp:training}
In all experiments, we use BigBird-Base \cite{zaheer2020big} as the LM architecture, in order to handle input sequences of up to 4096 tokens.
Unless otherwise specified, we use a learning rate of $2\mathrm{e}{-5}$ and an effective batch size of 32.
For all results, we report the mean over three seeds, as well as the standard deviation.
To measure the statistical significance of ER's improvements, we use the unpaired Welch's t-test between each ER model and the baseline No-ER model, with $p < 0.05$.
For our t-test, the alternative hypothesis is that the given ER model's mean performance is greater than the No-ER model's mean performance.
Further implementation details can be found in \textsection \ref{sec:app:training_details}. 

\begin{table*}[ht]
\centering
\scalebox{0.70}{
\begin{tabular}{cccccccc}
    \toprule
    \multirow{4}{*}{\makecell[c]{\textbf{Machine} \\ \textbf{Rationale} \\ \textbf{Extractor}}} & \multirow{4}{*}{\makecell[c]{\textbf{Rationale} \\ \textbf{Alignment} \\ \textbf{Criterion}}} & \multicolumn{3}{c}{\textbf{Sentiment Analysis}} & \multicolumn{3}{c}{\textbf{NLI}} \\
    \cmidrule(lr){3-5} \cmidrule(lr){6-8}
    & & \multicolumn{3}{c}{IMDb} & \multicolumn{3}{c}{LIT} \\
    \cmidrule(lr){3-5} \cmidrule(lr){6-8}
    & & Original Acc ($\uparrow$) & Contrast Acc ($\uparrow$) & Consistency ($\uparrow$) & Original Acc ($\uparrow$) & Contrast Acc ($\uparrow$) & Consistency ($\uparrow$)\\
    \midrule
    {-} & {No-ER} & 88.39~($\pm$2.05) & 85.11~($\pm$2.72) & 73.90~($\pm$4.64) & 46.15~($\pm$4.38) & 43.73~($\pm$2.81) & 16.84~($\pm$3.18)\\
    \midrule
    \multirow{6}*{IxG}& {MSE} & 88.11~($\pm$2.33) & 86.07~($\pm$2.48) & 78.07~($\pm$5.79) & \cellcolor{lblue} 54.23~  ($\pm$2.67) & 51.95~($\pm$1.21) & 16.37~($\pm$1.30)\\
    & {MAE} & 91.12~($\pm$0.59) & 89.82~($\pm$1.20) & \cellcolor{lblue}81.48~($\pm$1.86) & 52.41~($\pm$4.50) & 52.02~($\pm$1.49) & 17.48~($\pm$0.40)\\
    & {Huber} & 89.20~($\pm$1.67) & 86.13~($\pm$1.74) & 75.82~($\pm$3.37) & 53.97~($\pm$3.11) & 52.32~($\pm$1.04) & 16.34~($\pm$0.96)\\
    & {BCE} & 89.55~($\pm$1.42) & 87.30~($\pm$4.03) & 77.25~($\pm$5.30)& 53.68~($\pm$4.15) & 52.37~($\pm$1.42) & 16.90~($\pm$0.63)\\
    & {KLDiv} & 89.82~($\pm$1.71) & 87.91~($\pm$2.14) & 78.28~($\pm$3.50)& 52.39~($\pm$5.59) & 45.07~($\pm$7.32) & 14.91~($\pm$1.92)\\
    & {Order} & 86.00~($\pm$5.27) & 83.40~($\pm$6.16) & 69.74~($\pm$11.50) & 55.26~($\pm$3.56) & 52.78~($\pm$0.74) & 16.20~($\pm$0.54)\\
    \midrule
    \multirow{6}*{Attention} & {MSE} & 91.46~($\pm$0.97) & 89.14~($\pm$1.95) & \cellcolor{lblue}81.01~($\pm$2.19) & 50.56~($\pm$2.24) & 56.42~($\pm$3.17) & 18.87~($\pm$0.95)\\
    & {MAE} & 91.46~($\pm$0.63) & 89.41~($\pm$0.12) & \cellcolor{lblue}81.28~($\pm$0.60) & 52.56~($\pm$1.93) & 52.48~($\pm$1.77) & 17.97~($\pm$0.48)\\
    & {Huber} & 91.33~($\pm$0.24) & 88.66~($\pm$1.17) & \cellcolor{lblue}80.40~($\pm$1.40) & 50.46~($\pm$1.92) & 55.46~($\pm$3.75) & 19.29~($\pm$1.51) \\
    & {BCE} & 91.33~($\pm$1.36) & 89.89~($\pm$1.39) &\cellcolor{lblue}81.76~($\pm$2.67) & 53.43~($\pm$5.12) & 50.62~($\pm$6.28) & 16.72~($\pm$1.77) \\
    & {KLDiv} & 91.39~($\pm$0.82) & 86.81~($\pm$0.83) & 78.48~($\pm$1.43)& 51.78~($\pm$3.87) & 51.91~($\pm$2.41) & 17.83~($\pm$1.80)\\
    & {Order} & 91.70~($\pm$0.14) & 89.34~($\pm$3.19) & \cellcolor{lblue}81.56~($\pm$3.47)& 54.50~($\pm$3.54) & 50.78~($\pm$1.83) & 17.75~($\pm$0.85)\\
    \midrule
    \multirow{6}*{UNIREX} & {MSE} & 77.12~($\pm$17.69) & 70.77~($\pm$14.95) & 48.30~($\pm$32.40) & 46.21~($\pm$2.90) & 55.25~($\pm$2.74) & 17.45~($\pm$2.62)\\
    & {MAE} & 89.96~($\pm$1.25) & 85.86~($\pm$2.08) & 76.30~($\pm$2.40) & 45.18~($\pm$2.88) & 54.53~($\pm$2.79) & 16.50~($\pm$1.47)\\
    & {Huber} & 87.84~($\pm$5.15) & 81.01~($\pm$3.96) & 69.13~($\pm$9.15) & 46.37~($\pm$2.43) & 52.12~($\pm$6.06) & 15.67~($\pm$3.57)\\
    & {BCE} & 67.83~($\pm$18.57) & 61.34~($\pm$14.75) & 29.30~($\pm$33.07)& 39.43~($\pm$2.64) & 47.12~($\pm$5.21) & 11.99~($\pm$1.80)\\
    & {KLDiv} & 88.46~($\pm$0.52) & 82.99~($\pm$1.79) & 71.93~($\pm$2.30)& 48.32~($\pm$3.01) & 49.46~($\pm$5.07) & 14.88~($\pm$2.23)\\
    & {Order} & 85.93~($\pm$8.76) & 80.94~($\pm$11.05) & 67.35~($\pm$19.83) & 43.94~($\pm$2.15) & 54.10~($\pm$1.95) & 16.94~($\pm$0.64)\\

    \bottomrule 
\end{tabular}
}
\caption{\small \textbf{RQ1 - Contrast Set Tests (\textsection \ref{sec:exp:rq1:contrast}).} We compare various ER \textit{rationale alignment criteria} (as well as the No-ER baseline), with respect to performance on both original test sets (ID) and contrast sets (OOD). Performance is reported in terms of original test set accuracy (Original Acc), contrast set accuracy (Contrast Acc), and contrast consistency (Consistency).
Numbers highlighted in \colorbox{lblue}{blue} indicate statistically significant improvement over the No-ER baseline ($p < 0.05$).}
\label{tab:exp:rq1:contrast}
\vspace{-0.0cm}
\end{table*}

\subsection{RQ1: \textit{Which rationale alignment criteria are most effective for ER?}}
\label{sec:exp:rq1}

In RQ1, we use \method to analyze how different rationale alignment criteria impact ER models' ID and OOD generalization ability.

\subsubsection{Setup}
\label{sec:exp:rq1:setup}

In ER, the rationale alignment criterion $\Phi$ is used to push the model's machine rationales to be more similar to human rationales (\textsection \ref{sec:background}).
For RQ1, we consider the three machine rationale extractors (IxG, Attention, UNIREX) and six rationale alignment criteria (MSE, MAE, Huber, BCE, KLDiv, Order) described in \textsection \ref{sec:eval:rationale_extractors}-\ref{sec:eval:er_criteria}.
For all RQ1 settings, we assume instance-level rationales are available for all training instances.
Due to the high computational costs of UNIREX's faithfulness optimization, we only optimize UNIREX rationale extractors for plausibility (equivalent to ER objective) and task performance.
See \textsection \ref{sec:app:training_details} for more details.

Below, we discuss our findings from using \method to explore RQ1.
The RQ1 results obtained via unseen dataset tests (\textsection \ref{sec:exp:rq1:unseen}), contrast set tests (\textsection \ref{sec:exp:rq1:contrast}), and functional tests (\textsection \ref{sec:exp:rq1:functional}) are shown in Table \ref{tab:exp:rq1:unseen}, Table \ref{tab:exp:rq1:contrast}, and Fig. \ref{fig:exp:rq1:functional}, respectively.









\subsubsection{Unseen Dataset Tests}
\label{sec:exp:rq1:unseen}

Table \ref{tab:exp:rq1:unseen} shows the results for unseen dataset tests on RQ1.
First, on both sentiment analysis and NLI, no ER models achieve significantly higher ID (seen dataset) performance than No-ER.
However, when considering OOD (unseen dataset) performance, a considerable number of ER models (\eg IxG+MAE) yield significant improvements over No-ER.
Second, we find that Attention-based and IxG-based ER models generally achieve the best OOD performance on sentiment analysis and NLI, respectively.
On the other hand, although UNIREX-based ER models perform competitively on some datasets, they are noticeably worse on others.
This may be due to our UNIREX extractors not being optimized for faithfulness.
Third, all ER models achieve about the same ID performance, making it hard to distinguish between different ER criteria via ID performance.
Meanwhile, there is much greater variance in OOD performance among ER models, making it easier to identify which criteria are more effective.
For sentiment analysis, ER models using MAE consistently achieve the best overall OOD performance across all rationale extractors.
For NLI, no particular criterion consistently outperforms the others.

Why does MAE perform better than other criteria?
Although ER considers binary human rationales by default, in sentiment analysis, it is unlikely that every token with the same rationale annotation is equally important.
This means the model should have the flexibility to assign different levels of importance to each token.
Nonetheless, MAE, Huber, BCE, and KLDiv heavily penalize any deviation from human rationales, which may push the model to be overconfident in its token importance predictions on OOD data.
Meanwhile, Order's soft ranking objective may be too relaxed, since it is perfectly satisfied even if the important tokens' scores are barely higher than unimportant tokens' scores. 
This may cause the model to have weak learning signal and also fail on OOD data.
This suggests that MAE achieves the best balance of supervision strictness on sentiment analysis.
\subsubsection{Contrast Set Tests}
\label{sec:exp:rq1:contrast}

Table \ref{tab:exp:rq1:contrast} shows the results for contrast set tests on RQ1.
First, for both sentiment analysis and NLI, we observe that IxG-based and Attention-based ER models generally outperform No-ER, in terms of original accuracy, contrast accuracy, and contrast consistency.
In particular, for almost all criteria, the Attention extractor consistently yields high performance, especially on sentiment analysis.
This further demonstrates the benefits of ER on OOD generalization.
Meanwhile, UNIREX-based ER models tend to perform worse than No-ER on both sentiment analysis and NLI.
Again, this may be due to our UNIREX rationale extractors not being optimized for faithfulness.
Second, when comparing criteria, the MAE criterion yields the highest overall performance on sentiment analysis, while no particular criterion dominates on NLI.
This corroborates our findings from the unseen dataset tests.

\begin{figure*}[t!]
\hspace{-0.3cm}
\includegraphics[width=1.04\linewidth]{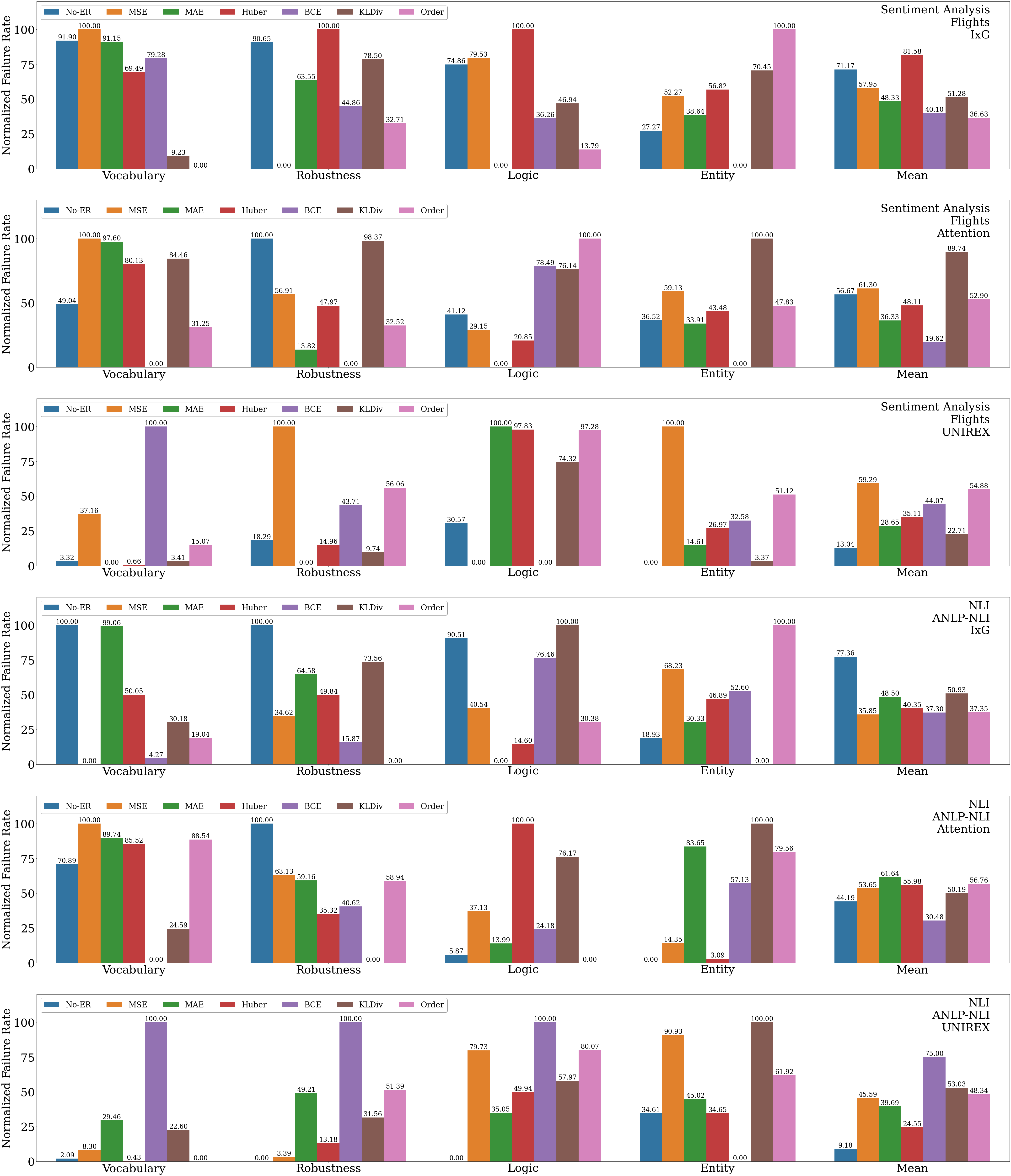}
\vspace{-0.5cm}
\caption{\small \textbf{RQ1 - Functional Tests (\textsection \ref{sec:exp:rq1:functional}).} We compare various ER \textit{rationale alignment criteria} (as well as the No-ER baseline), with respect to performance on a range of functional tests (OOD). Performance is reported in terms of normalized failure rate ($\downarrow$) for four functional test types (Vocabulary, Robustness, Logic, Entity) as well the mean normalized failure rate across all functional tests (Mean).}
\label{fig:exp:rq1:functional}
\end{figure*}




\subsubsection{Functional Tests}
\label{sec:exp:rq1:functional}

Fig. \ref{fig:exp:rq1:functional} shows the results for functional tests on RQ1.
First, we observe that models that perform well on one functional test may not perform well on other functional tests.
This suggests that each functional test evaluates a distinctly different linguistic capability.
Thus, when comparing different models' generalization ability, it is important to consider the mean performance across all four functional tests.
Second, across all functional tests for both tasks, we find that IxG-based ER models consistently achieve lower failure rates than No-ER.
For IxG, all rationale alignment criteria except Huber yield significantly lower failure rates than No-ER, with IxG+Order performing best overall.
However, the results are more mixed for Attention, with Attention+BCE performing best overall.
Meanwhile, UNIREX-based ER models consistently achieve higher failure rates than No-ER.
To some extent, this mirrors our general conclusions from the unseen dataset tests and contrast set tests.
This shows that ER functional test performance can be very sensitive to the choices of both rationale extractor and rationale alignment criteria.

\begin{table*}[ht]
\vspace{-0.2cm}
\centering
\scalebox{0.70}{
\begin{tabular}{ccccccccccc}
    \toprule
     \multirow{4}{*}{\makecell[c]{\textbf{Machine} \\ \textbf{Rationale} \\ \textbf{Extractor}}} & \multirow{4}{*}{\makecell[c]{\textbf{Rationale} \\ \textbf{Alignment} \\ \textbf{Criterion}}} & \multirow{4}{*}{\makecell[c]{\textbf{Human} \\ \textbf{Rationale} \\ \textbf{Type}}} & \multicolumn{4}{c}{\textbf{Sentiment Analysis}}  \\
    \cmidrule(lr){4-7} 
    & & & Seen Acc ($\uparrow$) & \multicolumn{3}{c}{Unseen Acc ($\uparrow$)}  \\
    \cmidrule(lr){4-4} \cmidrule(lr){5-7} 
    & & & SST & Amazon & Yelp & Movies  \\
    \midrule
    - & - & No-ER & 94.22~($\pm$0.77) & 90.72 ($\pm$1.36) & 92.07~($\pm$2.66) & 89.83~($\pm$6.79)  \\
    \midrule
    \multirow{5}{*}{IxG} & \multirow{2}{*}{MAE} & {Instance-Level} &  94.11~($\pm$0.38) & \cellcolor{lblue} 92.02~($\pm$0.25) & \cellcolor{lblue} 94.55~($\pm$0.30) & \cellcolor{lblue} 95.50~($\pm$1.32)  \\
     &  & {Task-Level} & 94.53~($\pm$0.60) & \cellcolor{lblue} 92.02~($\pm$0.45) & \cellcolor{lblue} 94.10~($\pm$0.91) & \cellcolor{lblue} 95.83~($\pm$1.02) \\
     \cmidrule{2-7}
    & \multirow{2}{*}{Huber}  & {Instance-Level} & 94.19~($\pm$0.19) & 90.43 ($\pm$1.45) & 92.38~($\pm$2.11) & 91.83~($\pm$3.75)  \\
    & & {Task-Level} & 93.81~($\pm$0.47) & 91.05~($\pm$1.45) & \cellcolor{lblue} 93.88~($\pm$0.41) & \cellcolor{lblue} 94.00~($\pm$0.40) \\
    \midrule
    \multirow{5}{*}{Attention} & \multirow{2}{*}{MAE}  & {Instance-Level} & 93.89~($\pm$0.89) & \cellcolor{lblue} 92.18~($\pm$0.59) & \cellcolor{lblue} 94.75~($\pm$0.22) & \cellcolor{lblue} 96.17~($\pm$2.02) \\
    & & {Task-Level} & 94.42~($\pm$0.92) & \cellcolor{lblue} 91.73~($\pm$0.51) & \cellcolor{lblue} 94.65 ~($\pm$1.00) & \cellcolor{lblue} 94.33~($\pm$0.62) \\
    \cmidrule{2-7}
    & \multirow{2}{*}{Huber} & {Instance-Level} & 93.92~($\pm$0.94) & 91.93~($\pm$0.75) &  \cellcolor{lblue} 94.55~($\pm$0.78) &  \cellcolor{lblue} 96.00~($\pm$0.00) \\
    & & {Task-Level} & 94.88~($\pm$0.07) & \cellcolor{lblue} 91.90~($\pm$0.10) & \cellcolor{lblue} 94.08 ~($\pm$0.65) & \cellcolor{lblue} 95.83~($\pm$0.84) \\
    \midrule
    \multirow{5}{*}{UNIREX} & \multirow{2}{*}{MAE} & {Instance-Level} & 94.69~($\pm$0.93) & 91.28~($\pm$0.74) & 93.28 ~($\pm$2.16) & \cellcolor{lblue} 94.83~($\pm$2.08) \\
     & & {Task-Level} & 94.65~($\pm$0.46) & 90.15~($\pm$1.19) & 92.07~($\pm$1.59) & 87.33~($\pm$8.36) \\
     \cmidrule{2-7}
    & \multirow{2}{*}{Huber} & {Instance-Level} & 94.03~($\pm$0.93) & 87.93~($\pm$4.45) & 89.90~($\pm$4.88) & 88.67~($\pm$6.81) \\
    & & {Task-Level} & 94.48~($\pm$1.18) & 90.50~($\pm$0.48) & 92.05 ~($\pm$1.00) & 94.33~($\pm$1.24) \\
    \bottomrule 
\end{tabular}
}
\caption{\small \textbf{RQ2 - Unseen Dataset Tests (\textsection \ref{sec:exp:rq2:unseen}).} (\textsection \ref{tab:exp:rq2:unseen}). We compare ER model performance using instance-level rationales versus using \textit{task-level rationales}, with respect to performance on seen (ID) and unseen (OOD) datasets. Performance is measured using accuracy (Acc). Numbers highlighted in \colorbox{lblue}{blue} indicate statistically significant improvement over the No-ER baseline ($p < 0.05$).}
\label{tab:exp:rq2:unseen}
\end{table*}

\subsection{RQ2: \textit{How effective are task-level human rationales for ER?}}
\label{sec:exp:rq2}

In RQ2, we use \method to compare the effectiveness of instance-level and task-level human rationales when used for ER on the same task.



\subsubsection{Setup}
\label{sec:exp:rq2:setup}
Due to computational constraints, we only consider a subset of the settings used in RQ1.
First, we focus on the sentiment analysis task.
Second, although we consider all three extractors (IxG, Attention, UNIREX), we only consider the MAE and Huber criteria, since they yielded the best task performance on the ID development set.
Third, to generate task-level rationales, we use the AFINN \cite{afinn} and SenticNet \cite{senticnet} lexicons.
See \textsection \ref{sec:app:rq2} for more details.

\begin{figure*}[ht!]
\hspace{-0.3cm}
\includegraphics[width=1.04\linewidth]{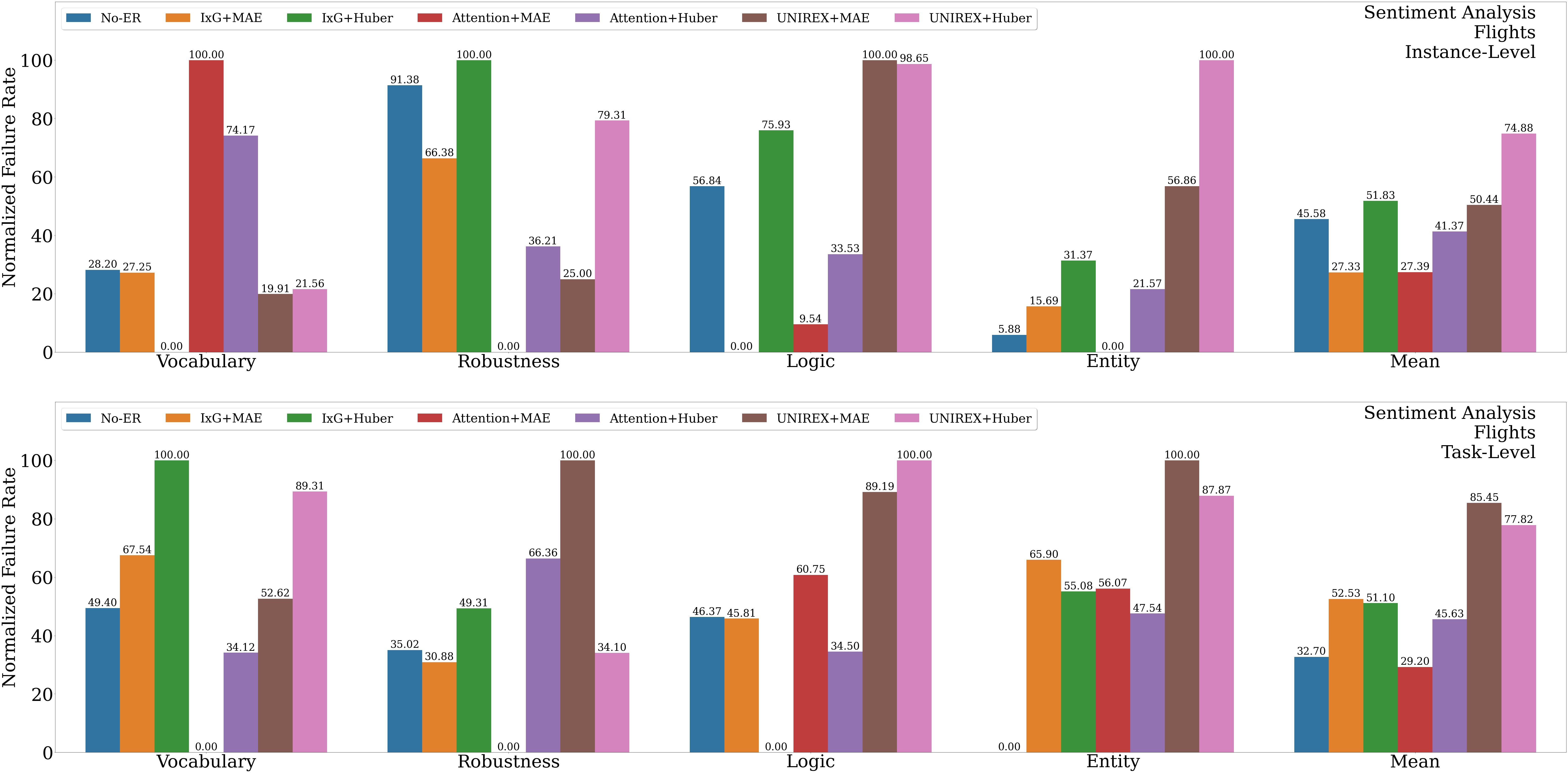}
\vspace{-0.5cm}
\caption{\small \textbf{RQ2 - Functional Tests (\textsection \ref{sec:exp:rq2:functional}).} We compare ER model performance using instance-level rationales versus using \textit{task-level rationales}, with respect to performance on a range of functional tests (OOD). Performance is reported in terms of normalized failure rate ($\downarrow$) for four functional test types (Vocabulary, Robustness, Logic, Entity) as well the mean normalized failure rate across all functional tests (Mean).}
\label{fig:exp:rq2:functional}
\end{figure*}

Below, we discuss our findings from using \method to explore RQ2.
The RQ2 results obtained via unseen dataset tests (\textsection \ref{sec:exp:rq2:unseen}), contrast set tests (\textsection \ref{sec:exp:rq2:contrast}), and functional tests (\textsection \ref{sec:exp:rq2:functional}) are shown in Table \ref{tab:exp:rq2:unseen}, Table \ref{tab:exp:rq2:contrast}, and Fig. \ref{fig:exp:rq2:functional}, respectively.


\begin{table*}[ht!]
\centering
\scalebox{0.70}{
\begin{tabular}{cccccccccc}
    \toprule
    \multirow{4}{*}{\makecell[c]{\textbf{Machine} \\ \textbf{Rationale} \\ \textbf{Extractor}}} & \multirow{4}{*}{\makecell[c]{\textbf{Rationale} \\ \textbf{Alignment} \\ \textbf{Criterion}}} &
    \multirow{4}{*}{\makecell[c]{\textbf{Human} \\ \textbf{Rationale} \\ \textbf{Type}}} & \multicolumn{3}{c}{\textbf{Sentiment Analysis}} \\
    \cmidrule(lr){4-6}
    & & & \multicolumn{3}{c}{IMDb} \\
    \cmidrule(lr){4-6}
    & & & Original Acc ($\uparrow$) & Contrast Acc ($\uparrow$) & Consistency ($\uparrow$) \\
    \midrule
    \multirow{4}*{IxG} & \multirow{2}*{MAE} & {Instance-Level} & 91.12~($\pm$0.59) & 89.82~($\pm$1.20) & \cellcolor{lblue}81.48~($\pm$1.86) \\
    &  & {Task-Level} &  91.46~($\pm$0.72) &89.82~($\pm$2.46) &\cellcolor{lblue}83.47~($\pm$0.47) \\
    \cmidrule{2-6}
    & \multirow{2}*{Huber} & {Instance-Level} & 89.20~($\pm$1.67) & 86.13~($\pm$1.74) & 75.82~($\pm$3.37) \\
    & & {Task-Level} &  90.64~($\pm$1.25) &89.82~($\pm$2.46) & \cellcolor{lblue}79.58~($\pm$1.72) \\
    \midrule
    \multirow{4}*{Attention} & \multirow{2}*{MAE} & {Instance-Level} & 91.46~($\pm$0.63) & 89.41~($\pm$0.12) & \cellcolor{lblue}81.28~($\pm$0.60) \\
    &  & {Task-Level} &  91.19~($\pm$0.20) & 91.94~($\pm$0.43) & \cellcolor{lblue}83.47~($\pm$0.47)\\
    \cmidrule{2-6}
    & \multirow{2}*{Huber} & {Instance-Level} & 91.33~($\pm$0.24) & 88.66~($\pm$1.17) & \cellcolor{lblue}80.40~($\pm$1.40) \\
    & & {Task-Level} &  91.12~($\pm$0.12) & 91.94~($\pm$0.43) & \cellcolor{lblue}79.58~($\pm$2.07)\\
    \midrule
    \multirow{4}*{UNIREX} & \multirow{2}*{MAE} & {Instance-Level} & 89.96~($\pm$1.25) & 85.86~($\pm$2.08) & 76.30~($\pm$2.40) \\
    &  & {Task-Level} &  84.77~($\pm$9.03)&  77.94~($\pm$13.45) &  62.98 ~($\pm$22.43)\\
    \cmidrule{2-6}
     & \multirow{2}*{Huber} & {Instance-Level} & 87.84~($\pm$5.15) & 81.01~($\pm$3.96) & 69.13~($\pm$9.15)\\
    & & {Task-Level} &  89.89~($\pm$0.52) & 84.02~($\pm$1.55) & 74.25~($\pm$2.17) \\
    \bottomrule 
\end{tabular}
}
\caption{\small \textbf{RQ2 - Contrast Set Tests (\textsection \ref{sec:exp:rq2:contrast}).} We compare ER model performance using instance-level rationales versus using \textit{task-level rationales}, with respect to performance on both original test sets (ID) and contrast sets (OOD). Performance is reported in terms of original test set accuracy (Original Acc), contrast set accuracy (Contrast Acc), and contrast consistency (Consistency). Numbers highlighted in \colorbox{lblue}{blue} indicate statistically significant improvement over the No-ER baseline ($p < 0.05$).}
\label{tab:exp:rq2:contrast}
\vspace{0.2cm}
\end{table*}

\subsubsection{Unseen Dataset Tests}
\label{sec:exp:rq2:unseen}
Table \ref{tab:exp:rq2:unseen} shows the results for unseen dataset tests on RQ2. 
Note that the results for instance-level rationales are copied from RQ1's unseen dataset test results in Table \ref{tab:exp:rq1:unseen}.
For both instance-level and task-level rationales, all ER models perform similarly to No-ER on the seen dataset (SST).
Meanwhile, for both instance-level and task-level rationales, most ER models significantly outperform No-ER on the unseen datasets (Amazon, Yelp, Movies).
In particular, task-level rationales yield notable gains on the Yelp and Movie datasets, sometimes even beating their instance-level counterparts on the same extractors.
We believe task-level rationales' advantage in some settings is due to their lexicon being task-specific (\ie for sentiment analysis) and dataset-agnostic.
In other words, task-level rationales may contain more general knowledge (\ie sentiment-related terms) that is also applicable to unseen datasets, whereas the instance-based rationales contain more SST-specific knowledge.



\subsubsection{Contrast Set Tests}
\label{sec:exp:rq2:contrast}

Table \ref{tab:exp:rq2:contrast} shows the results for contrast set tests on RQ2.
Note that the results for instance-level rationales are copied from RQ1's contrast set test results in Table \ref{tab:exp:rq1:contrast}.
First, for both instance-level and task-level rationales, we observe that IxG-based and Attention-based ER models generally outperform No-ER on all three metrics. 
Second, for some settings, we find that task-level rationales can even yield higher contrast accuracy and contrast consistency than instance-level rationales, while achieving similar original accuracy.
For both IxG-based and Attention-based ER models, task-level rationales often outperform instance-level rationales in contrast accuracy and contrast consistency.
These results suggest task-level rationales can serve as an effective yet annotation-efficient substitute for instance-level rationales.




\begin{figure*}
\vspace{0.2cm}
    \centering
    \includegraphics[width=0.98\linewidth]{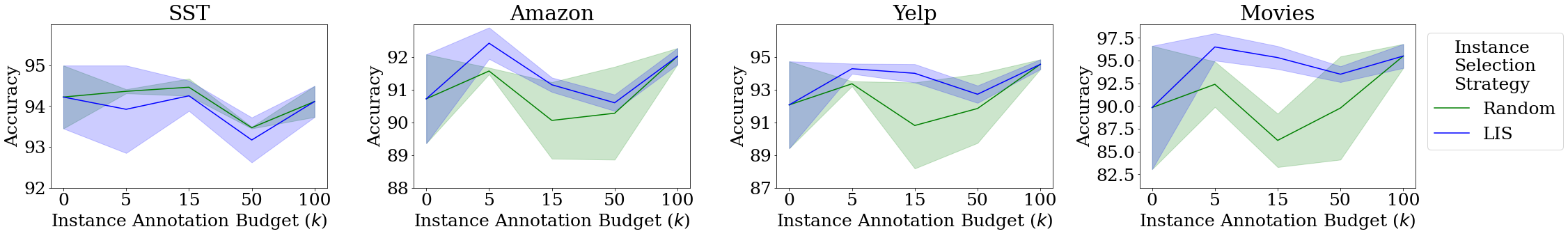}
    \vspace{-0.2cm}
    \caption{\small \textbf{RQ3 - Unseen Dataset Tests (\textsection \ref{sec:exp:rq3:unseen}).} 
    For the strongest \textit{instance selection strategy} (LIS) and key baselines (No-ER, Random), we plot task performance (Accuracy) as a function of instance annotation budget ($k$). Table \ref{tab:exp:rq3:unseen} presents more comprehensive results comparing No-ER, Random, and LIS to other instance selection strategies (LC, HC, HIS). 
    }
    \label{fig:exp:rq3:unseen}
    \vspace{0.2cm}
\end{figure*}

\begin{table*}[ht!]
\centering
\scalebox{0.70}{
\begin{tabular}{cccccccccc}
    \toprule
    \multirow{4}{*}{\makecell[c]{\textbf{Instance} \\ \textbf{Annotation} \\ \textbf{Budget} ($k$)}} & \multirow{4}{*}{\makecell[c]{\textbf{Instance} \\ \textbf{Selection} \\ \textbf{Strategy}}} & \multicolumn{4}{c}{\textbf{Sentiment Analysis}}  \\
    \cmidrule(lr){3-6} 
    & & Seen Acc ($\uparrow$) & \multicolumn{3}{c}{Unseen Acc ($\uparrow$)}  \\
    \cmidrule(lr){3-3} \cmidrule(lr){4-6}
    & & SST & Amazon & Yelp & Movies  \\
    \midrule
    0$\%$ & - & 94.22~($\pm$0.77) & 90.72 ($\pm$1.36) & 92.07~($\pm$2.66) & 89.83~($\pm$6.79)  \\
    100$\%$ & - & 94.11~($\pm$0.38) & \colorbox{lblue}{92.02~($\pm$0.25)} &\colorbox{lblue}{ 94.55~($\pm$0.30) }&\colorbox{lblue}{95.50~($\pm$1.32)}  \\
    \midrule
    \multirow{5}{*}{5$\%$} & {Random} & 94.36~($\pm$0.05) & 91.57~($\pm$0.10) & 93.36~($\pm$0.15) & 92.39~($\pm$2.50) \\
    & {LC} & 93.14~($\pm$1.97) & 90.72~($\pm$0.43) & 93.50~($\pm$0.53) & \colorbox{lblue}{93.17~($\pm$1.26)} \\
    & {HC} & 94.32~($\pm$0.42)& \colorbox{lblue}{91.57~($\pm$0.19)} & 93.03~($\pm$0.81) & 91.33~($\pm$3.09) \\
    & {LIS} & 93.92~($\pm$1.07) & \colorbox{lblue}{92.42~($\pm$0.48)}& \colorbox{lblue}{94.28~($\pm$0.31)} & \colorbox{lblue}{96.50~($\pm$1.50)} \\
    & {HIS} & 93.94~($\pm$0.83) & 90.58~($\pm$0.95) & 91.47~($\pm$2.37) & 92.00~($\pm$4.58) \\
    \midrule
    \multirow{5}{*}{15$\%$} & {Random} & 94.46~($\pm$0.21) & \colorbox{lblue}{90.06~($\pm$1.17)} & 90.81~($\pm$2.63) & 86.22~($\pm$2.94) \\
    & {LC} & 93.48~($\pm$0.80) & 90.12~($\pm$2.66) & 90.90~($\pm$5.30) & 83.67~($\pm$14.02) \\
    & {HC} & 94.39~($\pm$0.27) & 90.38~($\pm$1.12) & \colorbox{lblue}{93.48~($\pm$0.64)} & \colorbox{lblue}{91.33~($\pm$5.11)}\\
    & {LIS} & 94.25~($\pm$0.37) & 91.15~($\pm$0.22) & 94.00~($\pm$0.56) & \colorbox{lblue}{95.33~($\pm$1.26)} \\
    & {HIS} & 94.47~($\pm$0.22) & 91.13~($\pm$0.60) & 92.67~($\pm$0.98) & 93.50~($\pm$3.12) \\
    \midrule
    \multirow{5}{*}{50$\%$} & {Random} & 93.47~($\pm$0.02) & 90.28~($\pm$1.42) & 91.85~($\pm$2.11) & 89.78~($\pm$5.68) \\
    & {LC} & 87.07~($\pm$5.15) & 78.82~($\pm$20.68) & 77.73~($\pm$26.53) & 76.67~($\pm$19.08) \\
    & {HC} & 92.93~($\pm$0.17) & 92.15~($\pm$0.36) & 94.48~($\pm$0.94) & 91.00~($\pm$6.50) \\
    & {LIS} & 93.17~($\pm$0.55) & 90.60~($\pm$0.25) & 92.72~($\pm$0.53) & 93.50~($\pm$0.87) \\
    & {HIS} & 94.23~($\pm$0.65) & 88.85~($\pm$2.67) & 91.47~($\pm$1.47) & 93.67~($\pm$1.89) \\
    \bottomrule 
\end{tabular}
}
\caption{\small \textbf{RQ3 - Unseen Dataset Tests (\textsection \ref{sec:exp:rq3:unseen}).} 
We compare ER model performance for five \textit{instance selection strategies} across different instance annotation budgets ($k$\% of training data), with respect to performance on seen (ID) and unseen (OOD) datasets. Performance is measured using accuracy (Acc).
Numbers highlighted in \colorbox{lblue}{blue} indicate statistically significant improvement over the No-ER baseline ($p < 0.05$).
}
\label{tab:exp:rq3:unseen}
\end{table*}


\subsubsection{Functional Tests}
\label{sec:exp:rq2:functional}

Fig. \ref{fig:exp:rq2:functional} shows the functional test results for RQ2.
First, for instance-level rationales, we see that multiple ER models (\ie IxG+MAE, Attention+MAE, Attention+Huber) yield lower failure rates than No-ER, although ER models perform especially poorly on entity tests.
In particular, IxG+MAE achieves the lowest mean failure rate for instance-level rationales, with Attention+MAE coming at a close second.
This supports our earlier finding that MAE is a generally effective rationale alignment criterion.
Second, for task-level rationales, we find that only Attention+MAE achieves the lowest failure rate but is the only ER model that achieves a lower mean failure rate than No-ER.
Again, ER models perform especially poorly on entity tests.
Although task-level rationales can be useful in certain situations (\eg unseen dataset tests and contrast tests), these results show the limitations of using task-level rationales instead of instance-level rationales.
We hypothesize that relying on such task-level lexicons may sometimes hinder ER models from generalizing to harder instances that require dataset-specific knowledge, which can only be obtained via instance-level rationale annotations.



\subsection{RQ3: \textit{How is ER affected by the number and choice of training instances with human rationales?}}
\label{sec:exp:rq3}

In RQ1 and RQ2, we considered the ER setting where human rationales (whether instance-level or task-level) are available for all training instances.
However, if rationale annotation resources are limited and task-level rationales are not feasible, we need a way to determine which training instances should be prioritized for instance-level rationale annotation.
Using \method, RQ3 explores the impact of different annotation budgets (\ie \textit{number} of training instances) and instance selection methods (\ie \textit{choice} of training instances) on ER model generalization.

\subsubsection{Setup}
\label{sec:exp:rq3:setup}

Following \textsection \ref{sec:eval:inst_select}, we consider budgets of $k = \{ 5, 15, 50 \}$, where $k\%$ of training instances are annotated with human rationales.
For instance selection, we consider the five strategies (Random, LC, HC, LIS, HIS) described in \textsection \ref{sec:eval:inst_select}.
As reference points, we also include the No-ER model ($k=0$) and the fully-supervised ER model ($k=100$).
Due to computational constraints, we limit our RQ3 experiments to sentiment analysis and IxG+MAE, which yielded the highest development ID performance for RQ1 (\textsection \ref{sec:app:training_details}).
See \textsection \ref{sec:app:rq3} for more details.

\begin{figure*}[t!]
\hspace{-0.3cm}
\includegraphics[width=1.04\linewidth]{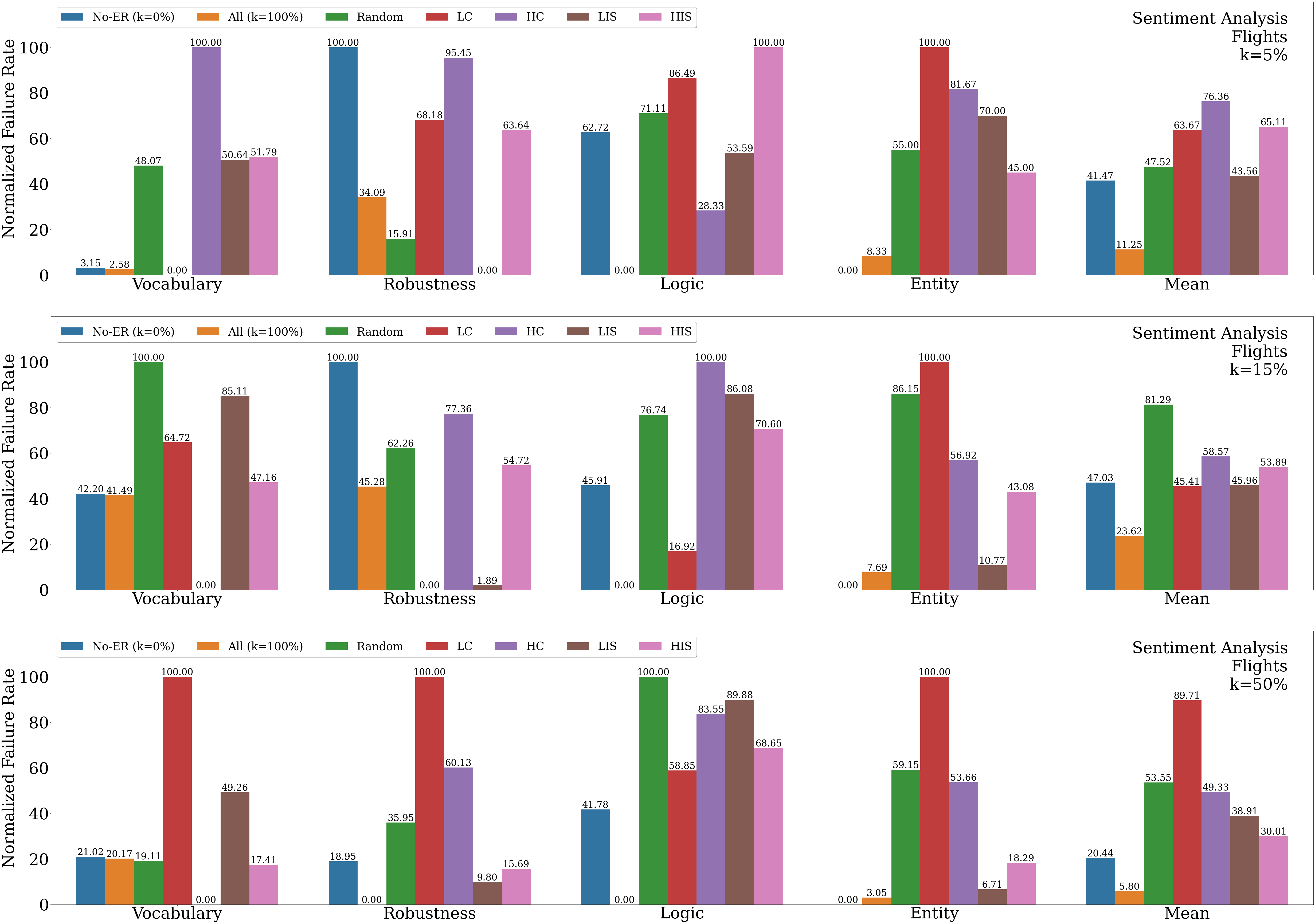}
\vspace{-0.5cm}
\caption{\small \textbf{RQ3 - Functional Tests (\textsection \ref{sec:exp:rq3:functional}).} We compare ER model performance for five \textit{instance selection strategies} across different instance annotation budgets ($k$\% of training data), with respect to performance on a range of functional tests (OOD). Performance is reported in terms of normalized failure rate ($\downarrow$) for four functional test types (Vocabulary, Robustness, Logic, Entity) as well the mean normalized failure rate across all functional tests (Mean).}
\label{fig:exp:rq3:functional}
\end{figure*}

Below, we discuss our findings from using \method to explore RQ3.
The RQ3 results obtained via unseen dataset tests (\textsection \ref{sec:exp:rq3:unseen}), contrast set tests (\textsection \ref{sec:exp:rq3:contrast}), and functional tests (\textsection \ref{sec:exp:rq3:functional}) are shown in Table \ref{tab:exp:rq3:unseen} (as well as Fig. \ref{fig:exp:rq3:unseen}), Table \ref{tab:exp:rq3:contrast}, and Fig. \ref{fig:exp:rq3:functional}, respectively.

\begin{table*}[ht]
\centering
\scalebox{0.70}{
\begin{tabular}{ccccc}
  \toprule
    \multirow{4}{*}{\makecell[c]{\textbf{Instance} \\ \textbf{Annotation} \\ \textbf{Budget} ($k$)}} & \multirow{4}{*}{\makecell[c]{\textbf{Instance} \\ \textbf{Selection} \\ \textbf{Strategy}}} & \multicolumn{3}{c}{\textbf{Sentiment Analysis}} \\
    \cmidrule(lr){3-5}
    & & \multicolumn{3}{c}{IMDb} \\
    \cmidrule(lr){3-5}
    & & Original Acc ($\uparrow$) & Contrast Acc ($\uparrow$) & Consistency ($\uparrow$) \\
    \midrule
    0$\%$ & - & 88.39~($\pm$2.05) & 85.11~($\pm$2.72) & 73.90~($\pm$4.64) \\
    100$\%$ & - & 91.12~($\pm$0.59) & 89.82~($\pm$1.20) & \cellcolor{lblue}81.48~($\pm$1.86) \\
    \midrule
    \multirow{5}{*}{5$\%$} & {Random} &  90.03~($\pm$1.71) &85.63~($\pm$1.76) & 76.05~($\pm$3.37) \\
    & {LC} &  90.71~($\pm$0.24) &89.07~($\pm$0.31)  &\cellcolor{lblue}80.26~($\pm$0.30) \\
    & {HC} & 90.98~($\pm$1.02)  &88.66~($\pm$0.12) &\cellcolor{lblue}80.12~($\pm$1.08) \\
    & {LIS} & 91.73~($\pm$0.12)&89.34~($\pm$0.89)&\cellcolor{lblue}81.56~($\pm$1.25) \\
     & {HIS} & 88.32~($\pm$4.14)&84.77~($\pm$5.93)&73.57~($\pm$10.11) \\
     \midrule
   \multirow{5}{*}{15$\%$} & {Random} &  89.41~($\pm$2.51) & 86.32~($\pm$2.84) & 76.18~($\pm$5.40)\\
    & {LC} &  87.43~($\pm$6.08) & 87.02~($\pm$8.64) & 74.81~($\pm$14.77)\\
    & {HC} &  90.44~($\pm$1.33) & 86.75~($\pm$1.89) & 77.60~($\pm$3.13)\\
    & {LIS} &  91.46~($\pm$0.66) & 88.93~($\pm$0.20) & \cellcolor{lblue}80.87~($\pm$0.43)\\  
    & {HIS} &  90.51~($\pm$1.17) & 87.23~($\pm$2.56) & 78.35~($\pm$3.80)\\
    \midrule
    \multirow{5}{*}{50$\%$} & {Random} &  87.86~($\pm$3.15)&  84.29~($\pm$4.35) &  72.56 ~($\pm$7.11)\\
    & {LC} &  88.18~($\pm$2.96) & 87.70~($\pm$1.98) & 76.30~($\pm$5.10) \\
    & {HC} & 87.70~($\pm$3.63) & 84.22~($\pm$3.50) & 72.47~($\pm$7.11) \\
  & {LIS} &  89.69~($\pm$1.36) & 86.89~($\pm$1.95) & 77.05~($\pm$3.35) \\
    & {HIS} &  88.87~($\pm$1.03) & 84.02~($\pm$3.19) & 73.36~($\pm$3.36) \\
    \bottomrule 
\end{tabular}}
\caption{\small \textbf{RQ3 - Contrast Set Tests (\textsection \ref{sec:exp:rq3:contrast}).} We compare ER model performance for five \textit{instance selection strategies} across different instance annotation budgets ($k$\% of training data), with respect to performance on both original test sets (ID) and contrast sets (OOD). Performance is reported in terms of original test set accuracy (Original Acc), contrast set accuracy (Contrast Acc), and contrast consistency (Consistency). Numbers highlighted in \colorbox{lblue}{blue} indicate statistically significant improvement over the No-ER baseline ($p < 0.05$). }
\label{tab:exp:rq3:contrast}
\vspace{-0.4cm}
\end{table*}

\subsubsection{Unseen Dataset Tests}
\label{sec:exp:rq3:unseen}


Table \ref{tab:exp:rq3:unseen} and \ref{fig:exp:rq3:unseen} present the results for unseen dataset tests on RQ3.
First, like we saw in RQ1 and RQ2, almost all compared models have similar ID (seen dataset) performance.
Thus, we continue to focus on comparing models with respect to OOD (unseen dataset) performance.

Second, among the different instance selection strategies, we see that LIS achieves the best overall OOD performance, across all datasets and instance annotation budgets.
This demonstrates that feature importance scores can provide useful signal for ranking instances to annotate.
Interestingly, although Random generally does not perform well, we find that Random does not always yield the worst performance, with LC typically performing worse.
In particular, for $k=50\%$, LC achieves much lower accuracy (with high variance) than other strategies do.
This indicates that label confidence scores do not provide reliable signal for ranking instances.

Third, as expected, the ER model trained on all rationale annotations ($k=100\%$) generally outperforms both No-ER ($k=0\%$) and ER models with other budgets ($k=\{5\%, 15\%, 50\%\}$).
However, we counterintuitively find that $k=5\%$ tends to outperform both $k=15\%$ and $k=50\%$.
In some cases (\ie LIS on Movies), $k=5\%$ even slightly outperforms $k=100\%$.
This suggests that, despite its success in some settings, feature importance scores alone cannot provide sufficient signal for ranking instances to annotate.
We leave further investigation of other feature importance algorithms and other instance ranking strategies to future work.




\subsubsection{Contrast Set Tests}
\label{sec:exp:rq3:contrast}

Table \ref{tab:exp:rq3:contrast} presents the results for contrast set tests on RQ3.
First, among the different instance selection strategies, we see that LIS performs best overall on the three metrics, across all instance annotation budgets.
As expected, Random yields the worst overall performance, since it does not rank instances in any intelligent way.
Furthermore, after Random, HIS yields the second-worst overall performance, since it provides the opposite ranking as LIS.
This demonstrates that feature importance scores can provide useful signal for ranking instances to annotate.
On the other hand, although LC and HC also perform well for $k=5\%$, they perform noticeably worse for $k=15\%$ and/or $k=50\%$.
In particular, LC beats HC for $k=50\%$, while HC beats LC for $k=15\%$.
This inconsistent signal shows that label confidence scores are not effective for ranking instances to annotate.
Second, as expected, the ER model trained on all rationale annotations ($k=100\%$) generally outperforms both No-ER ($k=0\%$) and ER models with other budgets ($k=\{5\%, 15\%, 50\%\}$).
However, we counterintuitively find that $k=5\%$ tends to outperform both $k=15\%$ and $k=50\%$.
In some cases (\ie LIS), $k=5\%$ even slightly outperforms $k=100\%$.
This suggests that, despite its success in some settings, feature importance scores alone cannot provide sufficient signal for ranking instances to annotate.
Overall, our conclusions from the contrast set tests are in line with those from the unseen dataset tests.
We leave further investigation of other feature importance algorithms and other instance ranking strategies to future work.


\subsubsection{Functional Tests}
\label{sec:exp:rq3:functional}

Fig. \ref{fig:exp:rq3:functional} presents the results for functional tests on RQ3.
First, we find that the $k=100\%$ ER model consistently outperforms No-ER and other ER models.
Yet, for all rationale annotation budgets except $k=100\%$, the ER model fails to outperform No-ER.
This shows that having sufficiently abundant rationale annotations is important for training an ER model that captures the linguistic capabilities evaluated in these functional tests.
Second, for budgets $k=\{5\%, 15\%, 50\%\}$, no instance ranking strategy consistently beats other strategies, even when considering the Random strategy.
This shows that LIS may not necessarily provide useful signal for ranking instances to annotate.
Also, this further supports the notion that, for functional tests, no instance ranking strategy is strong enough to overcome insufficient rationale annotations.

\subsection{RQ4: \textit{How is ER affected by the time taken to annotate human rationales?}}
\label{sec:exp:rq4}

\begin{figure*}
    \centering
    \includegraphics[width=0.97\linewidth]{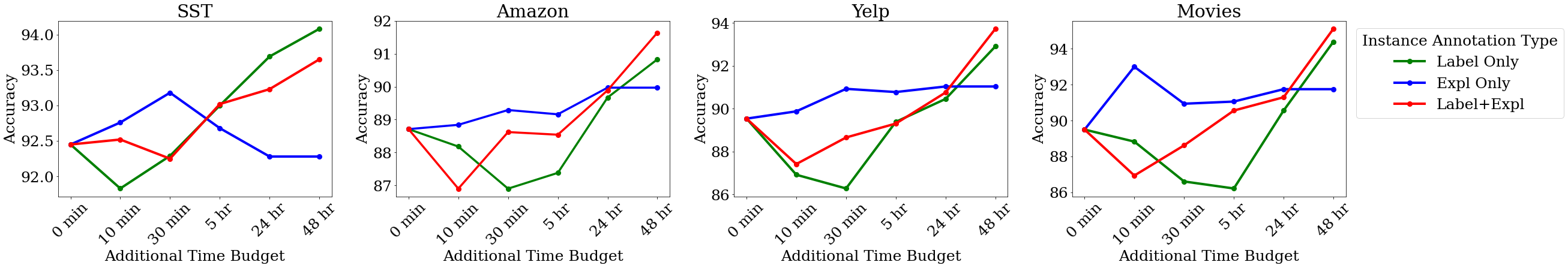}
    \vspace{-0.1cm}
    \caption{\small \textbf{RQ4 - Unseen Dataset Tests (\textsection \ref{sec:exp:rq4:unseen}).}
    We compare ER model performance for three \textit{instance annotation types} across different time budgets, with respect to performance on seen (ID) and unseen (OOD) datasets. For each instance annotation type (Label Only, Expl Only, Label+Expl), we plot task performance (Accuracy) as a function of additional time budget for annotation.
    Note that the model trained with 0 min additional time budget corresponds to the No-ER baseline.
    }
    \label{fig:exp:rq4:unseen}
\end{figure*}

Previously, we considered ER models trained on instance-level (RQ1, RQ3) and task-level (RQ2) human rationale annotations.
However, instead of doing ER, it is also possible to improve LM generalization by simply providing more label-annotated instances.
For ER to be practical, rationale annotation needs to be more cost-efficient than label annotation.
In light of this, RQ4 compares the time cost of label and rationale annotation, in terms of their respective impact on model generalization (\ie time budget \versus ID/OOD performance).

\subsubsection{Setup}
\label{sec:exp:rq4:setup}

We conduct an Amazon Mechanical Turk\footnote{\hyperlink{https://www.mturk.com/}{https://www.mturk.com/}} (MTurk) human study to compare the effectiveness of three instance annotation types across various time budgets.
For \textbf{Label Only}, the Turkers (\ie MTurk annotators) are asked to annotate a given instance's task label.
For \textbf{Expl Only}, the Turkers are provided a given instance's ground truth task label, then asked to annotate an extractive rationale by highlighting input tokens that support this label.
For \textbf{Label+Expl}, the Turkers are asked to annotate both the task label and the rationale.
In this study, we consider an initial training set $\mathcal{D}_{\text{init}}$ of 1000 instances and different time budgets $b$.
For Label Only, the Turkers use $b$ to add new instances $\mathcal{D}_{\text{L}}^{(b)}$ with label annotations, yielding combined training set $\{ \mathcal{D}_{\text{init}}, \mathcal{D}_{\text{L}}^{(b)}\}$.
For Expl Only, the Turkers use $b$ to annotate rationales for a subset of instances $\mathcal{D}_{\text{E}}^{(b)} \subseteq \mathcal{D}_{\text{init}}$.
For Label+Expl, the Turkers use $b$ to add new instances $\mathcal{D}_{\text{L+E}}^{(b)}$ with both label and rationale annotations, yielding combined training set $\{ \mathcal{D}_{\text{init}}, \mathcal{D}_{\text{L+E}}^{(b)}\}$.

Since it is difficult to track Turkers' annotation progress over long time periods (\eg 48 hours), we first estimate the annotation time per instance based on a sample of timed instance annotations, then use these estimates to create proportional training sets (w.r.t. number of instances) for each time budget level.
For each annotation type, we obtain the time estimates by asking Turkers to collectively annotate the same 200 SST training instances, which are randomly selected via stratified sampling with respect to sentiment label.
We considered a 200-instance sample for time estimation because we felt that this was a reasonable trade-off between estimation accuracy and annotation cost.
Across the three annotation types, we employ 178 total Turkers, with three Turkers per instance.
On these 200 instances, the Turkers yielded mean $\pm$ std annotation times of $140.56 \pm 8.45$ seconds per instance (Label Only), $110.31 \pm 3.21$ seconds per instance (Expl Only), and $263.10 \pm 7.31$ seconds per instance (Label+Expl).

Based on these estimates, for each annotation type and time budget, we construct training sets by uniformly sampling the following numbers of additional instances from the training set (excluding the initial 1000-instance training set $\mathcal{D}_{\text{init}}$).
For Label Only, we sample training sets of 4 (10 min), 13 (30 min), 128 (5 hr), 615 (24 hr), and 1229 (48 hr) instances.
For Expl Only, we sample training sets of 5 (10 min), 16 (30 min), 163 (5 hr), 783 (24 hr), and 1556 (48 hr) instances.
For Label+Expl, we sample training sets of 2 (10 min), 7 (30 min), 68 (5 hr), 328 (24 hr), and 657 (48 hr) instances.
Due to computational constraints, we limit our RQ4 experiments to sentiment analysis and IxG+MAE, which yielded the highest development ID performance for RQ1 (\textsection \ref{sec:app:training_details}).
See \textsection \ref{sec:app:rq4} for more details.

\begin{table*}[ht]
\hspace{-0.2cm}
\centering
\scalebox{0.70}{
\begin{tabular}{ccccc}
\toprule
\multirow{4}{*}{\makecell[c]{\textbf{Additional Time} \\ \textbf{Budget}}} &\multirow{4}{*}{\makecell[c]{\textbf{Instance} \\ \textbf{Annotation} \\ \textbf{Type}}} & \multicolumn{3}{c}{\textbf{Sentiment Analysis}} \\
\cmidrule(lr){3-5}
& & \multicolumn{3}{c}{IMDb} \\
\cmidrule(lr){3-5}
& & Original Acc ($\uparrow$) & Contrast Acc ($\uparrow$) & Consistency ($\uparrow$) \\
\midrule
0 min & None & 88.02~($\pm$2.34) &83.36~($\pm$4.49)&71.84~($\pm$6.82)\\
\midrule
\multirow{3}*{10 min} & {Label Only} & 85.15~($\pm$4.98)&83.08~($\pm$5.38)  &68.17~($\pm$10.23)\\
 & {Expl Only} & 88.11~($\pm$0.24)&85.05~($\pm$1.98)&73.82~($\pm$2.38)\\
 & {Label+Expl} & 85.22~($\pm$1.19)&79.39~($\pm$2.25)&65.00~($\pm$3.42)\\
\midrule

\multirow{3}*{30 min} & {Label Only} & 86.20~($\pm$4.18)&82.17~($\pm$5.15)  &68.74~($\pm$9.22)\\
& {Expl Only} & 87.80~($\pm$2.18)&83.61~($\pm$1.85)&72.15~($\pm$4.38)\\
 & {Label+Expl} & 83.89~($\pm$4.84)&80.97~($\pm$4.18)&65.22~($\pm$16.89)\\
\midrule

\multirow{3}*{5 hr} & {Label Only} & 87.48~($\pm$1.27)&83.63~($\pm$1.69)  &71.58~($\pm$3.02)\\
& {Expl Only} &87.57~($\pm$1.70)&84.40~($\pm$3.03)&72.40~($\pm$4.56)\\
 & {Label+Expl} & 88.39~($\pm$2.02)&85.68~($\pm$2.87)&74.48~($\pm$4.96)\\
\midrule

\multirow{3}*{24 hr} & {Label Only} & 87.64~($\pm$3.25)&83.49~($\pm$5.46)  &71.56~($\pm$8.78)\\
& {Expl Only} & 88.19~($\pm$2.56)&84.81~($\pm$4.55)&73.39~($\pm$5.42)\\
 & {Label+Expl} & 86.86~($\pm$4.88)&83.45~($\pm$7.02)&70.67~($\pm$11.90)\\
\midrule

\multirow{3}*{48 hr} & {Label Only} & 89.73~($\pm$0.52)&86.86~($\pm$6.11)  &\cellcolor{lblue}77.12~($\pm$1.54)\\
& {Expl Only} & 88.19~($\pm$2.56)&84.81~($\pm$4.55)&73.39~($\pm$5.42)\\
 & {Label+Expl} & 89.78~($\pm$0.18)&87.97~($\pm$4.52)&\cellcolor{lblue}78.74~($\pm$1.08)\\

\bottomrule
\end{tabular}}
\caption{\small \textbf{RQ4 - Contrast Set Tests (\textsection \ref{sec:exp:rq4:contrast}).} We compare ER model performance for three \textit{instance annotation types} across different time budgets, with respect to performance on both original test sets (ID) and contrast sets (OOD). Performance is reported in terms of original test set accuracy (Original Acc), contrast set accuracy (Contrast Acc), and contrast consistency (Consistency). Numbers highlighted in \colorbox{lblue}{blue} indicate statistically significant improvement over the No-ER baseline ($p < 0.05$).}
\label{tab:exp:rq4:contrast}
\end{table*}

\begin{figure*}[t!]
\hspace{-0.3cm}
\includegraphics[width=1.04\linewidth]{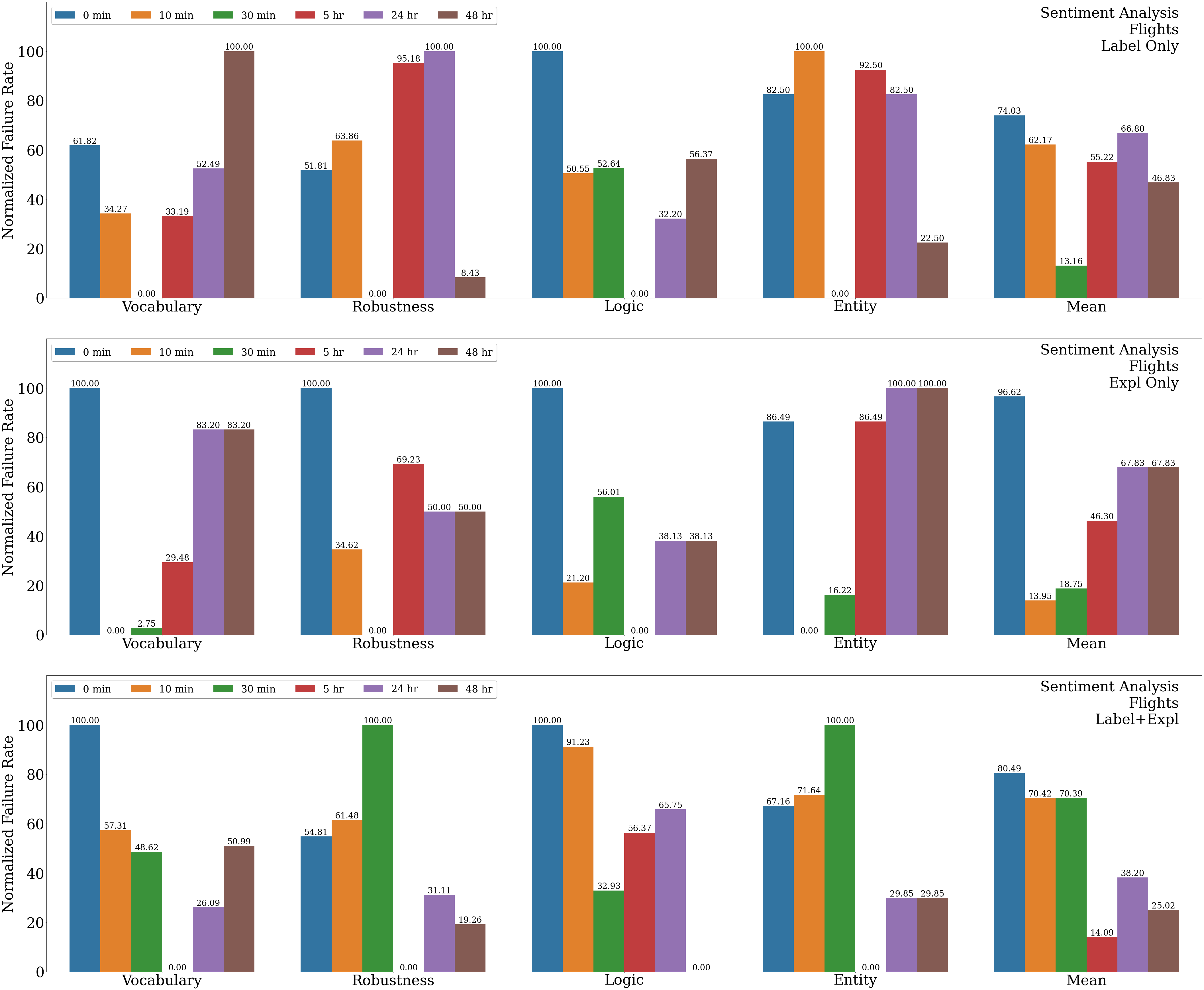}
\vspace{-0.5cm}
\caption{\small \textbf{RQ4 - Functional Tests (\textsection \ref{sec:exp:rq4:functional}).} We compare ER model performance for three \textit{instance annotation types} across different time budgets, with respect to performance on a range of functional tests (OOD). Performance is reported in terms of normalized failure rate ($\downarrow$) for four functional test types (Vocabulary, Robustness, Logic, Entity) as well the mean normalized failure rate across all functional tests (Mean).}
\label{fig:exp:rq4:functional}
\end{figure*}

Below, we discuss our findings from using \method to explore RQ4.
The RQ4 results obtained via unseen dataset tests (\textsection \ref{sec:exp:rq4:unseen}), contrast set tests (\textsection \ref{sec:exp:rq4:contrast}), and functional tests (\textsection \ref{sec:exp:rq4:functional}) are shown in Fig. \ref{fig:exp:rq4:unseen}, Table \ref{tab:exp:rq4:contrast}, and Fig. \ref{fig:exp:rq4:functional}, respectively.

\subsubsection{Unseen Dataset Tests}
\label{sec:exp:rq4:unseen}

Fig. \ref{fig:exp:rq4:unseen} shows the results for unseen dataset set tests on RQ4.
When provided a nonzero additional time budget (\ie 10 min and above), all three instance annotation types can yield models that perform better than models with zero additional time budget (\ie 0 min).
However, the performance trajectory with respect to additional time budget varies significantly across different instance annotation types.

For Label Only, on both seen (SST) and unseen datasets (Amazon, Yelp, Movies), model performance tends to decrease slightly for lower nonzero budgets, before steadily increasing for higher nonzero budgets.
Generally, Label Only annotations require at least 24 hours of additional annotation time before the model's performance begins to noticeably outperform the baseline 0-minute ER model.
Given a 48-hour budget, Label Only yields even greater improvements.
Also, we see that Label+Expl tends to yield similar trends as Label Only, except with smaller performance decreases for lower nonzero budgets and larger performance increases for higher nonzero budgets.

On the other hand, Expl Only immediately yields improvements at the 10-minute mark, but does not always steadily increase with higher budgets.
For the seen dataset (SST), the Expl Only model's performance begins to drop after the time budget exceeds 30 minutes, although the performance stays within a relatively small small range across all time budgets.
For unseen datasets (Amazon, Yelp, Movies), the Expl Only model's performance generally increases with increasing time budgets, but not as drastically as the Label Only or Label+Expl models'.
On average, we find that 30 minutes of Expl Only annotation yields similar performance as 24 hours of Label Only or Label+Expl annotation.

These results suggest that additional annotations may sometimes introduce an annotation distribution shift within the training set, as the annotators employed in our study are different from those in the original dataset.
For annotation types involving the task label (Label Only, Label+Expl), the distribution shift is more drastic, so more annotations (\ie higher budgets) are required to balance out the annotation distribution.
Meanwhile, the opposite appears to be true for Expl Only.
This is likely due to the fact that Label Only and Label+Only both involve adding new instances, whereas Expl Only only involves adding rationale annotations to existing instances.
Ultimately, we conclude that Expl Only is more effective for smaller budgets, while Label Only and Label+Expl are more suitable for larger budgets.
Since annotation budgets tend to be low in real-world scenarios, Expl Only may often be a more practical annotation strategy than Label Only and Label+Expl.


\subsubsection{Contrast Set Tests}
\label{sec:exp:rq4:contrast}

Table \ref{tab:exp:rq4:contrast} shows the results for contrast set tests on RQ4.
For Expl Only, we see that increasing the time budget from 0 min (None) to any other budget under 48 hours yields decreased performance.
However, increasing the Expl Only budget to 48 hours yields significant performance improvements.
Also, for Label+Expl, we see a more extreme version of this trend, with a lower initial performance decrease followed by a higher eventual performance increase.
Thus, Label+Expl with a 48-hour budget yields the highest overall performance.
On the other hand, across all nonzero time budgets under 48 hours, we see that Expl Only generally yields higher performance than None, Label Only, and Label+Expl, yet fails to achieve significant performance improvements as the time budget increases to 48 hours.
This mirrors the findings from our unseen dataset tests on RQ4, hence supporting our conclusion that Expl Only works better for smaller budgets whereas Label Only and Label+Expl work better for larger budgets.

\subsubsection{Functional Tests}
\label{sec:exp:rq4:functional}

Fig. \ref{fig:exp:rq4:functional} shows the results for functional tests on RQ4.
For all instance annotation types, we see that models with nonzero additional time budgets (\ie 10 min and higher) achieve lower mean failure rates than the model with zero additional time budget (\ie 0 min).

For Label Only, using a 30-minute budget yields the lowest mean failure rate by far, while other budgets yield failure rates that are not significantly lower than the 0-minute failure rate.
This Label Only trend is quite different from the trends observed in the unseen dataset tests and contrast set tests, where the best performance was consistently attained using the 48-hour budget.

For Expl Only, using 10-minute and 30-minute budgets yields much lower mean failure rates than the other budgets do, with the failure rate generally increasing with the time budget.
This Expl Only trend is also different from previous trends, in which the performance either increased or stayed about the same as the budget increased.

For Label+Expl, using a 5-hour budget yields the lowest mean failure rate.
Although the failure rate generally decreases as the budget increases, this Label+Expl is still different from previous trends, where the best performance was consistently attained using the 48-hour budget.

In summary, we find that nonzero time budgets lead to lower failure rates than zero time budgets, although there are mixed trends in how the failure rate changes as the nonzero time budget increases.
Like in previous tests, Expl Only shines for lower nonzero budgets but yields diminishing returns for higher nonzero budgets.
However, the opposite tends to be true for Label+Expl.
For Label Only, the results are less conclusive, with the failure rate oscillating dramatically as the budget increases.


\section{Related Work} 
\label{sec:related_work}


\paragraph{Rationale Extraction}
Much of the language model (LM) explainability literature has focused on rationale extraction, which is the process of producing extractive rationales.
An extractive rationale explains an LM's output on a given task instance by scoring input tokens' influence on the LM's output \cite{denil2014extraction, sundararajan2017axiomatic, li2016understanding, jin2019towards, lundberg2017unified, chan2022unirex}.
This token scoring can be done via input gradients~\citep{sundararajan2017axiomatic, lundberg2017unified, denil2014extraction, li2015visualizing}, input perturbation \cite{li2016understanding, poerner2018evaluating, kadar2017representation}, attention weights~\citep{pruthi2020evaluating, stacey2022supervising, wiegreffe2019attention}, or learned rationale extraction models
\cite{chan2022unirex, jain2020learning, situ2021learning}.
In this work, we study how extractive rationales can be used to regularize LMs.

\paragraph{Explanation-Based Learning}
To improve LM behavior, many methods have been proposed for explanation-based learning \cite{hase2021can, hartmann-sonntag-2022-survey}, especially using human-annotated explanations \cite{tan-2022-diversity}.
For extractive rationales, one common paradigm is explanation regularization (ER), which regularizes the LM so that its extractive machine rationales (reflecting LM's reasoning process) are aligned with extractive human rationales (reflecting humans' reasoning process) \cite{ross2017right, huang2021exploring, ghaeini2019saliency, zaidan2008modeling, kennedy2020contextualizing, rieger2020interpretations, liu2019incorporating}.
In ER, the human rationale can be obtained by annotating each instance individually \cite{zaidan2008modeling, lin2020triggerner, camburu2018snli, rajani2019explain, deyoung2019eraser} or by applying domain-level lexicons across all instances \cite{rieger2020interpretations, ross2017right, ghaeini2019saliency, kennedy2020contextualizing, liu2019incorporating}.
Beyond ER, there are other ways to learn from extractive rationales.
\citet{lu-etal-2022-rationale} used human-in-the-loop feedback on machine rationales for data augmentation.
\citet{ye-durrett-2022-explanations} used machine rationales to calibrate black box models and improve their performance on low-resource domains.

\paragraph{Evaluating ER Models}
Existing works have primarily evaluated ER models via ID generalization \cite{zaidan2008modeling, lin2020triggerner, huang2021exploring}, which only captures one aspect of ER's impact.
However, a few works have considered auxiliary evaluations, such as machine-human rationale alignment \cite{huang2021exploring, ghaeini2019saliency}, task performance on unseen datasets \cite{ross2017right, kennedy2020contextualizing}, and social group fairness \cite{rieger2020interpretations, liu2019incorporating}.
\citet{carton-etal-2022-learn} showed that maximizing machine-human rationale alignment does not always improve task performance, while human rationales vary in their ability to provide useful information for task prediction.
Meanwhile, \method jointly evaluates ER models' OOD generalization along three dimensions: unseen dataset tests, contrast set tests \cite{gardner2020evaluating}, and functional tests \cite{ribeiro2020beyond, li-etal-2020-linguistically}.

\section{Conclusion} 
\label{sec:conclusion}



\paragraph{Summary of Contributions}

In this paper, we primarily study ER from the perspective of OOD generalization.
We propose \method, a framework for evaluating ER models' OOD generalization along three dimensions: unseen dataset tests, contrast set tests, and functional tests.
Using \method, we investigate four research questions: (A) Which rationale alignment \textit{criteria} are most effective? (B) Is ER effective with \textit{task-level} human rationales? (C) How is ER affected by the \textit{number and choice} of rationale-annotated instances? (D) How does ER performance vary with the rationale annotation \textit{time budget}?

For two text classification tasks and six datasets, \method shows that ER has little impact on ID performance but yields large gains on OOD performance, with the best ER criteria being task-dependent (\textsection \ref{sec:exp:rq1}). 
Furthermore, ER can improve OOD performance even with distantly-supervised (\textsection \ref{sec:exp:rq2}) or few (\textsection \ref{sec:exp:rq3}) human rationales.
Finally, we find that rationale annotation is more time-efficient than label annotation, in terms of impact on OOD performance (\textsection \ref{sec:exp:rq4}).
These results from \method help demonstrate ER's utility and establish best practices for using ER effectively.


\paragraph{Future Work}

Our findings suggest the promise of several directions for future work in improving LM generalization.
Here, we discuss each of these future directions.

First, in this paper, we focused on applying \method to extractive rationales, which use input token scoring to explain the reasoning process behind a given task output.
Meanwhile, free-text rationales (FTRs) explain reasoning processes via natural language \cite{camburu2018snli, rajani2019explain, wei2022chain, wang2022pinto, chan2022knife}.
Compared to extractive rationales, FTRs may be more intuitive to humans, can reference things beyond the task input, and support high flexibility in content, style, and length \cite{wiegreffe2021measuring, chan2022frame}.
In the future, it would be interesting to repurpose \method to support the evaluation of FTR-based LM regularization.

Second, existing ER works generally consider the offline setting where human feedback is only collected once, and the ER model is only trained once using this human feedback \cite{ross2017right, huang2021exploring, ghaeini2019saliency, zaidan2008modeling, kennedy2020contextualizing, rieger2020interpretations, liu2019incorporating}.
Thus, \method also follows this setting.
However, as the ER model is updated, it may benefit from multiple rounds of adaptive human feedback, with each round tailored to the ER model's specific weaknesses during that point in training.
In the future, it would be interesting to repurpose \method for online, human-in-the-loop (HITL) LM debugging \cite{idahl2021towards, lertvittayakumjorn-etal-2020-find, zylberajch-etal-2021-hildif, ribeiro2016should, lee2022xmd}.
This would likely involve exploring how active learning can be more effectively used to select instances for collecting human feedback.

Third, in this work, we focused on using \method to measure the extent to which rationales can improve models' OOD generalization ability.
However, rationales can also be evaluated with respect to explainability desiderata such as faithfulness, plausibility, and human utility \cite{deyoung2019eraser, chan2022unirex, chan2022frame, joshi2022measuring}.
In the future, it would be interesting to incorporate elements from \method to improve benchmarks for evaluating explainability performance.

\section{Limitations}



\textbf{\method's contrast set tests and functional tests are generally expensive to create.}
In our experiments, we considered off-the-shelf contrast set tests and functional tests released by prior works.
However, not many tasks have such pre-constructed tests already available to use.
Plus, each instantiation of these tests requires significant manual effort to create, which does not scale well to the numerous existing NLP tasks.
This limits the applicability of \method to tasks for which researchers have invested considerable resources to create contrast set tests and functional tests.

\textbf{Certain tests in \method may not be applicable to all NLP tasks.}
For token classification tasks (\eg NER), it is not straightforward to construct contrast set tests.
This is because creating contrast set tests for token classification would require pivoting each instance with respect to each token in the instance's input sequence, yet it is unclear how this would be done feasibly.
This essentially limits the applicability of \method to sequence classification tasks.

\textbf{\method covers a limited region in the space of OOD evaluation.}
As described earlier, \method consists of three types of tests: unseen dataset tests, contrast set tests, and functional tests.
However, these tests alone do not account for all aspects of OOD generalization.
In the future, it would be interesting to augment \method with additional tests that evaluate other aspects of OOD generalization.
With a wider range of tests, we can better understand ER's impact on OOD generalization and design better methods for improving LMs' OOD generalization.




\section{Ethics Statement}

\paragraph{Data}
All datasets used in our work are freely available for public use and have been duly attributed to their original authors. 

\paragraph{User Study}
For the RQ4 user study in \textsection \ref{sec:exp:rq4}, we measured the time needed for Turkers to annotate task labels and rationales for different sentiment analysis instances.
All annotation instructions are presented to Turkers in an accessible manner, as shown in Fig. \ref{fig:app:rq4:label}-\ref{fig:app:rq4:label+expl}.
Plus, we actively corresponded with Turkers over email to address their questions/concerns regarding the rejection of human intelligence task (HIT) submissions.
As a result of our thorough email discussions with the Turkers, we actually reversed many of our HIT rejections, thus ensuring that Turkers were properly rewarded for their efforts.
Finally, we made sure to provide fair compensation to all Turkers, paying them above the United States minimum wage of \$16 per hour.
See \textsection \ref{sec:app:rq4} for more details. 
\section{Acknowledgments}

This research is supported in part by the Office of the Director of National Intelligence (ODNI), Intelligence Advanced Research Projects Activity (IARPA), via Contract No. 2019-19051600007, NSF IIS 2048211, and gift awards from Google, Amazon, JP Morgan, and Sony.
We would like to thank all of our collaborators at USC NLP Group, USC INK Research Lab, and Meta AI for their constructive feedback on this work.
\bibliography{references}
\bibliographystyle{acl_natbib}

\appendix
\section{Appendix} 
\label{sec:appendix}

\subsection{ER Training Details}
\label{sec:app:training_details}

Though ER involves many design choices (\ie hyperparameters), it is infeasible to comprehensively tune all of these hyperparameters. 
Hence, for hyperparameters that yielded little sensitivity in our initial experiments, were not ER-specific, and/or were not the focus of our RQs, we used fixed values across all of our subsequent experiments.
We discuss these global hyperparameter values below.
First, we use a learning rate of $2\mathrm{e}{-5}$.
Second, we use a batch size of $32$.
Third, we set the ER strength to $\lambda_{\text{ER}}=1$.
Fourth, we train all models for a maximum of $25$ epochs, with early stopping.
We perform early stopping based on total development set loss (\ie task loss + ER loss) and a patience of $10$ epochs.
Fifth, before calculating the ER loss between machine rationales (\ie attribution scores) and human rationales, we use the sigmoid function to normalize the attribution scores as probabilities.
However, for a given instance, if all tokens' attribution scores are low, then all token probabilities will be close to $0.5$ and provide little signal for identifying important tokens.
Thus, we scale the raw attribution scores by $100$, so that the normalized attribution scores become closer to $0$ or $1$.

\subsection{Development Set Results}
In \textsection \ref{sec:exp}, we only reported ER model performance on test sets, even though we tuned our ER hyperparameters based on the development sets (for the seen dataset).
Furthermore, recall that we used RQ1's development set results to obtain the respective subsets of ER settings used for RQ2-RQ4.
Thus, for reference, we also report the development set results for RQ1, RQ2, RQ4, and RQ4 in Tables \ref{app:tab:rq1:dev}, \ref{app:tab:rq2:dev}, \ref{app:tab:rq3:dev}, and \ref{app:tab:rq4:dev}, respectively.

\begin{table}[!ht]
\centering
\scalebox{0.70}{
\begin{tabular}{cccccccccc}
    \toprule
    \multirow{4}{*}{\makecell[c]{\textbf{Machine} \\ \textbf{Rationale} \\ \textbf{Extractor}}} & \multirow{4}{*}{\makecell[c]{\textbf{Rationale} \\ \textbf{Alignment} \\ \textbf{Criterion}}} & \textbf{Sentiment Analysis} & \textbf{NLI}\\
    \cmidrule{3-4} 
    & & SST & e-SNLI \\
    \cmidrule{3-4} 
    & & Acc ($\uparrow$) & F1 ($\uparrow$) \\
    \midrule
    \multirow{6}{*}{IxG} & {MSE} & 93.39~($\pm$0.93) & 78.66~($\pm$0.42) \\
    & {MAE} & \textbf{93.93}~($\pm$0.24) & 78.46~($\pm$0.79)  \\
    & {Huber} & 93.89~($\pm$0.61) &  78.69~($\pm$0.29)  \\
    & {BCE} & 93.35~($\pm$0.70) & 78.57~($\pm$0.81)\\
    & {KLDiv} & 92.89~($\pm$0.30) & 73.41~($\pm$5.09)  \\
    & {Order} & 90.83~($\pm$1.91) & \textbf{78.93}~($\pm$0.74)  \\
    \midrule
      \multirow{6}{*}{Attention} &  {MSE} & 93.27~($\pm$0.78) & 75.41~($\pm$0.42) \\
    & {MAE} & 93.65~($\pm$0.63) & 76.27~($\pm$0.92)  \\
    & {Huber} & 93.08~($\pm$0.07) &  75.98~($\pm$0.68)  \\
    & {BCE} & 93.46~($\pm$0.90) & 75.55~($\pm$0.40)\\
    & {KLDiv} & 93.50~($\pm$0.54) & \textbf{76.69}~($\pm$0.49)  \\
    & {Order} & \textbf{93.98}~($\pm$0.08) & 75.99~($\pm$0.24)  \\
    \midrule
    \multirow{6}{*}{UNIREX} & {MSE} & 92.43~($\pm$0.20) & 73.77~($\pm$1.81) \\
    & {MAE} & \textbf{93.31}~($\pm$0.57) & 72.92~($\pm$1.45)  \\
    & {Huber} & 92.47~($\pm$0.76) &  73.33~($\pm$2.12)  \\
    & {BCE} & 92.89~($\pm$0.11) & 68.81~($\pm$1.54)\\
    & {KLDiv} & 92.43~($\pm$0.23) & \textbf{74.05}~($\pm$1.20)  \\
    & {Order} & 93.00~($\pm$0.23) & 72.26~($\pm$0.82)  \\
    \bottomrule 
\end{tabular}
}
\caption{\small \textbf{RQ1 - Development Set Performance (\textsection \ref{sec:exp:rq1:unseen}).}
For each machine rationale extractor and task/dataset, the best-performing rationale alignment criterion is indicated in \textbf{bold}.
}
\label{app:tab:rq1:dev}
\end{table}

\begin{table}[!ht]
\centering
\scalebox{0.70}{
\begin{tabular}{cccc}
    \toprule
    \multirow{4}{*}{\makecell[c]{\textbf{Human} \\ \textbf{Rationale} \\ \textbf{Type}}} & \multirow{4}{*}{\makecell[c]{\textbf{Machine} \\ \textbf{Rationale} \\ \textbf{Extractor}}} & \multirow{4}{*}{\makecell[c]{\textbf{Rationale} \\ \textbf{Alignment} \\ \textbf{Criterion}}} & {\textbf{Sentiment Analysis}} \\
    \cmidrule{4-4}
    & & & SST \\
    \cmidrule{4-4}
    & & & Acc ($\uparrow$) \\
    \midrule
    \multirow{4}{*}{IxG} & \multirow{2}{*}{MAE} & {Instance-level} &  92.93~($\pm$0.24)  \\
    & & {Task-level} &  \textbf{93.46}~($\pm$0.11)  \\
    \cmidrule{2-4}
    & \multirow{2}{*}{Huber} & {Instance-level} &  92.89~($\pm$0.61) \\
    & & {Task-level} &  \textbf{93.27}~($\pm$0.48) \\
    \midrule
    \multirow{4}{*}{Attention} & \multirow{2}{*}{MAE} & {Instance-level} &  93.65~($\pm$0.63)  \\
    & & {Task-level} &  \textbf{94.31}~($\pm$0.71)  \\
    \cmidrule{2-4}
    & \multirow{2}{*}{Huber} & {Instance-level} &  93.08~($\pm$0.07) \\
    & & {Task-level} &  \textbf{93.98}~($\pm$0.33) \\
    \midrule
    \multirow{4}{*}{UNIREX} & \multirow{2}{*}{MAE} & {Instance-level} & \textbf{93.31}~($\pm$0.57) \\
    & & {Task-level} &  92.34~($\pm$1.11)  \\
    \cmidrule{2-4}
    & \multirow{2}{*}{Huber} & {Instance-level} &  92.47~($\pm$0.76) \\
    & & {Task-level} &  \textbf{93.45}~($\pm$0.86) \\
    \bottomrule 
\end{tabular}
}
\caption{\small \textbf{RQ2 - Development Set Performance (\textsection \ref{sec:exp:rq2:unseen}).}
For each machine rationale extractor and rationale alignment criterion, the best-performing human rationale type is indicated in \textbf{bold}.
}

\label{app:tab:rq2:dev}
\end{table}

\begin{table}[!ht]
\centering
\scalebox{0.75}{
\begin{tabular}{cccc}
    \toprule
    \multirow{4}{*}{\makecell[c]{\textbf{Instance} \\ \textbf{Annotation} \\ \textbf{Budget} ($k$)}}& \multirow{4}{*}{\makecell[c]{\textbf{Instance} \\ \textbf{Selection} \\ \textbf{Strategy}}} & \textbf{Sentiment Analysis} \\
    \cmidrule{3-3} 
    & & SST \\
    \cmidrule{3-3} 
    & & Acc($\uparrow$) \\
    \midrule
    \multirow{5}{*}{5\%} & {Random} & \textbf{93.58}~($\pm$0.23)  \\
    & {LC} & 92.55~($\pm$0.72)   \\
    & {HC} & 93.08~($\pm$0.35) \\
    & {LIS} & 92.62~($\pm$0.40) \\
    & {HIS} & 92.97~($\pm$0.52)  \\
    \midrule
      \multirow{5}{*}{15\%} 
      &  {Random} & 93.08~($\pm$0.28)  \\
    & {LC} & 92.24~($\pm$0.78)   \\
    & {HC} & 92.58~($\pm$0.46) \\
    & {LIS} & 93.08~($\pm$0.66) \\
    & {HIS} & \textbf{93.31}~($\pm$0.18)  \\
    \midrule
    \multirow{5}{*}{50\%} 
      &  {Random} & 92.29~($\pm$0.06)  \\
    & {LC} & 87.58~($\pm$5.11)   \\
    & {HC} & \textbf{92.62}~($\pm$0.81) \\
    & {LIS} & 92.24~($\pm$0.54) \\
    & {HIS} & 92.51~($\pm$0.37)  \\
    \bottomrule 
\end{tabular}
}
\caption{\small \textbf{RQ3 - Development Set Performance (\textsection \ref{sec:exp:rq3:unseen}).}
For each instance annotation budget, the best-performing instance selection strategy is indicated in \textbf{bold}.
}
\label{app:tab:rq3:dev}
\end{table}

\begin{table}[!ht]
\centering
\scalebox{0.75}{
\begin{tabular}{cccc}
    \toprule
    \multirow{4}{*}{\makecell[c]{\textbf{Instance} \\ \textbf{Annotation} \\ \textbf{Type}}} & \multirow{4}{*}{\makecell[c]{\textbf{Additional} \\ \textbf{Time} \\ \textbf{Budget}}} & \textbf{Sentiment Analysis} \\
    \cmidrule{3-3}
    & & SST \\
    \cmidrule{3-3}
    & & Acc ($\uparrow$) \\
    \midrule
    None & 0 min & 91.00~($\pm$0.18)\\
    \midrule
    \multirow{3}{*}{10 min} & {Label Only} &  91.09~($\pm$0.18)  \\
    & {Expl Only} &  \textbf{91.50}~($\pm$0.31)  \\
    & {Label+Expl} &  90.66~($\pm$0.27)  \\

    \midrule

    \multirow{3}{*}{30 min} & {Label Only} &  90.88~($\pm$0.66) \\
    & {Expl Only} &  \textbf{91.23}~($\pm$0.21) \\
    & {Label+Expl} &  91.08~($\pm$0.18) \\

    \midrule

    \multirow{3}{*}{5 hr} & {Label Only} & \textbf{91.62}~($\pm$0.32) \\  
    & {Expl Only} & 90.86~($\pm$0.77) \\
    & {Label+Expl} & 91.30~($\pm$0.70) \\   

    \midrule
    
    \multirow{3}{*}{24 hr} & {Label Only} & \textbf{92.42}~($\pm$0.67)  \\ 
    & {Expl Only} & 91.24~($\pm$0.06)  \\
    & {Label+Expl} & 91.97~($\pm$0.53)  \\  

    \midrule

    \multirow{3}{*}{48 hr} & {Label Only} & \textbf{92.70}~($\pm$0.47)  \\ 
    & {Expl Only} & 91.24~($\pm$0.06)  \\
    & {Label+Expl} & 92.37~($\pm$0.53)  \\

    \bottomrule 
\end{tabular}
}
\caption{\small \textbf{RQ4 - Development Set Performance (\textsection \ref{sec:exp:rq4:unseen}).}
For each time budget, the best-performing instance annotation type is indicated in \textbf{bold}.
}
\label{app:tab:rq4:dev}
\end{table}

\subsection{RQ1: \textit{Which rationale alignment criteria are most effective for ER?}}
\label{sec:app:rq1}

In this section, we present additional RQ1 experiments beyond those presented in \textsection \ref{sec:exp:rq1}.

\subsubsection{Named Entity Recognition}
\label{sec:app:rq1:ner}

Besides sentiment analysis and NLI, we also consider the named entity recognition (NER) task in unseen dataset tests for RQ1.
For NER, we use CoNLL-2003 \cite{sang2003introduction, lin2020triggerner} as the seen dataset (since it has rationale annotations) and OntoNotes v5.0 \cite{pradhan2013towards} as the unseen dataset.
CoNLL-2003 contains only text from Reuters news stories, while OntoNotes v5.0 contains text from newswires, magazines, telephone conversations, websites, and other sources.

Table \ref{tab:app:task_perf_ner} displays the NER unseen dataset test results for RQ1.
For the seen dataset, we see more variance (versus what we saw in Table \ref{tab:exp:rq1:unseen}) in task performance among different ER criteria, although the variance is still quite small among the best criteria (MSE, MAE, Huber).
Here, MAE yields the highest performance, while BCE yields the lowest by far.
For the unseen dataset, MAE still performs best, while MSE and Huber are competitive.
Meanwhile, BCE again performs the worst.

\begin{table}[!ht]
\centering
\scalebox{0.70}{
\begin{tabular}{cccc}
    \toprule
    \multirow{4}{*}{\makecell[c]{\textbf{Machine} \\ \textbf{Rationale} \\ \textbf{Extractor}}} & \multirow{4}{*}{\makecell[c]{\textbf{Rationale} \\ \textbf{Alignment} \\ \textbf{Criterion}}} & \multicolumn{2}{c}{\textbf{NER}} \\
   \cmidrule(lr){3-4}
  & & Seen Acc ($\uparrow$) & Unseen Acc ($\uparrow$) \\
  \cmidrule(lr){3-3} \cmidrule(lr){4-4} 
    & & CoNLL-2003 & OntoNotes v5.0\\
    \midrule
    - & {No-ER} & 77.24~($\pm$0.20) & 20.78~($\pm$0.41)\\
    \midrule
    \multirow{5}{*}{IxG} &{MSE}  & 78.02~($\pm$0.69) & 21.60~($\pm$0.46)\\
    &{MAE}  & \textbf{78.34}~($\pm$0.81) & \textbf{21.73}~($\pm$0.31)\\
    &{Huber} & 77.83~($\pm$1.09) & 21.38~($\pm$0.16)\\
    &{BCE} & 64.53~($\pm$13.22) & 17.32~($\pm$3.59)\\
    &{Order} & 72.62~($\pm$5.01) & 19.14~($\pm$1.75)\\

    \bottomrule 
\end{tabular}
}
\caption{\small \textbf{RQ1 - Unseen Dataset Tests for NER (\textsection \ref{sec:app:rq1:ner}).}
For NER, we compare various ER \textit{rationale alignment criteria} using the IxG machine rationale extractor (as well as the No-ER baseline), with respect to performance on seen (ID) and unseen (OOD) datasets.
Performance is measured using accuracy (Acc).
For each dataset and metric, the best-performing criterion is indicated in \textbf{bold}.
}
\label{tab:app:task_perf_ner}
\end{table}

\subsubsection{Hate Speech Detection}
\label{sec:app:rq1:hate}

\begin{table*}[!h]
\centering
\scalebox{0.70}{
\begin{tabular}{cccccccc}
    \toprule
    \multirow{4}{*}{\makecell[c]{\textbf{Machine} \\ \textbf{Rationale} \\ \textbf{Extractor}}} & \multirow{4}{*}{\makecell[c]{\textbf{Rationale} \\ \textbf{Alignment} \\ \textbf{Criterion}}} & \multicolumn{6}{c}{\textbf{Hate Speech Detection}} \\
    \cmidrule(lr){3-8}
    \cmidrule(lr){3-4} \cmidrule(lr){5-8}
    & & \multicolumn{2}{c}{Stf} & \multicolumn{2}{c}{HatEval} & \multicolumn{2}{c}{GHC} \\
    \cmidrule(lr){3-4} \cmidrule(lr){5-6} \cmidrule(lr){7-8}
    & & Seen Acc~($\uparrow$) & Seen FPRD~($\downarrow$) & Unseen Acc~($\uparrow$) & Unseen FPRD~($\downarrow$) & Unseen Acc~($\uparrow$) & Unseen FPRD~($\downarrow$) \\
    \midrule
    - & {No-ER} & 89.50~($\pm$0.20) & 1.11~($\pm$0.58) & 63.68~($\pm$0.78) & 1.64~($\pm$0.66) & 89.43~($\pm$0.98) & 1.09~($\pm$0.12)\\
    \midrule
    \multirow{5}{*}{IxG} & {MSE}  & 89.46~($\pm$0.21) & 2.18~($\pm$0.47) & 64.30~($\pm$1.52) & 1.99~($\pm$0.26) & 88.19~($\pm$0.62) & 1.50~($\pm$0.10)\\
    & {MAE}  & \textbf{89.59}~($\pm$0.06) & 1.39~($\pm$0.62) & 63.30~($\pm$0.49) & 1.80~($\pm$0.59) & 88.07~($\pm$1.66) & 1.43~($\pm$0.24)\\
    & {Huber} & 89.50~($\pm$0.51) & 1.90~($\pm$0.35) & \textbf{64.85}~($\pm$1.50) & 2.11~($\pm$0.27) & 87.77~($\pm$1.21) & 1.84~($\pm$0.34)\\
    & {BCE} & 89.42~($\pm$0.71) & 1.87~($\pm$0.45) & 63.54~($\pm$0.57) & 1.87~($\pm$0.45) & 88.99~($\pm$0.83) & 1.36~($\pm$0.58)\\
    & {Order} & 89.21~($\pm$1.18) & \textbf{0.56}~($\pm$0.09) & 64.46~($\pm$1.18) & \textbf{0.92}~($\pm$0.92) & \textbf{92.84}~($\pm$0.46) & \textbf{0.59}~($\pm$0.25)\\
    
    \bottomrule 
\end{tabular}
}
\caption{\small \textbf{RQ1 - Unseen Dataset Tests for Hate Speech Detection (\textsection \ref{sec:app:rq1:hate}).}
For hate speech detection, we compare various ER \textit{rationale alignment criteria} using the IxG machine rationale extractor (as well as the No-ER baseline), with respect to performance on seen (ID) and unseen (OOD) datasets.
Performance is measured using accuracy (Acc) and false positive difference rate (FPRD).
For each metric, the best-performing criterion is indicated in \textbf{bold}.
Note that we only consider task-level rationales for hate speech detection, since instance-level rationales are unavailable.
}
\label{tab:exp:ext:hate_results}
\end{table*}

Besides sentiment analysis, NLI, and NER, we also consider the hate speech detection task in unseen dataset tests for RQ1.
Unlike the other tasks we consider, hate speech detection datasets typically provide task-level rationales by default, while instance-level rationales are unavailable. 

\paragraph{Task-Level Rationales}
Many existing hate speech detection models are largely oversensitive to certain group identifier words (\eg ``black'', ``Muslim'', ``gay''), almost always predicting hate speech for text containing these words \cite{kennedy2020contextualizing}.
To address this, prior works first manually annotated a lexicon of group identifiers that should be ignored for hate speech detection.
Then, for all training instances, they automatically marked only tokens belonging to the lexicon as unimportant (and the rest as important).
By using these human rationales for ER, they trained the LM to be less biased with respect to these group identifiers \cite{kennedy2020contextualizing, jin-etal-2021-transferability}.

\paragraph{Datasets}
For hate speech detection, we use Stormfromt (Stf) dataset \cite{de-gibert-etal-2018-hate} as the seen dataset, for which we use the lexicons from \cite{jin-etal-2021-transferability} to generate distantly-supervised rationales.
Each instance in the Stf dataset is matched to one or more lexicons by simple character-level matching, and the rationales are generated as described above.
Then, we train model $\mathcal{F}$ on the Stf dataset.
Meanwhile, we use HatEval \cite{barbieri-etal-2020-tweeteval} and Gab Hate Corpus (GHC) \cite{kennedyghc} as the unseen datasets.
All of these datasets contain binary labels for hateful and non-hateful content.
The Stf dataset is collected from a white-supremacist forum, whereas HatEval instances are tweets and GHC instances are taken from the Gab forum.

\paragraph{Fairness Evaluation}
In addition to task performance, we evaluate models with respect to fairness (\ie bias against group identifiers in the lexicons).
We measure fairness using the false positive rate difference (FPRD) metric \cite{jin-etal-2021-transferability}.
FPRD is computed as $\sum_{z} |\text{FPR}_{z} - \text{FPR}_{\text{overall}}|$, where $\text{FPR}_{z}$ is the model's false positive rate across all test instances mentioning group identifier $z$, and $\text{FPR}_{\text{overall}}$ is the model's false positive rate across all test instances.
In other words, FPRD evaluates the extent to which $\mathcal{F}$ is biased against group identifier $z$.
Lower FPRD indicates that $\mathcal{F}$ has lower bias against the group identifiers in the lexicons.

\paragraph{Results}
Table \ref{tab:exp:ext:hate_results} presents the hate speech detection results for unseen dataset tests with respect to RQ1.
Like in \textsection \ref{sec:exp:rq1}, compared to No-ER, ER does not yield significant increases or decreases in task performance (accuracy) on the seen dataset (Stf).
However, ER yields slightly different results for the unseen datasets.
For GHC, the ER model trained with Order criterion is the only ER model to achieve significant accuracy improvement over No-ER.
Meanwhile, for HatEval, although Huber performs best, none of the ER models achieve significant accuracy improvement over No-ER.

For fairness evaluation, we find that ER models tend to yield higher FPRD than No-ER.
However, we observe that the Order criterion consistently yields the lowest FPRD among all models on both seen and unseen datasets.
In particular, Order's FPRD is significantly lower than No-ER's.
This suggests that using a more relaxed rationale alignment objective is helpful for improving group fairness.
Our findings are in line with those in \citet{huang2021exploring}, which proposed the Order criterion. 

\subsection{RQ2: \textit{How effective are task-level human rationales for ER?}}
\label{sec:app:rq2}

In this section, we provide additional details about our RQ2 experiments (\textsection \ref{sec:exp:rq2}).

 

\subsubsection{Creating Task-Level Rationales}
\label{sec:app:rq2:creating_tasklevel_rationales}

As stated in \textsection \ref{sec:eval:human_rationales} and \textsection \ref{sec:exp:rq2:setup}, task-level rationales are created using task-level lexicons.
Below, we provide details about how the lexicons are used to create rationales.

For a given task $T$, let $L_{T}$ be a human-annotated lexicon (\ie set) of words/phrases that are known to be either important or unimportant to $T$.
For example, sentiment analysis lexicons generally contain words/phrases that are strongly indicative of sentiment (\ie important) \cite{afinn, senticnet}, whereas hate speech detection lexicons generally contain words/phrases that should be ignored when detecting hate speech (\ie unimportant) \cite{jin-etal-2021-transferability}.

Given token $x$, let $\mathbbm{1}_{L_{T}}(\cdot)$ be an indicator function, such that $\mathbbm{1}_{L_{T}}(x) = 1$ if $x \in L_{T}$ (\ie $x$ is part of at least one word/phrase in $L_{T}$) and $\mathbbm{1}_{L_{T}}(x) = 0$ otherwise.
Recall that $\mathbf{x}_i = [x_{i}^{t}]_{t=1}^{n}$ denotes the $n$-token sequence for a task instance $i$ (\textsection \ref{sec:background}).
By applying $\mathbbm{1}_{L_{T}}(\cdot)$ to each token $x_{i}^{t}$ of each instance $i$ in a dataset for $T$, we can obtain a distantly-supervised rationale $\mathbf{\dot{r}}_{i} = [\mathbbm{1}_{L_{T}}(x_{i}^{t})]_{t=1}^{n}$ for every instance in the dataset.




\paragraph{Sentiment Analysis}
For sentiment analysis, we used AFINN \cite{afinn} and SenticNet \cite{senticnet} as source lexicons to create task-level rationales.
By combining AFINN and SenticNet, we obtain a unified lexicon of 170K total words/phrases.
Since some words/phrases appear in both source lexicons, 136 of these words/phrases have conflicting sentiments (\ie have positive sentiment in one lexicon and negative sentiment in the other), so we discard these 136 words/phrases from the unified lexicon.

We find that 93\% of the SST training instances (\ie 6441 instances) have at least one token in the lexicon, which means that the remaining 7\% do not have task-level rationales.
Nonetheless, we use all training instances for ER training.
If a given instance has a task-level rationale, then we use both ER loss and task loss for that instance.
Otherwise, we only use task loss for that instance.
To facilitate a fair comparison with instance-based rationales, we assume that instance-level rationales are also unavailable for the same 7\% of training instances for which task-level rationales are unavailable.


\subsection{RQ3: \textit{How is ER affected by the number and choice of training instances with human rationales?}}
\label{sec:app:rq3}

In this section, we provide additional details about our RQ3 experiments (\textsection \ref{sec:exp:rq3}).

\subsubsection{Setup}
\label{sec:app:rq3:setup}


Since we average all results over three seeds in \textsection \ref{sec:exp}, we describe how this applies to instance selection in RQ3.
For random instance selection, we uniformly sample three subsets from the training set, with each subset containing $k\%$ of the training instances.
For each non-random instance selection strategy (LC, HC, LIS, HIS), we obtain an aggregate selection score for each training instance by first using the given strategy to compute the selection score for each seed, then taking the mean of these three seed-level scores.
After that, we rank all instances in descending order of selection score and select the top-$k\%$ instances based on this ranking.

Given that most training instances do not have rationale annotations in these low-resource settings, it is possible that a given training batch does not contain any rationale-annotated instances, which makes ER impossible for this batch.
To address this, our RQ3 experiments use a modified dataloader that ensures at least one-third of the instances in each training batch have rationale annotations.
As a result, the ER loss can be computed for a substantial number of instances per batch, so that the ER loss' impact is not dwarfed by the task loss'.
Note that the ER loss is only used for rationale-annotated instances in the batch, whereas the task loss is used for all instances in the batch.



\subsection{RQ4: \textit{How is ER affected by the time taken to annotate human rationales?}}
\label{sec:app:rq4}

In this section, we provide additional details about our RQ4 experiments (\textsection \ref{sec:exp:rq4}).

\subsubsection{Setup}
\label{sec:app:rq4:setup}

\begin{figure*}[ht!]
    \centering
    \includegraphics[width=0.95\linewidth]{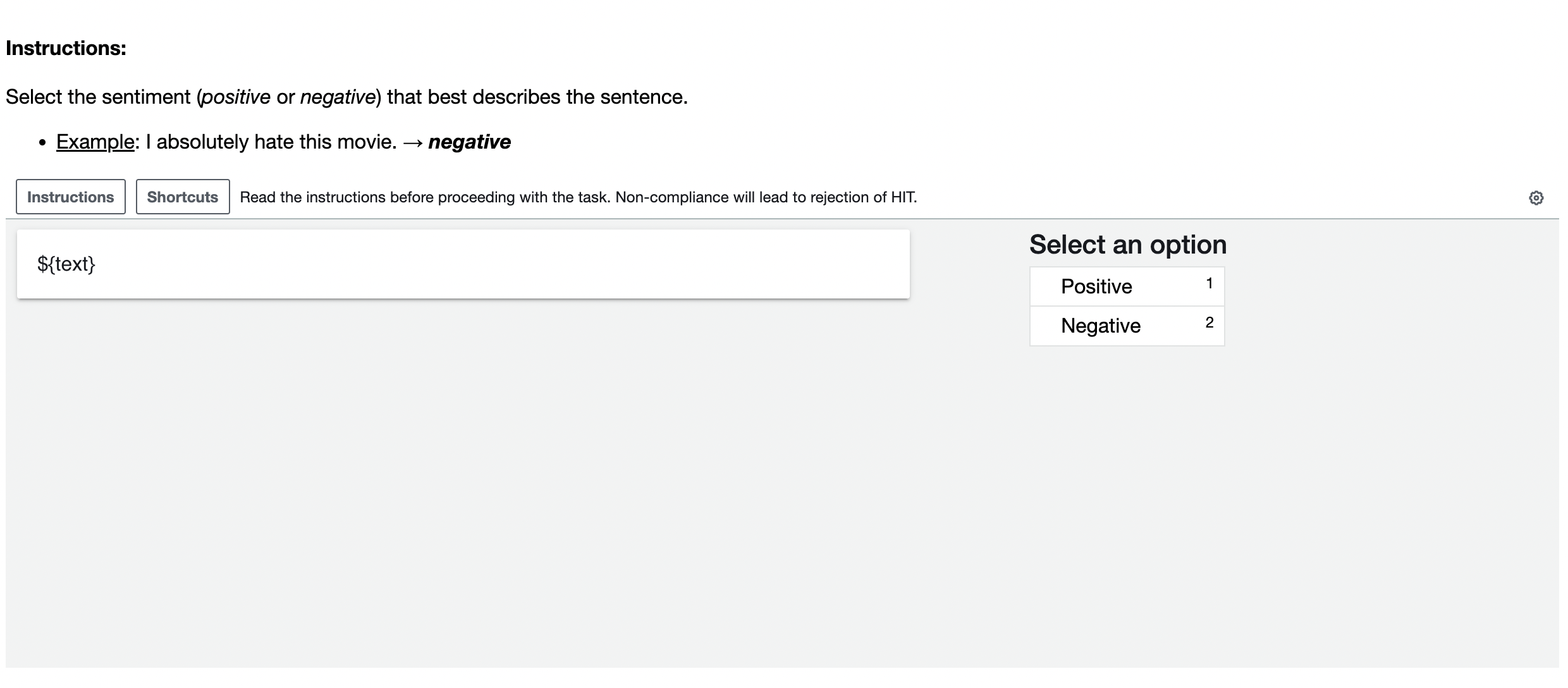}
    \caption{\textbf{RQ4 - Label Only UI.} UI used for Label Only MTurk annotations.}
    \label{fig:app:rq4:label}
\end{figure*}
\begin{figure*}[ht!]
    \centering
    \includegraphics[width=0.95\linewidth]{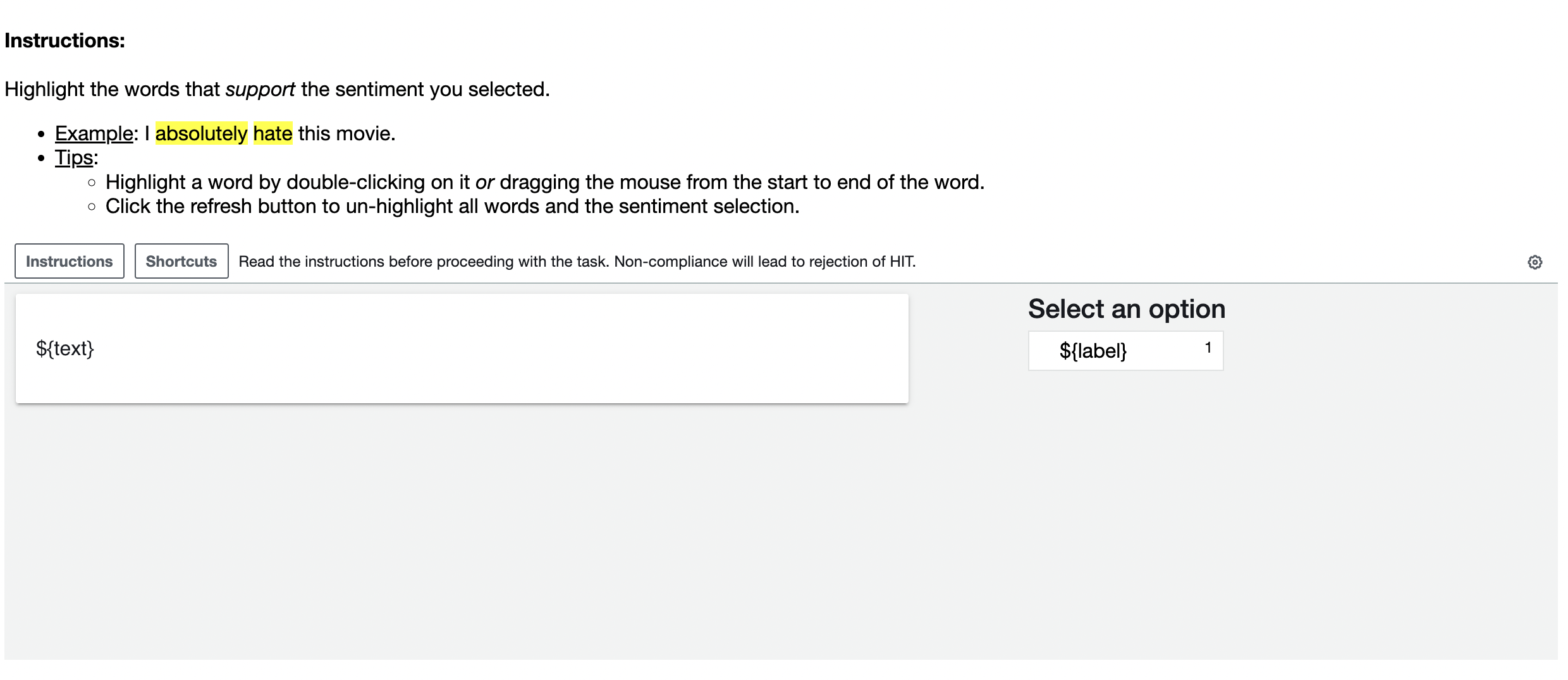}
    \caption{\textbf{RQ4 - Expl Only UI.} UI used for Expl Only MTurk annotations.}
    \label{fig:app:rq4:expl}
\end{figure*}
\begin{figure*}[ht!]
    \centering
    \includegraphics[width=0.95\linewidth]{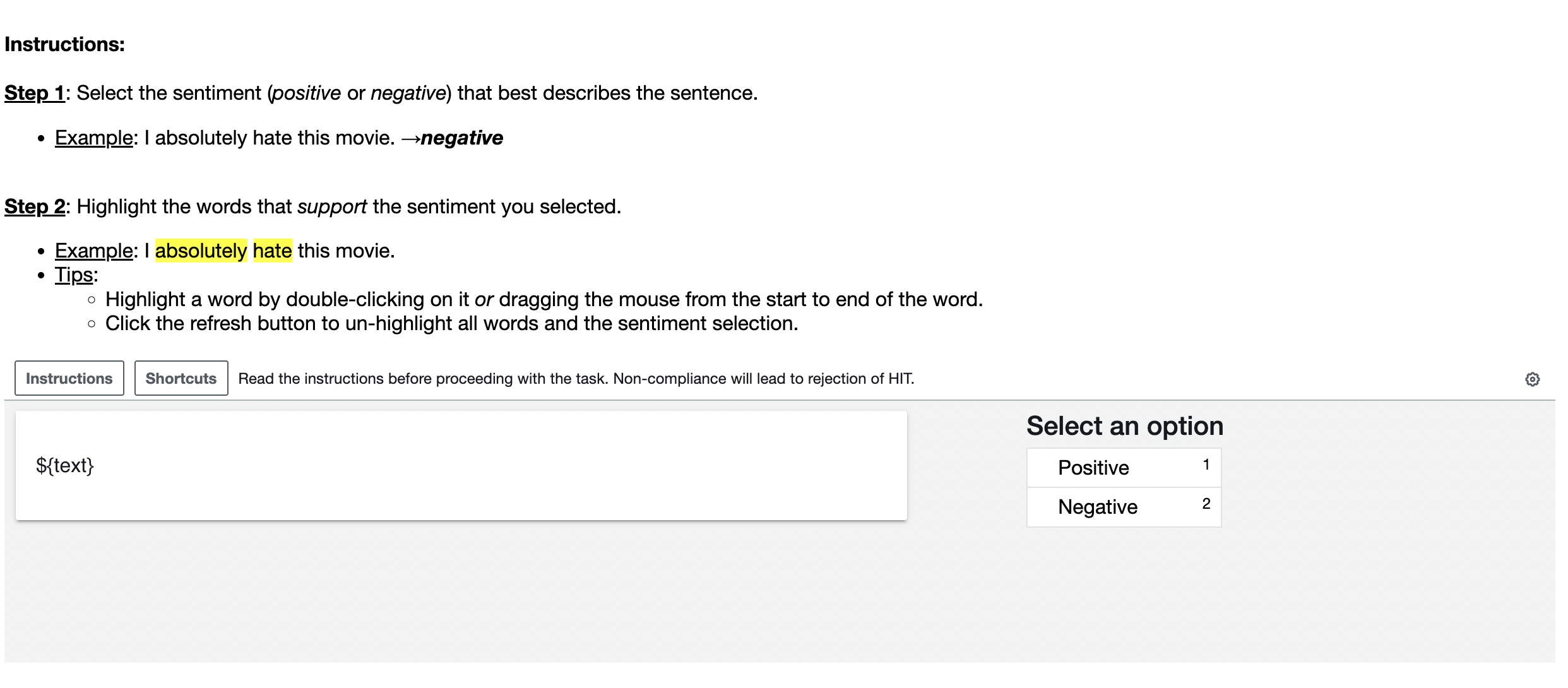}
    \caption{\textbf{RQ4 - Label+Expl UI.} UI used for Label+Expl MTurk annotations.}
    \label{fig:app:rq4:label+expl}
\end{figure*}

Figures \ref{fig:app:rq4:label}, \ref{fig:app:rq4:expl}, and \ref{fig:app:rq4:label+expl} display the user interfaces (UIs) provided to MTurk annotators in our time estimation experiments for Label Only, Expl Only, and Label+Expl, respectively.
All Turkers were paid at least \$16 per hour.
Based on rough a priori estimates of the annotation time per instance, we paid \$0.50 per instance for the qualifier task (\ie preliminary task used to select initial pool of 250 qualified Turkers), \$0.25 per instance for Label Only, \$0.30 per instance for Expl Only, \$0.35 per instance for Label+Expl.

To ensure that the annotation time estimates were based on high-quality annotations, we manually filtered out Turkers who submitted low-effort (\eg empty) responses. 
As a result, our time estimation experiments all yielded high inter-annotator agreement.
For Label Only and Label+Expl, we achieved Fleiss' kappa scores were $0.74$ and $0.70$, respectively.
For Expl Only and Label+Expl, we achieved rationale overlap rates \cite{zaidan2008modeling} of $0.78$ and $0.66$, respectively.
To further verify the quality of our MTurk results, we replicated these experiments in a small-scale study with nine computer science students and observed similar trends.
In this small-scale study, we considered the same 200 instances annotated by the Turkers, with each instance annotated by three students.



\subsection{Functional Tests}
\label{sec:app:functional}

There are four categories of functional tests: Vocabulary, Robustness, Logic, and Entity.
Below, we describe each category in more detail.
Also, each functional test category consists of one or more functional subtests.
However, for each category, \textsection \ref{sec:exp} only reported the aggregate performance across all subtests in the category.
Thus, Tables \ref{tab:app:functional:rq1:sent}-\ref{tab:app:functional:rq1:nli} also report the performance for all individual subtests, with respect to RQ1.

\paragraph{Vocabulary Tests}
Vocabulary tests are used to evaluate LMs' sensitivity to vocabulary changes, which may or may not change the text's meaning.
For sentiment analysis, we consider the following vocabulary tests: Add Sentiment Words, Paraphrase Neutral Words, Add Intensifiers, Add Reducers, Add Positive Phrases, and Add Negative Phrases \cite{ribeiro2020beyond}.
For NLI, we consider the following vocabulary tests: Antonym in Hypothesis, Synonym in Hypothesis, and Supertype in Hypothesis \cite{ribeiro2020beyond}.

\paragraph{Robustness Tests}
Robustness tests evaluate LMs' (in)sensitivity to character-level edits that should not change the text's meaning.
We consider the following robustness tests: Add Random URLs/Handles, Add Punctuation, Add One Typo, Add Two Typos, and Add Contractions \cite{jones-etal-2020-robust, wang-etal-2020-cat}.
For NLI, we consider the following vocabulary tests: Add Punctuation, Add One Typo, Add Two Typos, and Add Contractions \cite{jones-etal-2020-robust, wang-etal-2020-cat}.

\paragraph{Logic Tests} 
Logic tests evaluate LMs' sensitivity to perturbations that alter the logical semantics expressed by the text.
We consider the following logic tests: Positive $\rightarrow$ Negative, Negative $\rightarrow$ Positive, and Positive $\rightarrow$ Negative (w/ Distractors) \cite{talman2018testing, mccoy-etal-2019-right}.
For NLI, we consider the following vocabulary tests: Negate Hypothesis, Negate Premise, and Hypothesis is Premise \cite{talman2018testing, mccoy-etal-2019-right}.

\paragraph{Entity Tests}
Entity tests evaluate LMs' (in)sensitivity to entity changes that should not affect the text's meaning.
We consider the following entity tests: Replace Names, Replace Locations, and Replace Numbers \cite{ribeiro2020beyond}.
For NLI, we consider the following vocabulary tests: Replace Entity in Hypothesis \cite{ribeiro2020beyond}.

\paragraph{Results}
For each functional test category, we report the results for each functional subtest.
First, for sentiment analysis, we present the functional test/subtest for RQ1 (Table \ref{tab:app:functional:rq1:sent}), RQ2 (Table \ref{tab:app:functional:rq2}), RQ3 (Tables \ref{tab:app:functional:rq3_5}-\ref{tab:app:functional:rq3_50}), and RQ4 (Tables \ref{tab:app:functional:rq4_expl}-\ref{tab:app:functional:rq4_label_expl}). 
Second, for NLI, we present the functional test/subtest results for RQ1 (Table \ref{tab:app:functional:rq1:nli}).

\begin{table*}[t!]
\centering
\scalebox{0.55}{
\begin{tabular}{ccccccccc}
    \toprule
    \multirow{5}{*}{\textbf{Functional Test}} & \multirow{5}{*}{\textbf{Functional Subtest}} & \multicolumn{7}{c}{\textbf{Sentiment Analysis}}\\
    \cmidrule(lr){3-9}
    & & \multicolumn{7}{c}{Flights} \\
    \cmidrule(lr){3-9}
    & & \multicolumn{7}{c}{Failure Rate ($\downarrow$)} \\
    \cmidrule(lr){3-9}
    & & No-ER & IxG+MSE & IxG+MAE & IxG+Huber & IxG+BCE &  IxG+Order &IxG+KLDiv\\
\midrule
\multirow{8}{*}{Vocabulary} & Add Sentiment Words & 1.20~($\pm$0.74) & 0.60~($\pm$0.16) & 1.27~($\pm$0.84) & 1.13~($\pm$0.50) & 1.00~($\pm$0.86) &  0.80~($\pm$0.28) &\textbf{0.27} ~($\pm$0.25)\\
\cmidrule(lr){3-9}
& Paraphrase Neutral Words & 5.59~($\pm$0.16) & \textbf{5.13}~($\pm$0.90) & 5.40~($\pm$0.28) & 5.67~($\pm$0.74) &5.67~($\pm$0.68) &   5.60~($\pm$1.63) &5.87~($\pm$0.66)\\
\cmidrule(lr){3-9}
& Add Intensifiers & 2.13~($\pm$1.63) & 1.80~($\pm$0.16) & 1.40~($\pm$0.16) & 2.67~($\pm$0.96) & 2.67~($\pm$0.77) &  1.60~($\pm$0.65) & \textbf{1.27}~($\pm$0.19)\\
\cmidrule(lr){3-9}
& Add Reducers & 23.85~($\pm$7.18) & 35.00~($\pm$46.01) & 27.38~($\pm$5.95) & 17.46~($\pm$13.65) & 25.00~($\pm$25.00)  &  \textbf{0.77}~($\pm$0.43) &5.56~($\pm$7.86)\\
\cmidrule(lr){3-9}
& Add Positive Phrases & 1.40~($\pm$0.28) & 2.33~($\pm$1.84) & \textbf{0.67}~($\pm$0.50) & 2.33~($\pm$1.76) & 1.27~($\pm$1.00) &  2.07~($\pm$1.52) &1.07~($\pm$0.57) \\
\cmidrule(lr){3-9}
& Add Negative Phrases & 22.86~($\pm$7.43) & \textbf{14.80}~($\pm$1.40) & 20.67~($\pm$4.07) & 20.67~($\pm$3.35) & 17.40~($\pm$3.64) & 16.93~($\pm$1.91) & 16.67~($\pm$2.29)\\
\midrule
\multirow{7}{*}{Robustness} & Add Random URLs/Handles & 9.80~($\pm$0.48) & \textbf{7.27}~($\pm$2.23) & 9.07~($\pm$1.80) & 10.27~($\pm$0.9) & 7.87~($\pm$2.76) &  9.60~($\pm$2.47) & 9.60~($\pm$2.14)\\
\cmidrule(lr){3-9}
& Add Punctuation & 3.93~($\pm$0.89) & \textbf{1.93}~($\pm$0.41) & 3.00~($\pm$1.02) & 3.80~($\pm$0.28) &  2.87~($\pm$0.19) & 2.67~($\pm$0.34) & 2.67~($\pm$0.50)\\
\cmidrule(lr){3-9}
& Add One Typo & 2.60~($\pm$0.90) & 2.53~($\pm$0.82) & 2.60~($\pm$0.57) & 2.60~($\pm$0.75) & 3.13~($\pm$0.90) &  \textbf{2.00}~($\pm$0.86) &2.60~($\pm$0.43)\\
\cmidrule(lr){3-9}
& Add Two Typos & 3.93~($\pm$0.65) & 3.87~($\pm$1.24) & 4.27~($\pm$0.5) & 4.60~($\pm$0.43) & 4.13~($\pm$1.2) &  \textbf{3.33}~($\pm$0.25)&4.73~($\pm$0.25) \\
\cmidrule(lr){3-9}
& Add Contractions & 1.00~($\pm$0.00) & 0.80~($\pm$0.33) & 0.87~($\pm$0.25) & \textbf{0.47}~($\pm$0.09) & 0.80~($\pm$0.43) &  0.53~($\pm$0.50)&1.00~($\pm$0.16) \\
\midrule
\multirow{4}{*}{Logic} & Positive $\rightarrow$ Negative & 5.20~($\pm$2.75)  & 4.27~($\pm$1.65)  & 4.47~($\pm$3.07)  & \textbf{3.93}~($\pm$1.57) &  4.47~($\pm$1.75) & 5.67~($\pm$1.68)&4.93~($\pm$2.29)\\
\cmidrule(lr){3-9}
& Negative $\rightarrow$ Positive & 59.73~($\pm$9.48) & 59.00~($\pm$15.81)  & \textbf{37.47}~($\pm$10.41)  & 59.07~($\pm$14.97) & 63.27~($\pm$17.61) &  45.87~($\pm$24.13) &46.67~($\pm$18.75)\\
\cmidrule(lr){3-9}
& Positive $\rightarrow$ Negative (w/ Distractors) & 32.20~($\pm$14.65)  & 35.13~($\pm$1.91) & 35.00~($\pm$16.52)& 40.93~($\pm$4.31) & \textbf{19.00}~($\pm$8.66)  &  29.13~($\pm$10.60) & 38.00~($\pm$8.47)\\
\midrule
\multirow{4}{*}{Entity} & Replace Names &  \textbf{0.70}~($\pm$0.14 & 1.91~($\pm$0.71) & 1.11~($\pm$0.51) &  1.61~($\pm$0.62) & 0.81~($\pm$0.14) &  1.91~($\pm$1.51) &1.01~($\pm$0.75) \\
\cmidrule(lr){3-9}
& Replace Locations & 3.33~($\pm$0.74) & \textbf{2.73}~($\pm$1.15) & 3.40~($\pm$0.86) & 3.00~($\pm$0.33) & 3.07~($\pm$1.79) &  3.20~($\pm$1.57) & 3.53~($\pm$1.95)\\
\cmidrule(lr){3-9}
& Replace Numbers & 0.80~($\pm$0.00) & 0.53~($\pm$0.34) &  \textbf{0.47}~($\pm$0.41) & 0.60~($\pm$0.43) & 0.60~($\pm$0.33) &  0.67~($\pm$0.81)&0.87~($\pm$0.66)\\
\bottomrule 
\end{tabular}
}
\caption{\small \textbf{RQ1 - Functional Subtests for Sentiment Analysis (\textsection \ref{sec:exp:rq1:functional}).}
For sentiment analysis, we compare various ER \textit{rationale alignment criteria} using the IxG machine rationale extractor (as well as the No-ER baseline), with respect to performance on a range of functional tests/subtests (OOD). For each functional test, we report model performance on each of its individual functional subtests. Performance is reported in terms of failure rate. 
}
\label{tab:app:functional:rq1:sent}
\end{table*}

\begin{table*}[t!]
\centering
\scalebox{0.55}{
\begin{tabular}{cccccccc}
    \toprule
    \multirow{5}{*}{\textbf{Functional Test}} & \multirow{5}{*}{\textbf{Functional Subtest}} & \multicolumn{6}{c}{\textbf{NLI}}\\
    \cmidrule(lr){3-8}
    & & \multicolumn{6}{c}{ANLP-NLI} \\
    \cmidrule(lr){3-8}
    & & \multicolumn{6}{c}{Failure Rate ($\downarrow$)} \\
    \cmidrule(lr){3-8}
    & & No-ER & IxG+MSE & IxG+MAE & IxG+BCE & IxG+Huber & IxG+Order \\
\midrule
\multirow{4}{*}{Vocabulary} & Antonym in Hypothesis & 71.66~($\pm$20.98) & 64.77~($\pm$21.97) & 84.55~($\pm$11.53) & 65.88~($\pm$21.40) & 74.77~($\pm$20.41) & \textbf{62.55}~($\pm$13.16) \\
\cmidrule(lr){3-8}
& Synonym in Hypothesis & 32.61~($\pm$7.41) & \textbf{24.11}~($\pm$7.62) & 30.11~($\pm$6.42) & 25.88~($\pm$6.86) & 30.77~($\pm$7.07) & 29.27~($\pm$6.95) \\
\cmidrule(lr){3-8}
& Supertype in Hypothesis & 24.44~($\pm$15.95) & 11.00~($\pm$3.62) & 13.77~($\pm$6.71) & 9.31~($\pm$5.90) & \textbf{8.77}~($\pm$8.06) & 13.55~($\pm$7.10) \\
\midrule
\multirow{5}{*}{Robustness} & Add Punctuation & 14.55~($\pm$4.13) & 9.44~($\pm$2.79) & 11.33~($\pm$1.63) & \textbf{8.11}~($\pm$1.19) & 10.00~($\pm$2.58) & 9.88~($\pm$2.51) \\
\cmidrule(lr){3-8}
& Add One Typo & 15.88~($\pm$3.44) & 10.22~($\pm$3.04) & 12.33~($\pm$1.63) & \textbf{9.66}~($\pm$2.10) & 10.88~($\pm$2.68) & 10.77~($\pm$2.52) \\
\cmidrule(lr){3-8}
& Add Two Typos & 15.33~($\pm$3.68) & 9.77~($\pm$1.81) & 12.00~($\pm$1.76) & \textbf{9.44}~($\pm$2.31) & 11.11~($\pm$2.99) & 10.00~($\pm$2.66) \\
\cmidrule(lr){3-8}
& Add Contractions & 24.69~($\pm$6.98) & 24.69~($\pm$8.72) & 25.92~($\pm$9.07) & 22.22~($\pm$9.07) & 25.92~($\pm$7.40) & \textbf{14.81}~($\pm$5.23) \\
\midrule
\multirow{4}{*}{Logic} & Negate Hypothesis & 50.88~($\pm$32.25) & 27.77~($\pm$37.24) & \textbf{9.77}~($\pm$15.66) & 41.33~($\pm$41.54) & 15.22~($\pm$28.77) & 18.44~($\pm$23.21) \\
\cmidrule(lr){3-8}
& Negate Premise & 99.88~($\pm$0.31) & 98.54~($\pm$3.78) & 91.69~($\pm$20.37) & 98.65~($\pm$2.56) & \textbf{98.42}~($\pm$4.44) & 99.88~($\pm$0.31) \\
\cmidrule(lr){3-8}
& Hypothesis is Premise & \textbf{14.22}~($\pm$8.63) & 14.33~($\pm$10.14) & 19.44~($\pm$12.12) & 18.16~($\pm$12.69) & 14.38~($\pm$9.23) & 17.38~($\pm$10.16) \\
\midrule
\multirow{1}{*}{Entity} & Replace Entity in Hypothesis & \textbf{77.21}~($\pm$39.57) & 88.88~($\pm$24.11) & 79.91~($\pm$22.20) & 85.18~($\pm$30.04) & 83.83~($\pm$24.25) & 96.40~($\pm$4.85) \\
\bottomrule 
\end{tabular}
}
\caption{\small \textbf{RQ1 - Functional Subtests for NLI (\textsection \ref{sec:exp:rq1:functional}).}
For NLI, we compare various ER \textit{rationale alignment criteria} using the IxG machine rationale extractor (as well as the No-ER baseline), with respect to performance on a range of functional tests/subtests (OOD). For each functional test, we report model performance on each of its individual functional subtests. Performance is reported in terms of failure rate. 
}
\label{tab:app:functional:rq1:nli}
\end{table*}

\begin{table*}[t!]
\centering
\scalebox{0.55}{
\begin{tabular}{cccccc}
    \toprule
    \multirow{6}{*}{\textbf{Functional Test}} & \multirow{5}{*}{\textbf{Functional Subtest}} & \multicolumn{4}{c}{\textbf{Sentiment Analysis}}\\
    \cmidrule(lr){3-6}
    & & \multicolumn{4}{c}{Flights} \\
    \cmidrule(lr){3-6}
    & & \multicolumn{4}{c}{Failure Rate ($\downarrow$)} \\
    \cmidrule(lr){3-6}
    & & \multicolumn{2}{c}{IxG+MAE} & \multicolumn{2}{c}{IxG+Huber}\\
    \cmidrule(lr){3-4} \cmidrule(lr){5-6}
    & & Instance-Level & Task-Level & Instance-Level & Task-Level\\
\midrule
\multirow{8}{*}{Vocabulary} & Add Sentiment Words & \textbf{0.80}~($\pm$0.16) & 2.00~($\pm$0.82) & \textbf{1.13}~($\pm$0.50) & 1.27~($\pm$0.98)\\
\cmidrule(lr){3-6}
& Paraphrase Neutral Words & 5.87~($\pm$0.34) & \textbf{5.67}~($\pm$0.41) & \textbf{5.67}~($\pm$0.74) & 6.00~($\pm$0.59)\\
\cmidrule(lr){3-6}
& Add Intensifiers & 1.67~($\pm$0.41) & \textbf{1.60}~($\pm$0.49) & 2.67~($\pm$0.96) & \textbf{2.27}~($\pm$0.66) \\
\cmidrule(lr){3-6}
& Add Reducers & 55.03~($\pm$32.90) & \textbf{30.25}~($\pm$21.54) & \textbf{17.46}~($\pm$13.65) & 35.89~($\pm$1.61) \\
\cmidrule(lr){3-6}
& Add Positive Phrases & \textbf{0.60}~($\pm$0.33) & 1.47~($\pm$1.52) & 2.33~($\pm$1.76) & \textbf{0.67}~($\pm$0.57) \\
\cmidrule(lr){3-6}
& Add Negative Phrases & 20.47~($\pm$5.33) & \textbf{19.67}~($\pm$3.79) & \textbf{20.67}~($\pm$3.35) & 21.00~($\pm$5.94)  \\
\midrule
\multirow{7}{*}{Robustness} & Add Random URLs/Handles & 9.80~($\pm$0.48) & \textbf{7.27}~($\pm$2.23) & \textbf{10.27}~($\pm$0.90) & 10.40~($\pm$2.05)  \\
\cmidrule(lr){3-6}
& Add Punctuation & \textbf{2.07}~($\pm$0.25) & 4.00~($\pm$2.41) & 3.80~($\pm$0.28) & \textbf{3.40}~($\pm$1.34)  \\
\cmidrule(lr){3-6}
& Add One Typo & \textbf{2.40}~($\pm$0.86) & 2.47~($\pm$0.41) & \textbf{2.60}~($\pm$0.75) & 2.87~($\pm$0.25) \\
\cmidrule(lr){3-6}
& Add Two Typos & \textbf{3.87}~($\pm$0.94) & 4.47~($\pm$0.93) & 4.60~($\pm$0.43) & \textbf{4.33}~($\pm$0.52) \\
\cmidrule(lr){3-6}
& Add Contractions & \textbf{1.00}~($\pm$0.43) & 1.20~($\pm$0.43) & \textbf{0.47}~($\pm$0.09) & 0.87~($\pm$0.25) \\
\midrule
\multirow{4}{*}{Logic} & Positive $\rightarrow$ Negative & \textbf{4.60}~($\pm$1.88)  & 6.13~($\pm$1.82)  & 3.93~($\pm$1.57)  & \textbf{3.80}~($\pm$2.12) \\
\cmidrule(lr){3-6}
& Negative $\rightarrow$ Positive & 41.60~($\pm$21.11) & \textbf{40.87}~($\pm$20.03)  & 59.07~($\pm$14.97)  & \textbf{31.53}~($\pm$9.24)  \\
\cmidrule(lr){3-6}
& Positive $\rightarrow$ Negative (w/ Distractors) & \textbf{44.40}~($\pm$6.78)  & 49.80~($\pm$10.22) & 40.93~($\pm$4.31)& \textbf{32.13}~($\pm$13.47)  \\
\midrule
\multirow{4}{*}{Entity} & Replace Names &  \textbf{0.91}~($\pm$0.49 & 1.11~($\pm$0.38) & \textbf{1.61}~($\pm$0.62) &  1.91~($\pm$0.14)   \\
\cmidrule(lr){3-6}
& Replace Locations & \textbf{3.80}~($\pm$0.86) & 5.00~($\pm$1.84) & \textbf{3.00}~($\pm$0.33) & 4.07~($\pm$0.90)  \\
\cmidrule(lr){3-6}
& Replace Numbers & 0.87~($\pm$0.68) & \textbf{0.73}~($\pm$0.25) &  0.60~($\pm$0.43) & \textbf{0.53}~($\pm$0.19) \\
\bottomrule 
\end{tabular}
}
\caption{\small \textbf{RQ2 - Functional Subtests for Sentiment Analysis (\textsection \ref{sec:exp:rq1:functional}).}
For sentiment analysis, we compare various ER \textit{rationale alignment criteria} using the IxG machine rationale extractor (as well as the No-ER baseline), with respect to performance on a range of functional tests/subtests (OOD). For each functional test, we report model performance on each of its individual functional subtests. Performance is reported in terms of failure rate. 
}
\label{tab:app:functional:rq2}
\end{table*}

\begin{table*}[t!]
\centering
\scalebox{0.55}{
\begin{tabular}{ccccccccc}
    \toprule
    \multirow{5}{*}{\textbf{Functional Test}} & \multirow{5}{*}{\textbf{Functional Subtest}} & \multicolumn{7}{c}{\textbf{Sentiment Analysis}}\\
    \cmidrule(lr){3-9}
    & & \multicolumn{7}{c}{Flights} \\
    \cmidrule(lr){3-9}
    & & \multicolumn{7}{c}{Failure Rate ($\downarrow$)} \\
    \cmidrule(lr){3-9}
    & & 0\% & 100\% &Random & LC & HC & LIS & HIS \\
\midrule
\multirow{8}{*}{Vocabulary} & Add Sentiment Words & 1.20~($\pm$0.74) & 1.27~($\pm$0.84)  &1.09~($\pm$0.43)  & 2.00~($\pm$1.13)& \textbf{0.47}~($\pm$0.19) & 1.67~($\pm$0.74) &\textbf{0.47}~($\pm$0.25) \\
\cmidrule(lr){3-9}
& Paraphrase Neutral Words & 5.59~($\pm$0.16) & \textbf{5.40}~($\pm$0.28) &5.89~($\pm$1.25)   & 6.73~($\pm$2.04) &  7.07~($\pm$1.33) & 5.67~($\pm$0.68) &5.67~($\pm$0.84)  \\
\cmidrule(lr){3-9}
& Add Intensifiers & 2.13~($\pm$1.63) & \textbf{1.40}~($\pm$0.16) & 2.24~($\pm$0.53)  & 3.27~($\pm$1.52) & 1.93~($\pm$0.09) & 2.20~($\pm$0.43) &2.53~($\pm$0.90) \\
\cmidrule(lr){3-9}
& Add Reducers & 23.85~($\pm$7.18) & 27.38~($\pm$5.95) &45.42~($\pm$29.46)   & \textbf{20.96}~($\pm$11.10)  & 66.27~($\pm$24.67) & 42.59~($\pm$14.58) &43.73~($\pm$39.84) \\
\cmidrule(lr){3-9}
& Add Positive Phrases & 1.40~($\pm$0.28) & 0.67~($\pm$0.50) & 0.80~($\pm$0.43)  & 1.20~($\pm$0.28) & 0.80~($\pm$0.49) & \textbf{0.07}~($\pm$0.09)  &1.73~($\pm$1.39) \\
\cmidrule(lr){3-9}
& Add Negative Phrases & 22.86~($\pm$7.43) & 20.67~($\pm$4.07) & \textbf{20.47}~($\pm$0.50)  & 21.60~($\pm$6.65) &21.13~($\pm$4.72) & 24.80~($\pm$2.55) &23.33~($\pm$6.38) \\
\midrule
\multirow{7}{*}{Robustness} & Add Random URLs/Handles & 9.80~($\pm$0.48) & 9.07~($\pm$1.80)  &9.18~($\pm$1.98)  & 9.53~($\pm$1.00) &9.13~($\pm$0.98)  & 10.00~($\pm$1.13) &\textbf{9.00}~($\pm$3.85) \\
\cmidrule(lr){3-9}
& Add Punctuation & 3.93~($\pm$0.89) & \textbf{1.93}~($\pm$0.41) & 2.80~($\pm$1.07)  & 3.33~($\pm$0.82) & 2.60~($\pm$0.43) & 2.40~($\pm$0.49)&3.80~($\pm$1.73)  \\
\cmidrule(lr){3-9}
& Add One Typo & 2.60~($\pm$0.90) & 2.60~($\pm$0.57)  &2.49~($\pm$0.57) & 2.47~($\pm$0.34) & 2.80~($\pm$0.33) & \textbf{1.93}~($\pm$0.77)&2.33~($\pm$0.34)  \\
\cmidrule(lr){3-9}
& Add Two Typos & \textbf{3.93}~($\pm$0.65) & 4.27~($\pm$0.50)& 4.07~($\pm$0.54) & 4.20~($\pm$1.77) & 5.13~($\pm$0.34) & \textbf{3.93}~($\pm$0.52)&4.27~($\pm$0.09) \\
\cmidrule(lr){3-9}
& Add Contractions & 1.00~($\pm$0.00) & 0.87~($\pm$0.25) & 0.87~($\pm$0.28) & 1.00~($\pm$0.43) & 1.47~($\pm$0.96) & \textbf{0.80}~($\pm$0.16) &1.07~($\pm$0.25) \\
\midrule
\multirow{4}{*}{Logic} & Positive $\rightarrow$ Negative & 5.20~($\pm$2.75)  & 4.47~($\pm$3.07) & 6.60~($\pm$2.85)  & 7.13~($\pm$1.09) & 6.93~($\pm$2.1) & 7.27~($\pm$1.81) &\textbf{3.73}~($\pm$1.64) \\
\cmidrule(lr){3-9}
& Negative $\rightarrow$ Positive & 59.73~($\pm$9.48) & 37.47~($\pm$10.41) & 45.36~($\pm$23.87)  & 45.40~($\pm$17.25) & \textbf{32.87}~($\pm$7.25) & 38.87~($\pm$21.89) &64.93~($\pm$30.43) \\
\cmidrule(lr){3-9}
& Positive $\rightarrow$ Negative (w/ Distractors) & \textbf{32.20}~($\pm$14.65)  & 35.00~($\pm$16.52)& 47.89~($\pm$9.33)& 52.27~($\pm$7.17)  & 46.27~($\pm$17.01) & 48.07~($\pm$16.15) &40.47~($\pm$5.17) \\
\midrule
\multirow{4}{*}{Entity} & Replace Names &  0.70~($\pm$0.14)& 1.11~($\pm$0.51) & 1.07~($\pm$0.61) &  1.51~($\pm$0.25) & \textbf{0.50}~($\pm$0.14) & 0.81~($\pm$0.28) &1.31~($\pm$0.38)  \\
\cmidrule(lr){3-9}
& Replace Locations & \textbf{3.33}~($\pm$0.74) &3.40~($\pm$0.86) & 3.82~($\pm$1.57) & 4.20~($\pm$.91) & 4.53~($\pm$1.39) & 4.27~($\pm$1.51) &3.73~($\pm$1.76) \\
\cmidrule(lr){3-9}
& Replace Numbers & 0.80~($\pm$0.00) & \textbf{0.47}~($\pm$0.41) & 0.93~($\pm$0.55)  & 0.93~($\pm$0.19) & 1.27~($\pm$0.38) &  1.00~($\pm$0.33) &0.60~($\pm$0.16) \\
\bottomrule 
\end{tabular}
}
\caption{\small \textbf{RQ3 - Functional Subtests for Sentiment Analysis - $k$=5\% (\textsection \ref{sec:exp:rq1:functional}).}
For sentiment analysis, we compare various ER \textit{rationale alignment criteria} using the IxG machine rationale extractor (as well as the No-ER baseline), with respect to performance on a range of functional tests/subtests (OOD). For each functional test, we report model performance on each of its individual functional subtests. Performance is reported in terms of failure rate. 
}
\label{tab:app:functional:rq3_5}
\end{table*}

\begin{table*}[t!]
\centering
\scalebox{0.55}{
\begin{tabular}{ccccccccc}
    \toprule
    \multirow{5}{*}{\textbf{Functional Test}} & \multirow{5}{*}{\textbf{Functional Subtest}} & \multicolumn{7}{c}{\textbf{Sentiment Analysis}}\\
    \cmidrule(lr){3-9}
    & & \multicolumn{7}{c}{Flights} \\
    \cmidrule(lr){3-9}
    & & \multicolumn{7}{c}{Failure Rate ($\downarrow$)} \\
    \cmidrule(lr){3-9}
    & & 0\% & 100\% &Random & LC & HC & LIS & HIS \\
\midrule
\multirow{8}{*}{Vocabulary} & Add Sentiment Words & 1.20~($\pm$0.74) & 1.27~($\pm$0.84)  &0.98~($\pm$0.60)  & 1.20~($\pm$0.59)& \textbf{0.53}~($\pm$0.25) & 0.93~($\pm$0.25) &0.87~($\pm$0.09) \\
\cmidrule(lr){3-9}
& Paraphrase Neutral Words & 5.59~($\pm$0.16) & \textbf{5.40}~($\pm$0.28) &6.60~($\pm$1.35)   & 6.27~($\pm$1.36) &  5.93~($\pm$0.98) & 5.47~($\pm$0.94) &6.07~($\pm$0.74)  \\
\cmidrule(lr){3-9}
& Add Intensifiers & 2.13~($\pm$1.63) & \textbf{1.40}~($\pm$0.16) & 2.07~($\pm$1.12)  & 2.67~($\pm$0.34) & 1.93~($\pm$0.25) & 1.53~($\pm$0.41) &2.40~($\pm$0.71) \\
\cmidrule(lr){3-9}
& Add Reducers & 23.85~($\pm$7.18) & 27.38~($\pm$5.95) &42.96~($\pm$29.27)   & 34.97~($\pm$27.96)  & \textbf{16.67}~($\pm$16.67) & 42.86~($\pm$20.2) &35.35~($\pm$24.78) \\
\cmidrule(lr){3-9}
& Add Positive Phrases & 1.40~($\pm$0.28) & 0.67~($\pm$0.50) & \textbf{0.49}~($\pm$0.59)  & 1.07~($\pm$0.66) & 0.53~($\pm$0.50) & 0.80~($\pm$0.71)  &1.73~($\pm$1.61) \\
\cmidrule(lr){3-9}
& Add Negative Phrases & 22.86~($\pm$7.43) & 20.67~($\pm$4.07) & 23.51~($\pm$3.31)  & 18.47~($\pm$0.90) &17.20~($\pm$4.96) & 20.00~($\pm$4.26) &\textbf{12.33}~($\pm$2.02) \\
\midrule
\multirow{7}{*}{Robustness} & Add Random URLs/Handles & 9.80~($\pm$0.48) & 9.07~($\pm$1.80)  &8.93~($\pm$1.31)  & \textbf{8.20}~($\pm$2.29) & 8.87~($\pm$1.37) & 8.53~($\pm$2.92) &9.20~($\pm$2.75) \\
\cmidrule(lr){3-9}
& Add Punctuation & 3.93~($\pm$0.89) & \textbf{1.93}~($\pm$0.41) & 2.89~($\pm$0.89)  & 3.27~($\pm$2.39) & 3.53~($\pm$0.84) & 2.87~($\pm$0.34)&3.20~($\pm$0.71)  \\
\cmidrule(lr){3-9}
& Add One Typo & 2.60~($\pm$0.90) & 2.60~($\pm$0.57)  &2.56~($\pm$0.55) & 2.40~($\pm$0.98) & 3.27~($\pm$0.25) & \textbf{2.07}~($\pm$0.19)&2.73~($\pm$0.34)  \\
\cmidrule(lr){3-9}
& Add Two Typos & 3.93~($\pm$0.65) & 4.27~($\pm$0.50)& 4.60~($\pm$0.49) & 3.93~($\pm$0.50) & \textbf{3.67}~($\pm$0.66) & 4.27~($\pm$1.36)&3.87~($\pm$0.25) \\
\cmidrule(lr){3-9}
& Add Contractions & 1.00~($\pm$0.00) & 0.87~($\pm$0.25) & 1.29~($\pm$0.19) & \textbf{0.80}~($\pm$0.33) & 1.33~($\pm$0.57) & 0.93~($\pm$0.25) &1.07~($\pm$0.34) \\
\midrule
\multirow{4}{*}{Logic} & Positive $\rightarrow$ Negative & 5.20~($\pm$2.75)  & 4.47~($\pm$3.07) & 7.78~($\pm$2.17)  & 7.20~($\pm$1.85) & 7.80~($\pm$0.99) & 7.93~($\pm$1.65) &\textbf{3.27}~($\pm$1.91) \\
\cmidrule(lr){3-9}
& Negative $\rightarrow$ Positives & 59.73~($\pm$9.48) &\textbf{37.47}~($\pm$10.41) & 53.62~($\pm$24.45)  & 43.67~($\pm$34.97) & 59.93~($\pm$27.40) & 49.67~($\pm$12.45) &66.13~($\pm$22.82) \\
\cmidrule(lr){3-9}
& Positive $\rightarrow$ Negative (w/ Distractors) & \textbf{32.20}~($\pm$14.65)  & 35.00~($\pm$16.52)& 49.29~($\pm$13.80)& 33.53~($\pm$13.96)  & 53.20~($\pm$11.44) & 57.20~($\pm$7.69) &38.60~($\pm$14.01) \\
\midrule
\multirow{4}{*}{Entity} & Replace Names &  \textbf{0.70}~($\pm$0.14)& 1.11~($\pm$0.51) & 1.01~($\pm$0.74) &  \textbf{0.70}~($\pm$1.00) & 1.21~($\pm$0.25) & 1.11~($\pm$0.38) &1.61~($\pm$0.57)  \\
\cmidrule(lr){3-9}
& Replace Locations & 3.33~($\pm$0.74) &3.40~($\pm$0.86) & 4.42~($\pm$1.13) & 4.87~($\pm$1.89) & 3.73~($\pm$1.73) & \textbf{3.20}~($\pm$0.75) &3.40~($\pm$0.33) \\
\cmidrule(lr){3-9}
& Replace Numbers & 0.80~($\pm$0.00) & \textbf{0.47}~($\pm$0.41) & 1.09~($\pm$0.35)  & 1.20~($\pm$0.65) & 1.00~($\pm$0.16) &  0.73~($\pm$0.34) &0.67~($\pm$0.25) \\
\bottomrule 
\end{tabular}
}
\caption{\small \textbf{RQ3 - Functional Subtests for Sentiment Analysis - $k$=15\% (\textsection \ref{sec:exp:rq1:functional}).}
For sentiment analysis, we compare various ER \textit{rationale alignment criteria} using the IxG machine rationale extractor (as well as the No-ER baseline), with respect to performance on a range of functional tests/subtests (OOD). For each functional test, we report model performance on each of its individual functional subtests. Performance is reported in terms of failure rate. 
}
\label{tab:app:functional:rq3_15}
\end{table*}

\begin{table*}[t!]
\centering
\scalebox{0.55}{
\begin{tabular}{ccccccccc}
    \toprule
    \multirow{5}{*}{\textbf{Functional Test}} & \multirow{5}{*}{\textbf{Functional Subtest}} & \multicolumn{7}{c}{\textbf{Sentiment Analysis}}\\
    \cmidrule(lr){3-9}
    & & \multicolumn{7}{c}{Flights} \\
    \cmidrule(lr){3-9}
    & & \multicolumn{7}{c}{Failure Rate ($\downarrow$)} \\
    \cmidrule(lr){3-9}
    & & 0\% & 100\% &Random & LC & HC & LIS & HIS \\
\midrule
\multirow{8}{*}{Vocabulary} & Add Sentiment Words & 1.20~($\pm$0.74) & 1.27~($\pm$0.84)  &\textbf{1.02}~($\pm$0.55)  & 3.93~($\pm$2.12)& 2.27~($\pm$1.43) & 1.33~($\pm$0.09) &1.60~($\pm$1.13) \\
\cmidrule(lr){3-9}
& Paraphrase Neutral Words & 5.59~($\pm$0.16) & 5.40~($\pm$0.28) &5.98~($\pm$0.91)   & 9.73~($\pm$0.41) &  5.93~($\pm$0.98) & 4.87~($\pm$0.50) &\textbf{4.53}~($\pm$0.19)  \\
\cmidrule(lr){3-9}
& Add Intensifiers & 2.13~($\pm$1.63) & \textbf{1.40}~($\pm$0.16) & 2.07~($\pm$1.50)  & 3.27~($\pm$1.67) & 1.93~($\pm$0.25) & 2.60~($\pm$0.57) &3.53~($\pm$1.68) \\
\cmidrule(lr){3-9}
& Add Reducers & 23.85~($\pm$7.18) & 27.38~($\pm$5.95) &20.32~($\pm$19.45)   & 41.86~($\pm$7.54)  & 15.81~($\pm$2.82) & 27.86~($\pm$9.17) &\textbf{15.20}~($\pm$11.38) \\
\cmidrule(lr){3-9}
& Add Positive Phrases & 1.40~($\pm$0.28) & \textbf{0.67}~($\pm$0.50) & 1.87~($\pm$2.16)  & \textbf{0.67}~($\pm$0.41) & 1.00~($\pm$0.33) & 4.67~($\pm$1.81)  &6.73~($\pm$6.74) \\
\cmidrule(lr){3-9}
& Add Negative Phrases & 22.86~($\pm$7.43) & 20.67~($\pm$4.07) & 25.38~($\pm$8.40)  & \textbf{19.93}~($\pm$5.66) &21.47~($\pm$9.29) & 23.73~($\pm$3.14) &24.47~($\pm$15.92) \\
\midrule
\multirow{7}{*}{Robustness} & Add Random URLs/Handles & 9.80~($\pm$0.48) & \textbf{9.07}~($\pm$1.80)  &10.00~($\pm$3.02)  & 11.27~($\pm$3.36) & 12.27~($\pm$4.34) & 9.67~($\pm$1.39) &9.40~($\pm$7.64) \\
\cmidrule(lr){3-9}
& Add Punctuation & 3.93~($\pm$0.89) & \textbf{1.93}~($\pm$0.41) & 4.11~($\pm$1.75)  & 4.67~($\pm$1.67) & 3.27~($\pm$0.84) & 4.33~($\pm$0.66)&3.80~($\pm$2.83)  \\
\cmidrule(lr){3-9}
& Add One Typo & 2.60~($\pm$0.90) & 2.60~($\pm$0.57)  &2.89~($\pm$0.39) & 4.07~($\pm$1.16) & 2.40~($\pm$0.28) & \textbf{2.13}~($\pm$0.57)&3.00~($\pm$0.59)  \\
\cmidrule(lr){3-9}
& Add Two Typos & 3.93~($\pm$0.65) & 4.27~($\pm$0.50)& 4.56~($\pm$1.12) & 6.33~($\pm$1.20) & 5.47~($\pm$1.39) & \textbf{3.47}~($\pm$0.84)&4.40~($\pm$0.59) \\
\cmidrule(lr){3-9}
& Add Contractions & 1.00~($\pm$0.00) & 0.87~($\pm$0.25) & 1.00~($\pm$0.28) & 1.13~($\pm$0.41) & 1.00~($\pm$0.28) & 0.93~($\pm$0.09) &\textbf{0.40}~($\pm$0.16) \\
\midrule
\multirow{4}{*}{Logic} & Positive $\rightarrow$ Negative & 5.20~($\pm$2.75)  & 4.47~($\pm$3.07) & 6.09~($\pm$1.47)  & 10.13~($\pm$1.64) & 8.00~($\pm$0.85) & 3.80~($\pm$0.65) &\textbf{2.00}~($\pm$0.75) \\
\cmidrule(lr){3-9}
& Negative $\rightarrow$ Positives & 59.73~($\pm$9.48) & \textbf{37.47}~($\pm$10.41) & 68.09~($\pm$22.22)  & 41.73~($\pm$25.73) & 59.93~($\pm$27.40) & 69.80~($\pm$11.61) &85.27~($\pm$10.52) \\
\cmidrule(lr){3-9}
& Positive $\rightarrow$ Negative (w/ Distractors) & 32.20~($\pm$14.65)  & 35.00~($\pm$16.52)& 51.11~($\pm$12.86)& 53.53~($\pm$7.48)  & 56.07~($\pm$9.26) & 46.80~($\pm$2.21) &\textbf{22.87}~($\pm$7.56) \\
\midrule
\multirow{4}{*}{Entity} & Replace Names &  \textbf{0.70}~($\pm$0.14)& 1.11~($\pm$0.51) & 1.48~($\pm$0.79) &  1.61~($\pm$0.87) & 1.01~($\pm$0.14) & \textbf{0.70}~($\pm$0.38) &1.81~($\pm$1.08)  \\
\cmidrule(lr){3-9}
& Replace Locations & 3.33~($\pm$0.74) &3.40~($\pm$0.86) & 5.00~($\pm$1.04) & 6.27~($\pm$1.73) & 5.07~($\pm$1.37) & 3.67~($\pm$1.09) &\textbf{3.00}~($\pm$2.14) \\
\cmidrule(lr){3-9}
& Replace Numbers & 0.80~($\pm$0.00) & \textbf{0.47}~($\pm$0.41) & 1.27~($\pm$0.45)  & 1.87~($\pm$0.57) & 1.40~($\pm$0.00) &  0.80~($\pm$0.43) &0.93~($\pm$0.41) \\
\bottomrule 
\end{tabular}
}
\caption{\small \textbf{RQ3 - Functional Subtests for Sentiment Analysis - $k$=50\% (\textsection \ref{sec:exp:rq1:functional}).}
For sentiment analysis, we compare various ER \textit{rationale alignment criteria} using the IxG machine rationale extractor (as well as the No-ER baseline), with respect to performance on a range of functional tests/subtests (OOD). For each functional test, we report model performance on each of its individual functional subtests. Performance is reported in terms of failure rate. 
}
\label{tab:app:functional:rq3_50}
\end{table*}

\begin{table*}[t!]
\centering
\scalebox{0.55}{
\begin{tabular}{cccccccc}
    \toprule
    \multirow{5}{*}{\textbf{Functional Test}} & \multirow{5}{*}{\textbf{Functional Subtest}} & \multicolumn{6}{c}{\textbf{Sentiment Analysis}}\\
    \cmidrule(lr){3-8}
    & & \multicolumn{6}{c}{Flights} \\
    \cmidrule(lr){3-8}
    & & \multicolumn{6}{c}{Failure Rate ($\downarrow$)} \\
    \cmidrule(lr){3-8}
    & & 0 min & 10 min & 30 min & 5 hr & 24 hr &  48 hr \\
\midrule
\multirow{8}{*}{Vocabulary} & Add Sentiment Words & 2.22~($\pm$1.58) & 3.00~($\pm$4.79) & 1.27~($\pm$0.84) & 1.29~($\pm$0.80) & \textbf{0.85}~($\pm$0.85) &  1.11~($\pm$0.70) \\
\cmidrule(lr){3-8}
& Paraphrase Neutral Words & 4.91~($\pm$1.66) & 6.15~($\pm$1.52) & 5.40~($\pm$0.28) & \textbf{4.27}~($\pm$0.93) &5.52~($\pm$1.02) &   6.46~($\pm$0.89) \\
\cmidrule(lr){3-8}
& Add Intensifiers & 4.40~($\pm$2.05) & 4.18~($\pm$3.70) & \textbf{1.40}~($\pm$0.16) & 3.76~($\pm$1.34) & 2.75~($\pm$1.26) &  2.34~($\pm$0.42) \\
\cmidrule(lr){3-8}
& Add Reducers & 21.68~($\pm$16.20) & \textbf{13.11}~($\pm$11.58) & 27.38~($\pm$5.95) & 18.61~($\pm$18.37) & 25.78~($\pm$32.43)  &  28.25~($\pm$14.76) \\
\cmidrule(lr){3-8}
& Add Positive Phrases & 2.80~($\pm$2.56) & 4.80~($\pm$4.04) & \textbf{0.67}~($\pm$0.50) & 3.27~($\pm$2.08) & 2.05~($\pm$1.94) &  1.06~($\pm$1.01)  \\
\cmidrule(lr){3-8}
& Add Negative Phrases & 20.31~($\pm$11.60) & 23.78~($\pm$12.68) & 20.67~($\pm$4.07) & 16.44~($\pm$7.70) & \textbf{14.65}~($\pm$3.36) & 16.17~($\pm$4.42) \\
\midrule
\multirow{7}{*}{Robustness} & Add Random URLs/Handles & 8.82~($\pm$4.95) & 9.85~($\pm$4.57) & 9.07~($\pm$1.80) & \textbf{6.87}~($\pm$3.45) & 7.65~($\pm$2.79) &  7.66~($\pm$1.83) \\
\cmidrule(lr){3-8}
& Add Punctuation & 3.73~($\pm$2.68) & 4.45~($\pm$3.51) & 3.00~($\pm$1.02) & \textbf{2.31}~($\pm$1.22) &  2.55~($\pm$1.40) & 2.77~($\pm$0.92) \\
\cmidrule(lr){3-8}
& Add One Typo & \textbf{2.09}~($\pm$0.60) & 2.60~($\pm$0.77) & 2.60~($\pm$0.57) & 2.18~($\pm$0.46) & 2.68~($\pm$0.45) &  2.37~($\pm$0.61) \\
\cmidrule(lr){3-8}
& Add Two Typos & 4.56~($\pm$1.79) & 5.00~($\pm$0.78) & 4.27~($\pm$0.5) & \textbf{3.91}~($\pm$0.99) & 4.40~($\pm$1.03) &  4.20~($\pm$0.88) \\
\cmidrule(lr){3-8}
& Add Contractions & 0.80~($\pm$0.52) & 0.72~($\pm$0.22) & 0.87~($\pm$0.25) & \textbf{0.56}~($\pm$0.26) & 0.65~($\pm$0.37) &  0.77~($\pm$0.43) \\
\midrule
\multirow{4}{*}{Logic} & Positive $\rightarrow$ Negative & 5.04~($\pm$4.09)  & 4.62~($\pm$2.39)  & 4.47~($\pm$3.07)  & \textbf{2.93}~($\pm$1.38) &  5.10~($\pm$2.64) & 5.00~($\pm$2.27)\\
\cmidrule(lr){3-8}
& Negative $\rightarrow$ Positive & 84.56~($\pm$14.41) & 64.03~($\pm$25.19)  & \textbf{37.47}~($\pm$10.41)  & 81.29~($\pm$13.66) & 74.53~($\pm$17.39) &  49.94~($\pm$28.02) \\
\cmidrule(lr){3-8}
& Positive $\rightarrow$ Negative (w/ Distractors) & 30.40~($\pm$21.11)  & 34.20~($\pm$17.76) & 35.00~($\pm$16.52)& \textbf{25.53}~($\pm$18.38) & 32.88~($\pm$15.44)  &  38.20~($\pm$14.73) \\
\midrule
\multirow{4}{*}{Entity} & Replace Names &  1.28~($\pm$0.91) & 1.96~($\pm$1.33) & 1.11~($\pm$0.51) &  1.11~($\pm$0.38) & 1.02~($\pm$0.45) &  \textbf{0.91}~($\pm$0.65)  \\
\cmidrule(lr){3-8}
& Replace Locations & 3.47~($\pm$1.54) & 3.20~($\pm$1.76) & 3.40~($\pm$0.86) & 2.42~($\pm$1.31) & \textbf{3.05}~($\pm$1.12) &  2.89~($\pm$1.26) \\
\cmidrule(lr){3-8}
& Replace Numbers & 0.58~($\pm$0.60) & 0.75~($\pm$0.21) &  0.47~($\pm$0.41) & \textbf{0.38}~($\pm$0.37) & 0.42~($\pm$0.43) &  0.69~($\pm$0.57)\\
\bottomrule 
\end{tabular}
}
\caption{\small \textbf{RQ4 - Functional Subtests for Sentiment Analysis - Label Only (\textsection \ref{sec:exp:rq1:functional}).}
For sentiment analysis, we compare various ER \textit{rationale alignment criteria} using the IxG machine rationale extractor (as well as the No-ER baseline), with respect to performance on a range of functional tests/subtests (OOD). For each functional test, we report model performance on each of its individual functional subtests. Performance is reported in terms of failure rate. 
}
\label{tab:app:functional:rq4_label}
\end{table*}

\begin{table*}
\centering
\scalebox{0.55}{
\begin{tabular}{cccccccc}
    \toprule
    \multirow{5}{*}{\textbf{Functional Test}} & \multirow{5}{*}{\textbf{Functional Subtest}} & \multicolumn{6}{c}{\textbf{Sentiment Analysis}}\\
    \cmidrule(lr){3-8}
    & & \multicolumn{6}{c}{Flights} \\
    \cmidrule(lr){3-8}
    & & \multicolumn{6}{c}{Failure Rate ($\downarrow$)} \\
    \cmidrule(lr){3-8}
    & & 0 min & 10 min & 30 min & 5 hr & 24 hr &  48 hr \\
\midrule
\multirow{8}{*}{Vocabulary} & Add Sentiment Words & 2.22~($\pm$1.58) & \textbf{0.91}~($\pm$0.58) & 1.16~($\pm$0.62) & 2.07~($\pm$1.25) & 0.92~($\pm$0.52) & 0.92~($\pm$0.52) \\
\cmidrule(lr){3-8}
& Paraphrase Neutral Words & 4.91~($\pm$1.66) & 5.27~($\pm$1.47) & \textbf{4.82}~($\pm$1.05) & 5.47~($\pm$2.25) &5.82~($\pm$2.47) & 5.82~($\pm$2.47)   \\
\cmidrule(lr){3-8}
& Add Intensifiers & 4.40~($\pm$2.05) & 2.80~($\pm$1.19) & 2.76~($\pm$0.90) & 3.04~($\pm$1.09) & \textbf{2.00}~($\pm$1.11) &  \textbf{2.00}~($\pm$1.11)  \\
\cmidrule(lr){3-8}
& Add Reducers & 21.68~($\pm$16.20) & 11.06~($\pm$9.21) & \textbf{8.64}~($\pm$11.13) & 18.38~($\pm$10.12) & 29.43~($\pm$19.41)  &  29.43~($\pm$19.41) \\
\cmidrule(lr){3-8}
& Add Positive Phrases & 2.80~($\pm$2.56) & 2.49~($\pm$1.66) & 2.76~($\pm$1.80) & 1.76~($\pm$1.18) & \textbf{1.52}~($\pm$0.62) & \textbf{1.52}~($\pm$0.62) \\
\cmidrule(lr){3-8}
& Add Negative Phrases & 20.31~($\pm$11.60) & 18.51~($\pm$9.87) & 21.49~($\pm$8.73) & \textbf{16.76}~($\pm$5.49) & 19.45~($\pm$8.40) & 19.45~($\pm$8.40) \\
\midrule
\multirow{7}{*}{Robustness} & Add Random URLs/Handles & 8.82~($\pm$4.95) & 7.80~($\pm$3.71) & 7.69~($\pm$3.75) & 8.13~($\pm$2.99) & \textbf{7.15}~($\pm$4.06) &  \textbf{7.15}~($\pm$4.06) \\
\cmidrule(lr){3-8}
& Add Punctuation & 3.73~($\pm$2.68) & \textbf{2.84}~($\pm$1.37) & 3.42~($\pm$1.85) & 3.20~($\pm$1.48) &  3.65~($\pm$2.75) & 3.65~($\pm$2.75) \\
\cmidrule(lr){3-8}
& Add One Typo & \textbf{2.09}~($\pm$0.60) & 2.56~($\pm$0.67) & 2.51~($\pm$0.82) & 2.62~($\pm$1.01) & 2.78~($\pm$0.64) &  2.78~($\pm$0.64) \\
\cmidrule(lr){3-8}
& Add Two Typos & 4.56~($\pm$1.79) & 4.76~($\pm$1.11) & \textbf{4.09}~($\pm$0.96) & 4.49~($\pm$1.18) & 4.65~($\pm$1.02) &  4.65~($\pm$1.02) \\
\cmidrule(lr){3-8}
& Add Contractions & 0.80~($\pm$0.52) & 0.73~($\pm$0.46) & \textbf{0.56}~($\pm$0.53) & 0.73~($\pm$0.65) & 0.68~($\pm$0.33) &  0.68~($\pm$0.33) \\
\midrule
\multirow{4}{*}{Logic} & Positive $\rightarrow$ Negative & 5.04~($\pm$4.09)  & \textbf{3.36}~($\pm$1.30)  & 3.44~($\pm$1.50)  & 5.91~($\pm$3.16) &  5.98~($\pm$2.96) & 5.98~($\pm$2.96)\\
\cmidrule(lr){3-8}
& Negative $\rightarrow$ Positive & 84.56~($\pm$14.41) & 75.89~($\pm$26.78)  & 85.89~($\pm$12.74)  & \textbf{56.98}~($\pm$34.69) & 58.12~($\pm$27.91) &  58.12~($\pm$27.91) \\
\cmidrule(lr){3-8}
& Positive $\rightarrow$ Negative (w/ Distractors) & 30.40~($\pm$21.11)  & 28.38~($\pm$8.88) & \textbf{24.91}~($\pm$11.84)& 40.73~($\pm$23.29) & 46.75~($\pm$20.49)  &  46.75~($\pm$20.49) \\
\midrule
\multirow{4}{*}{Entity} & Replace Names &  1.28~($\pm$0.91) & 1.17~($\pm$0.97) & 1.28~($\pm$1.03) &  \textbf{1.01}~($\pm$0.55) & 1.21~($\pm$0.52) &  1.21~($\pm$0.52) \\
\cmidrule(lr){3-8}
& Replace Locations & 3.47~($\pm$1.54) & \textbf{2.69}~($\pm$1.49) & 2.87~($\pm$1.36) & 3.56~($\pm$2.05) & 3.18~($\pm$2.31) &  3.18~($\pm$2.31) \\
\cmidrule(lr){3-8}
& Replace Numbers & 0.58~($\pm$0.60) & 0.44~($\pm$0.39) &  \textbf{0.31}~($\pm$0.21) & 0.69~($\pm$0.58) & 1.02~($\pm$0.87) &  1.02~($\pm$0.87)\\
\bottomrule 
\end{tabular}
}
\caption{\small \textbf{RQ4 - Functional Subtests for Sentiment Analysis - Expl Only (\textsection \ref{sec:exp:rq1:functional}).}
For sentiment analysis, we compare various ER \textit{rationale alignment criteria} using the IxG machine rationale extractor (as well as the No-ER baseline), with respect to performance on a range of functional tests/subtests (OOD). For each functional test, we report model performance on each of its individual functional subtests. Performance is reported in terms of failure rate. 
}
\label{tab:app:functional:rq4_expl}
\end{table*}

\begin{table*}
\centering
\scalebox{0.55}{
\begin{tabular}{cccccccc}
    \toprule
    \multirow{5}{*}{\textbf{Functional Test}} & \multirow{5}{*}{\textbf{Functional Subtest}} & \multicolumn{6}{c}{\textbf{Sentiment Analysis}}\\
    \cmidrule(lr){3-8}
    & & \multicolumn{6}{c}{Flights} \\
    \cmidrule(lr){3-8}
    & & \multicolumn{6}{c}{Failure Rate ($\downarrow$)} \\
    \cmidrule(lr){3-8}
    & & 0 min & 10 min & 30 min & 5 hr & 24 hr &  48 hr \\
\midrule
\multirow{8}{*}{Vocabulary} & Add Sentiment Words & 2.22~($\pm$1.58) & 2.07~($\pm$2.44) & 2.84~($\pm$4.05) & 1.38~($\pm$0.73) & 1.53~($\pm$1.15) &  \textbf{1.07}~($\pm$0.83) \\
\cmidrule(lr){3-8}
& Paraphrase Neutral Words & 4.91~($\pm$1.66) & 6.18~($\pm$2.36) & 4.80~($\pm$1.98) & 5.27~($\pm$1.56) &5.36~($\pm$1.20) &   \textbf{4.78}~($\pm$1.14) \\
\cmidrule(lr){3-8}
& Add Intensifiers & 4.40~($\pm$2.05) & 3.00~($\pm$2.20) & 3.51~($\pm$3.21) & 4.04~($\pm$4.98) & 2.89~($\pm$1.08) &  \textbf{2.07}~($\pm$0.85) \\
\cmidrule(lr){3-8}
& Add Reducers & 21.68~($\pm$16.20) & 22.08~($\pm$29.84) & \textbf{16.55}~($\pm$15.31) & 22.27~($\pm$16.00) & 22.68~($\pm$11.99)  &  46.09~($\pm$19.50) \\
\cmidrule(lr){3-8}
& Add Positive Phrases & 2.80~($\pm$2.56) & 2.78~($\pm$2.22) & 1.44~($\pm$1.04) & 3.11~($\pm$2.83) & 1.91~($\pm$1.60) &  \textbf{1.40}~($\pm$1.10)  \\
\cmidrule(lr){3-8}
& Add Negative Phrases & 20.31~($\pm$11.60) & 19.07~($\pm$9.48) & \textbf{16.60}~($\pm$6.60) & 18.84~($\pm$17.92) & 25.87~($\pm$14.51) & 17.98~($\pm$3.68) \\
\midrule
\multirow{7}{*}{Robustness} & Add Random URLs/Handles & 8.82~($\pm$4.95) & 8.78~($\pm$4.78) & \textbf{7.27}~($\pm$2.48) & 9.02~($\pm$7.59) & 10.02~($\pm$5.74) &  8.44~($\pm$2.37) \\
\cmidrule(lr){3-8}
& Add Punctuation & 3.73~($\pm$2.68) & 3.22~($\pm$1.58) & 3.27~($\pm$2.24) & 3.91~($\pm$5.02) &  3.93~($\pm$2.24) & \textbf{2.38}~($\pm$0.71) \\
\cmidrule(lr){3-8}
& Add One Typo & \textbf{2.09}~($\pm$0.60) & 2.62~($\pm$0.73) & 2.42~($\pm$0.86) & 3.02~($\pm$1.22) & 2.44~($\pm$1.06) &  2.22~($\pm$0.64) \\
\cmidrule(lr){3-8}
& Add Two Typos & 4.56~($\pm$1.79) & 4.62~($\pm$0.89) & \textbf{3.84}~($\pm$1.49) & 4.58~($\pm$1.85) & 4.36~($\pm$1.14) &  3.98~($\pm$0.60) \\
\cmidrule(lr){3-8}
& Add Contractions & 0.80~($\pm$0.52) & 0.82~($\pm$0.38) & \textbf{0.62}~($\pm$0.57) & 0.82~($\pm$0.63) & 0.80~($\pm$0.23) &  0.73~($\pm$0.25) \\
\midrule
\multirow{4}{*}{Logic} & Positive $\rightarrow$ Negative & 5.04~($\pm$4.09)  & 8.31~($\pm$10.47)  & 4.47~($\pm$3.07)  & \textbf{4.13}~($\pm$2.99) &  4.84~($\pm$3.00) & 5.82~($\pm$2.81)\\
\cmidrule(lr){3-8}
& Negative $\rightarrow$ Positive & 84.56~($\pm$14.41) & 65.76~($\pm$21.49)  & 68.91~($\pm$28.29)  & 81.29~($\pm$13.66) & 70.98~($\pm$18.14) &  \textbf{63.73}~($\pm$22.93) \\
\cmidrule(lr){3-8}
& Positive $\rightarrow$ Negative (w/ Distractors) & 30.40~($\pm$21.11)  & 35.02~($\pm$25.60) & 36.38~($\pm$22.33)& 34.24~($\pm$15.02) & \textbf{28.24}~($\pm$16.92)  &  41.13~($\pm$11.90) \\
\midrule
\multirow{4}{*}{Entity} & Replace Names &  1.28~($\pm$0.91) & 1.21~($\pm$0.65) & 1.11~($\pm$0.43) &  1.51~($\pm$0.97) & 1.61~($\pm$1.18) &  \textbf{0.94}~($\pm$0.59)  \\
\cmidrule(lr){3-8}
& Replace Locations & 3.47~($\pm$1.54) & 3.53~($\pm$1.96) & \textbf{2.44}~($\pm$0.78) & 3.38~($\pm$1.92) & 3.07~($\pm$2.12) &  2.89~($\pm$1.02) \\
\cmidrule(lr){3-8}
& Replace Numbers & 0.58~($\pm$0.60) & 0.73~($\pm$0.70) &  0.71~($\pm$0.52) & \textbf{0.47}~($\pm$0.28) & 0.56~($\pm$0.41) &  0.69~($\pm$0.39)\\
\bottomrule 
\end{tabular}
}
\caption{\small \textbf{RQ4 - Functional Subtests for Sentiment Analysis - Label+Expl (\textsection \ref{sec:exp:rq1:functional}).}
For sentiment analysis, we compare various ER \textit{rationale alignment criteria} using the IxG machine rationale extractor (as well as the No-ER baseline), with respect to performance on a range of functional tests/subtests (OOD). For each functional test, we report model performance on each of its individual functional subtests. Performance is reported in terms of failure rate. 
}
\label{tab:app:functional:rq4_label_expl}
\end{table*}

\end{document}